%% file: main.tex
\documentclass[10pt,twocolumn,letterpaper]{article}

\usepackage{iccv}
\usepackage{times}
\usepackage{epsfig}
\usepackage{graphicx}
\usepackage{amsmath}
\usepackage{amssymb}

\usepackage{xcolor}
\usepackage{booktabs}
\usepackage{tabularx}
\usepackage{colortbl}
\usepackage{textcomp}
\usepackage{verbatim}
\usepackage{multirow}
\usepackage{makecell}
\usepackage{subcaption}
\usepackage{siunitx}
\usepackage{authblk}
\usepackage[T1]{fontenc}

\usepackage{pgf}
\usepackage{pgfplots}

\usepackage[pagebackref=true,
breaklinks=true,letterpaper=true,colorlinks, citecolor=cyan,bookmarks=false]{hyperref}
\usepackage[capitalize]{cleveref}
\usepackage[absolute,overlay]{textpos}

\iccvfinalcopy 

\def\iccvPaperID{****} 

\ificcvfinal\fi

\begin{document}

\begin{textblock*}{\textwidth}(18mm,12mm)
    \vspace{2mm}
    \tiny
    \centering
    Lorenzo Brigato, Stavroula Mougiakakou.\\
    ``No Data Augmentation?
    Alternative Regularizations for Effective Training on Small Datasets.''\\
    \textit{Proceedings of the IEEE/CVF International Conference on Computer Vision (ICCV) Workshops 2023.}\\
    \copyright\ 2023 IEEE. Personal use of this material is permitted.
    Permission from IEEE must be obtained for all other uses, in any current or future media, including reprinting/republishing this material for advertising or promotional purposes, creating new collective works, for resale or redistribution to servers or lists, or reuse of any copyrighted component of this work in other works.
    The final publication will be available at
    \href{https://ieeexplore.ieee.org/}{ieeexplore.ieee.org}.\\
    \vspace{2mm}
\end{textblock*}

\title{No Data Augmentation?\\
Alternative Regularizations for Effective Training on Small Datasets}

\author{Lorenzo Brigato, Stavroula Mougiakakou\\
AI in Health and Nutrition, ARTORG Center for Biomedical Engineering Research\\
University of Bern \\
{\tt\small lorenzo.brigato@unibe.ch, stavroula.mougiakakou@unibe.ch}
}

\maketitle
\ificcvfinal\thispagestyle{empty}\fi

\begin{abstract}
Solving image classification tasks given small training datasets remains an open challenge for modern computer vision.
Aggressive data augmentation and generative models are among the most straightforward approaches to overcoming the lack of data.
However, the first fails to be agnostic to varying image domains, while the latter requires additional compute and careful design.

In this work, we study alternative regularization strategies to push the limits of supervised learning on small image classification datasets.
In particular, along with the model size and training schedule scaling, we employ a heuristic to select (semi) optimal learning rate and weight decay couples via the norm of model parameters. 
By training on only 1\% of the original CIFAR-10 training set (i.e., 50 images per class) and testing on ciFAIR-10, a variant of the original CIFAR without duplicated images, we reach a test accuracy of 66.5\%, on par with the best state-of-the-art methods.
\end{abstract}

\section{Introduction}

In recent years, significant progress has been made in computer vision through large-scale pretraining on extensive datasets \cite{russakovsky2015imagenet, ridnik2021imagenet}.
However, improving the data efficiency of deep neural networks and enabling successful training on significantly smaller datasets, ranging from a few tens to hundreds of images per class, remains an ongoing area of research.
Better sample efficiency and generalization would greatly benefit domains where the high cost and limited accessibility of data collection and annotation are critical barriers (e.g., the medical domain).
The community has recently increased its focus toward studying limited-sample problems with deep learning through the organization of dedicated workshops and challenges \cite{bruintjes2021vipriors, lengyel2022vipriors, bruintjes2023vipriors}.
Furthermore, recent work has compared methods tailored explicitly for image classification with small datasets and established a dedicated benchmark  \cite{brigato2021tune, brigato2022image}.
A notable result of the latter analysis regards the importance of hyper-parameter tuning, particularly weight decay, which plays a significant role in the generalization ability of networks and has often been overlooked in previous works.
More in detail, a tuned vanilla cross-entropy classifier favorably compared against most of the evaluated data-efficient methods, powered by sophisticated techniques (e.g., \cite{bietti2019kernel, navon2021auxiliary}) and inductive biases (e.g., \cite{oyallon2017scaling, kayhan2020translation}).

\begin{figure}[t]
    \resizebox{\linewidth}{!}{\input{figures/paperdigest.pgf}}
    \caption{\textbf{Overview.}
    We examine each additional modification in our training setup (y-axis) and its corresponding impact (x-axis).
    The baseline WRN-16-1 model has been tuned for optimal learning rate and weight decay on a subset of the small training set.}
    \label{fig:topfig}
\end{figure}
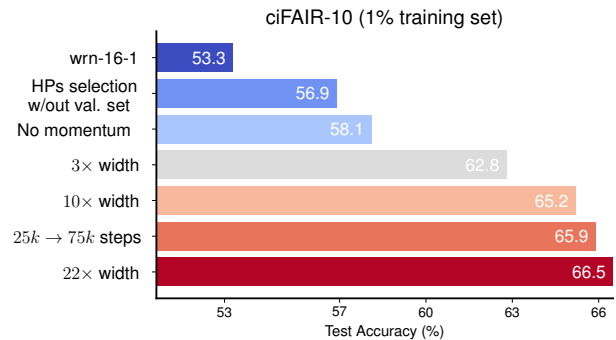

Classifiers augmented with aggressive data augmentation methods (e.g., AutoAugment \cite{cubuk2018autoaugment}) or generative models have recently scored the best results on multiple data-efficient image classification benchmarks \cite{azuri2021generative,chimera2022}.
While it is expected that additional data synthesis helps generalization, this family of approaches still presents challenges.
For instance, recent work has shown that data augmentation introduces strong class-dependent biases \cite{balestriero2022effects}.
Furthermore, the relevance of image transformations is domain dependent and requires domain expertise \cite{bendidi2023no}.
Generative models instead require sophisticated design, careful engineering, and multi-stage training \cite{zhang2019dada, azuri2021generative, chimera2022}. 


In this work, we investigate in detail the impact of optimization-related hyper-parameters (HPs) (i.e., learning rate, weight decay, and momentum), model size (in particular width), and training schedule length on the popular ciFAIR-10 small-data benchmark, which comprises 1\% of the original training set of CIFAR-10 and testing set without duplicated images \cite{barz2020cifair}.
Based on our empirical analysis, we devise a simple scheme to maximize the accuracy of a vanilla cross-entropy classifier by making it as data-efficient as state-of-the-art methods powered by strong data augmentation methods \cite{gauthier2022parametric,chimera2022}.

As visible in \cref{fig:topfig}, we start from a baseline Wide ResNet-16-1 (WRN-16-1) \cite{zagoruyko2016wrn}, tuned on the small validation set, which scores 53.3\% on the test set, and we reach a strong 66.5\% accuracy with WRN-16-22.
In particular, our proposed training setup involves a heuristic to select HPs without relying on validation sets (\cref{sec:hp_sel}), the removal of momentum (\cref{sec:no_mom}), the scaling of model size (\cref{sec:model_size}), and training length (\cref{sec:train_len}).

In summary, this paper builds a robust and easy-to-implement baseline for training efficiently vanilla cross-entropy classifiers on small datasets.
Furthermore, it provides insights regarding the impact of HPs, model scale, and training length.
We demonstrate that aggressive data augmentation is not the only way to reach the best performance in scenarios with limited data.
We hope that our empirical analysis could be helpful for practitioners and researchers involved in deploying and searching for more data-efficient image classifiers.

\section{Related Work}

\begin{figure*}[t]
\centering
    \resizebox{\linewidth}{!}{\includegraphics{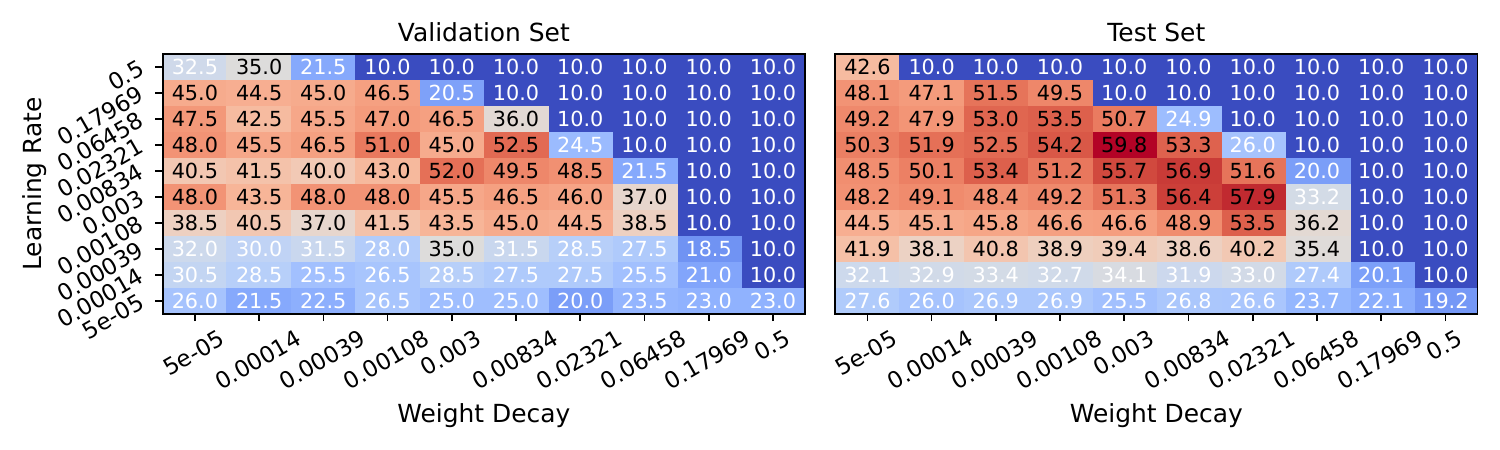}}
    \caption{\textbf{Impact of small validation sets.} Validation and test accuracy scored by a WRN-16-1 trained with momentum. Having only available a small training set can result in sub-optimal model selection on noisy validation sets.
    In this case, the best model on the validation set does not transfer to the best model on the test set.
    }
    \label{fig:val_test}
\end{figure*}

\paragraph{Impact of scaling model size and training length.}

Several studies have explored the effect of model scaling on performance.
For instance, convolutional networks can be scaled by depth \cite{he2016deep}, width \cite{zagoruyko2016wrn}, or the combination of the two along with the input resolution \cite{tan2019efficientnet}.
Other works studied the generalization of networks across data and model scaling \cite{hestness2017deep, rosenfeld2019constructive}, with some focusing on small data regimes \cite{bornschein2020small, brigato2021close}.
Differently from \cite{bornschein2020small, brigato2021close}, we experiment with a single dataset size and provide insights concerning the impact of optimization-related HPs, model, and training length scaling.

The relationship between generalization error and model size, with the empirical finding of the \textit{double descent} phenomenon, has been observed in older works \cite{vallet1989linear, opper1990ability, loog2020brief} and further investigated in the deep-learning era \cite{novak2018sensitivity, belkin2019reconciling, nakkiran2021deep, advani2020high}.
Although models of different sizes reach the same training errors, larger models tend to have smaller test errors \cite{advani2020high}.
While still under discussion in current research, possible explanations include that large models are more biased towards better minima \cite{du2018gradient, du2019gradient} or explore more features \cite{brutzkus2019larger}.
Finally, additional training iterations benefit generalization \cite{hoffer2017train, he2019rethinking}, and seem to generate a similar \textit{double descent} behavior but related to the length of training \cite{nakkiran2021deep, pezeshki2022multi}.
However, we are unaware of any preceding work studying the impact of the training length and focusing on image classification tasks with small datasets.

\paragraph{Scale-invariant networks.}

Normalization layers (e.g., Batch Normalization (BN) \cite{ioffe2015batch}) make modern neural networks almost fully scale-invariant. In other words, their output activations, and consequently, the loss function, does not change if the weights undergo scaling, implying that weight decay does not limit the model capacity as previously believed \cite{van2017l2}.
The training dynamics of Stochastic Gradient Descent (SGD) and variants have been widely investigated and are still under discussion from both an empirical and theoretical perspective \cite{hoffer2018norm, zhang2018three, li2020reconciling, wan2021spherical, kodryan2022training}. 
The parameters' norm strongly impacts the effective learning rate, the actual step which a scale-invariant network would take if optimized over the unit sphere \cite{wan2021spherical, kodryan2022training}. 
Recent work has practically studied predicting and scheduling optimal HPs by exploiting SGD symmetries as data scales \cite{yun2020weight, yun2022angular}.
Our paper not only focuses on HPs selection but also analyses the impact of model size and training length.

\paragraph{Image classification with small datasets.}
Learning from a  small sample is an actual challenge for deep learning.
and shares the goal of deploying data-efficient classifiers with other popular research areas, such as transfer learning \cite{pan2009survey, kornblith2019better}, domain adaptation \cite{wang2018deep}, and  few-shot learning \cite{wang2020generalizing}.
However, such research domains assume access to a generally extensive annotated database on which networks can be trained.
This assumption is not always satisfactory, notably when the domain where the network is transferred dramatically differs from the original one.

We refer the reader to \cite{brigato2022image} for a detailed overview concerning learning methods tailored explicitly for learning from scratch on small datasets.
Some methods benefit from employing \emph{geometric priors}, such as fixed or learnable filters based on wavelet transformations \cite{oyallon2017scaling, oyallon2018scattering, gauthier2022parametric} or discrete cosine transform \cite{ulicny2019harmonic, ulicny2019harmonicnet}. 
Invariance to input transformations (e.g., rotation, translation) is achieved by integrating steerable filters or circular harmonics \cite{worrall2017harmonic}, alternative padding strategies \cite{kayhan2020translation}, and specialized convolution blocks \cite{xu2020towards, sun2020visual}.  
Cost-based regularization strategies formulate objective functions and penalties to mitigate overfitting \cite{navon2021auxiliary}, such as the cosine loss and variants \cite{barz2020deep, kobayashi2021t, sun2020visual}.
Other cost-based regularizers include rotation invariance \cite{xu2020towards}, gradient penalties and spectral norms \cite{bietti2019kernel}, low-rank embedding \cite{lezama2018ole}, and temperature calibration \cite{bornschein2020small}.
Another set of approaches performs data augmentation on the input space by relying on generative models \cite{zhang2019dada, zhang2020deep, azuri2021generative, chimera2022}, or on the network's feature space \cite{ishii2019training, keshari2019guided, lin2020efficient, lin2020latent}.
Finally, some previous work warm-start the final classifier after solving a pretext task through layer-wise greedy initialization \cite{rueda2015supervised}, adaptive model complexity \cite{feng2015learning}, dictionary-based learning \cite{keshari2018learning}, or self-supervised pre-training \cite{zhao2020distilling, wad2022equivariance}.

Our work shares with previous work \cite{brigato2021tune, barz2021strong} the interest in improving vanilla cross-entropy classifiers on limited data settings.
Differently, we perform a comprehensive analysis concerning the impact of model size and training schedule length, which is completely missing in \cite{brigato2021tune}.
Further, we propose additional insights regarding the search for optimal optimization parameters and the impact of momentum.

\section{Preliminaries}
\label{seq:prelim}

We face an image classification problem in which we are given a small set of \(N\) labeled pairs \(\mathcal{D} = \{x_{i}, y_{i}\}_{i=1:N}\) sampled from distributions \(\mathcal{X}\) and \(\mathcal{Y}\).
We train function approximators \(f_{\theta}\) (WRNs) with mini-batches of dimension \(B\) to optimize the objective function \(J_{\theta} = \frac{1}{B}\sum_{x,y \sim \mathcal{D}} J(f_{\theta}(x), y))\).
The networks are trained for \(T\) iterations with SGD and its variants with momentum (\(\mu\)) and weight decay (\(\lambda\)).
The latter explicitly penalizes the \(L_{2}\) squared norm of the weights divided by two.
At each training step \(t\), the parameters follow the update rule:

\begin{equation}
\label{eq:sgd_update}
\begin{array}{ll}
v_{t+1} = \mu  v_{t} + \alpha_{t}  (\nabla J_{\theta} + \lambda  \theta_{t}) & \\
\theta_{t+1} = \theta_{t} - v_{t+1}
\end{array}
\end{equation}

with \(\alpha_{t}\) being the learning rate adjusted at each iteration step according to a defined learning rate schedule.
We instead refer to \(\alpha\) as the initial learning rate.
If we consider the simpler case without momentum (i.e., \(\mu = 0.0\)), the general SGD update reported in \cref{eq:sgd_update} can be decoupled into a weight decay step \(\theta_{t+1} = \theta_{t}  (1 - \alpha_{t} \lambda\)) and a gradient descent one \(\theta_{t+1} = \theta_{t} - \alpha_{t} \nabla J_{\theta}\).
The weight decay update is ruled by the product between \(\alpha_{t}\) and \(\lambda\), which is referred as the \textit{effective weight decay} in \cite{hanson1988comparing}.
If we assume scale-invariance\footnote{All layers of WRNs are scale-invariant except for the BN affine parameters and final classification head.}, i.e., \(J_{\theta} = J_{c\cdot\theta}, c > 0\), it follows that \(\nabla J_{\theta} \cdot \theta = 0\) \cite{van2017l2, li2020reconciling}.
Hence, each SGD step encompasses a combination of two conflicting forces.
The \textit{effective weight decay} diminishes the parameter norm, whereas the gradient amplifies it, resulting in a dynamic interplay between the two.

\section{Experiments}

To perform our empirical analyses, we choose the popular WRN architecture of depth 16 widely used in previous work on the small ciFAIR-10 dataset \cite{oyallon2017scaling, brigato2021tune}, and vary the width to increase model size when necessary.
We fix the batch size \(B\) for all the training runs to 10, given the success of small batches in small-data regimes \cite{brigato2021tune, brigato2022image}.
In addition, we incorporate the widely used cosine annealing schedule to adjust the learning rate during training \cite{loshchilov2016sgdr}.
To have a good glimpse of the impact of the learning rate and weight decay on the generalization performance, for most of the networks, we run grid searches with 100 models, sampling equally spaced learning rate and weight decay values in log-space from the interval \([5 \cdot 10^{-5}, 5 \cdot  10^{-1}]\).
We only run a sub-portion of the grid for bigger models that would have required an onerous amount of compute.
We finally employ minimal data augmentation composed of random horizontal flipping and translations of 4 pixels.

\subsection{Baseline setup}

\label{sec:model_sel}

\begin{figure*}[t]
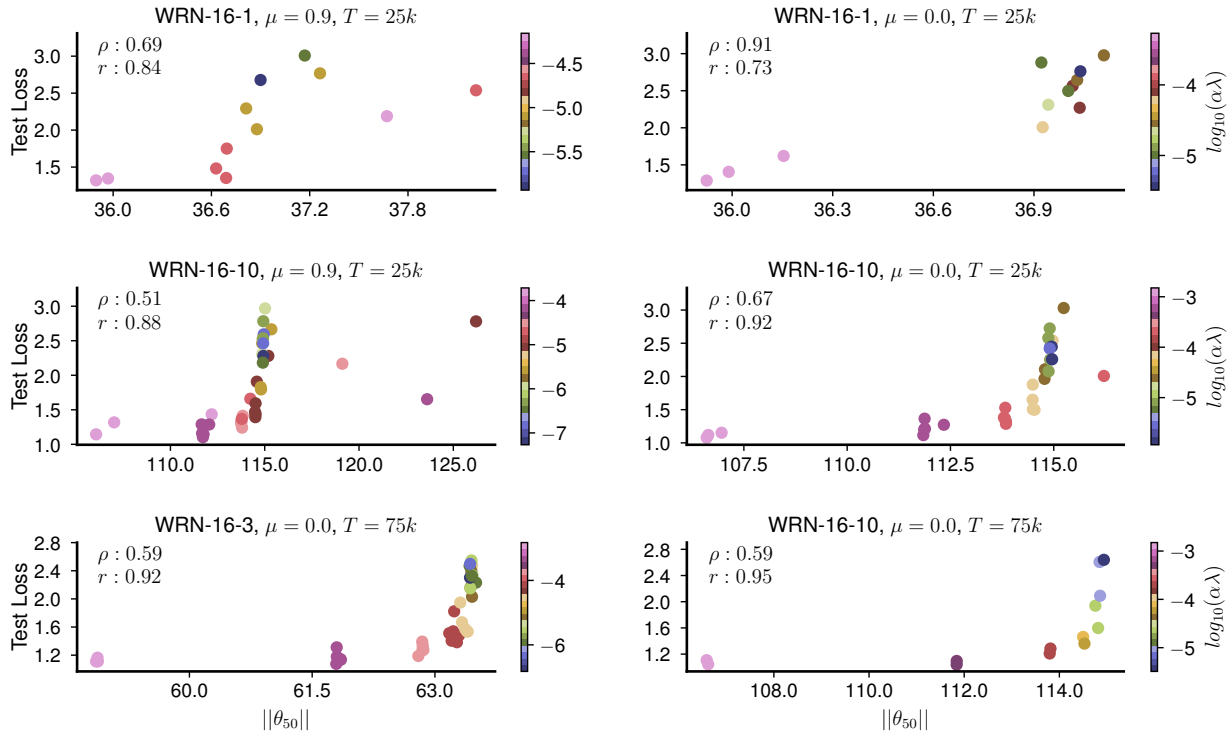

\centering
    \resizebox{0.48\linewidth}{!}{\input{figures/fig_correlation_wrn-16-1_mu-09.pgf}}
    \resizebox{0.48\linewidth}{!}{\input{figures/fig_correlation_wrn-16-1_mu-00.pgf}}
    \resizebox{0.48\linewidth}{!}{\input{figures/fig_correlation_wrn-16-10_mu-09.pgf}}
    \resizebox{0.48\linewidth}{!}{\input{figures/fig_correlation_wrn-16-10_mu-00.pgf}}
    \resizebox{0.48\linewidth}{!}{\input{figures/fig_correlation_wrn-16-3_mu-00_T-75k.pgf}}
    \resizebox{0.48\linewidth}{!}{\input{figures/fig_correlation_wrn-16-10_mu-00_T-75k.pgf}}
    
    \caption{\textbf{HPs selection via parameters norm.} Relationship between the norm of the parameters after one training epoch (50 iterations) and test loss for different configurations (specified in figure titles).
    Each network, represented as a dot, has scored 100\% training accuracy on the training set. 
    Colors represent the product between learning rate and weight decay, \(\mu\) is momentum, and \(T\) is the total number of training steps.
    Pearson and Spearman's coefficients are indicated with \(p\) and \(r\).
    }
    \label{fig:correlation}
\end{figure*}

As a base setup, we choose i) the smallest architecture of the WRN-16 family, i.e., WRN-16-1, which is computationally cheap to train; ii) a training schedule of \(25k\) steps as proposed in \cite{brigato2021tune};
iii) momentum \(\mu = 0.9\) as standard practice in deep learning; iv) HPs selection on a small validation set with the aforementioned grid search.
In particular, we employ the training-validation split proposed in \cite{brigato2021tune}.

In \cref{fig:val_test}, we show the results of the grid searches for both validation and test sets.
Given the accuracy score on the validation set, we select the model scoring around 53.3\% on the testing set.
However, we also note that the best learning rate and weight decay combination found in the validation set does not transfer to the optimal model, and the best-achieved accuracy on the test set is already higher than previously published results of larger networks, e.g., WRN-16-8 \cite{oyallon2017scaling, ulicny2019harmonicnet, brigato2021tune, brigato2022image}.
Reasonably, the search for HPs is particularly noisy and sub-optimal because we face a learning task in the small-sample regime.
Hence, we argue that better HPs selection has the potential to deliver networks that generalize better, particularly for larger models, as we have just observed that a tiny WRN-16-1 coupled with optimal parameters could outperform the best accuracy of a larger WRN-16-8. 

\subsection{HPs selection without validation sets}
\label{sec:hp_sel}

\begin{figure*}[t]
\centering
    \resizebox{\linewidth}{!}{\input{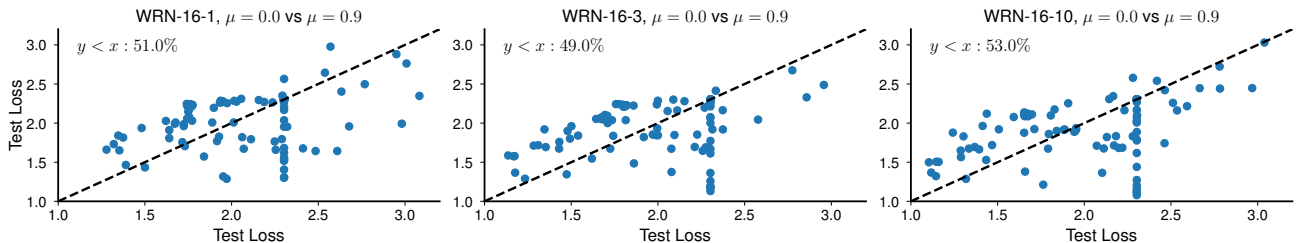}}
    \caption{\textbf{Impact of momentum.} Comparison among three couples of architectures in terms of testing loss without (\(\mu = 0.0\)) and with (\(\mu = 0.9\)) momentum.
    The losses of networks trained without momentum are shown on the y-axis.
    Momentum does not seem to provide clear benefits in the optimization leading to similarly performing networks.}
    \label{fig:momentum}
\end{figure*}

We devise a straightforward heuristic that effectively predicts the generalization performance of models by only monitoring training-related metrics.
In this manner, we circumvent the requirement of relying on held-out validation sets, which may be limited and noisy in small-sample regimes.
We first filter out all networks that do not fit the training set, i.e., those that do not score \(100\%\) training accuracy and hence do not have enough representation power to converge \cite{arpit2017closer}.
Secondly, out of this pool of models, we consider the parameter vector norm \(||\theta_{t}||\) at the beginning of training to be a good predictor for the testing loss.
Previous work supports our intuitive approach by showing that regularization (e.g., weight decay) mostly affects early training dynamics \cite{golatkar2019time}.

In \cref{fig:correlation}, we plot the test loss as a function of the norm after one epoch, which coincides with as few as 50 steps, i.e., \(||\theta_{50}||\).
We represent models that share the same learning rate-weight decay product in the same colors. 
A robust monotonic relationship exists between the two variables, as indicated by Spearman's rank coefficient surpassing 0.8 most of the time.
The models with the smallest norm are the ones that generalize better by scoring lower testing losses.
The monotonicity increases as the model size and training length increase.
Reasonably, models with similar initial \(\textit{effective weight decay}\) share norm magnitudes since their parameter vector is equally decayed.
The symmetries across the learning rate-weight decay space (left-to-right diagonals) are also visible in \cref{fig:val_test} (right).
However, not all the models generalize the same along a constant $\alpha \lambda$ since the gradient update is proportional to only \(\alpha\), not \(\alpha \lambda\).
Momentum introduces some additional noise, potentially attributable to the more complex dynamics of incorporating previous gradients.
However, the monotonic relationship remains reliable also if \(\mu = 0.9\).

By using the parameter's norm to select the HPs, we raise the accuracy of the base WRN-16-1 from 53.3\% to 56.9\%.
We will use this model-selection strategy in the next experiments.

\subsection{Removal of momentum}
\label{sec:no_mom}

Momentum is widely used in the deep learning community.
Recent work has shown that it reduces the distance traveled by the parameters over the loss landscape \cite{heo2020adamp}.
Furthermore, momentum makes the training dynamics slightly more complex due to past-gradients additions.
We conducted experiments to assess the effect of momentum in our constrained data conditions using three models: WRN-16 with width scales of 1, 3, and 10. All six models underwent training for 25,000 steps. The test losses for each architecture, both with and without momentum, were compared, as depicted in \cref{fig:momentum}. 
Remarkably, approximately 50\% of the time, the best models are either with \(\mu = 0.0\) or \(\mu = 0.9\), indicating a similar test performance.
These results suggest that making the SGD trajectories noisier may not necessarily penalize learning in limited data scenarios. 
To this end,  we remove momentum and maintain more predictable training dynamics.
In this manner, our momentum-free WRN-16-1 reaches a test accuracy of 58.1\%, higher than the previous 56.9\%.
Notably, by removing momentum and performing HPs selection with our newly introduced metric (parameters' norm), we made a small WRN-16-1 as data-efficient as a larger WRN-16-8 tuned with Asynchronous HyperBand with Successive Halving (ASHA) search, which scored on the same benchmark 58.2\% test accuracy \cite{brigato2021tune}.  

\begin{figure}
    \centering  
    \resizebox{0.9\linewidth}{!}{\input{figures/tlength_acc.pgf}}
    \caption{\textbf{Impact of training length.} Maximal achievable test accuracy as a function of the employed architecture and number of training iterations.
    A longer training schedule improves generalization.}
    \label{fig:tlength}
\end{figure}
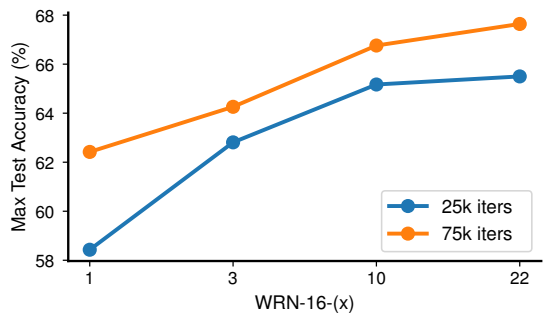

\subsection{Increased model size}
\label{sec:model_size}

\begin{figure*}[t]
\centering
    \resizebox{\linewidth}{!}{\includegraphics{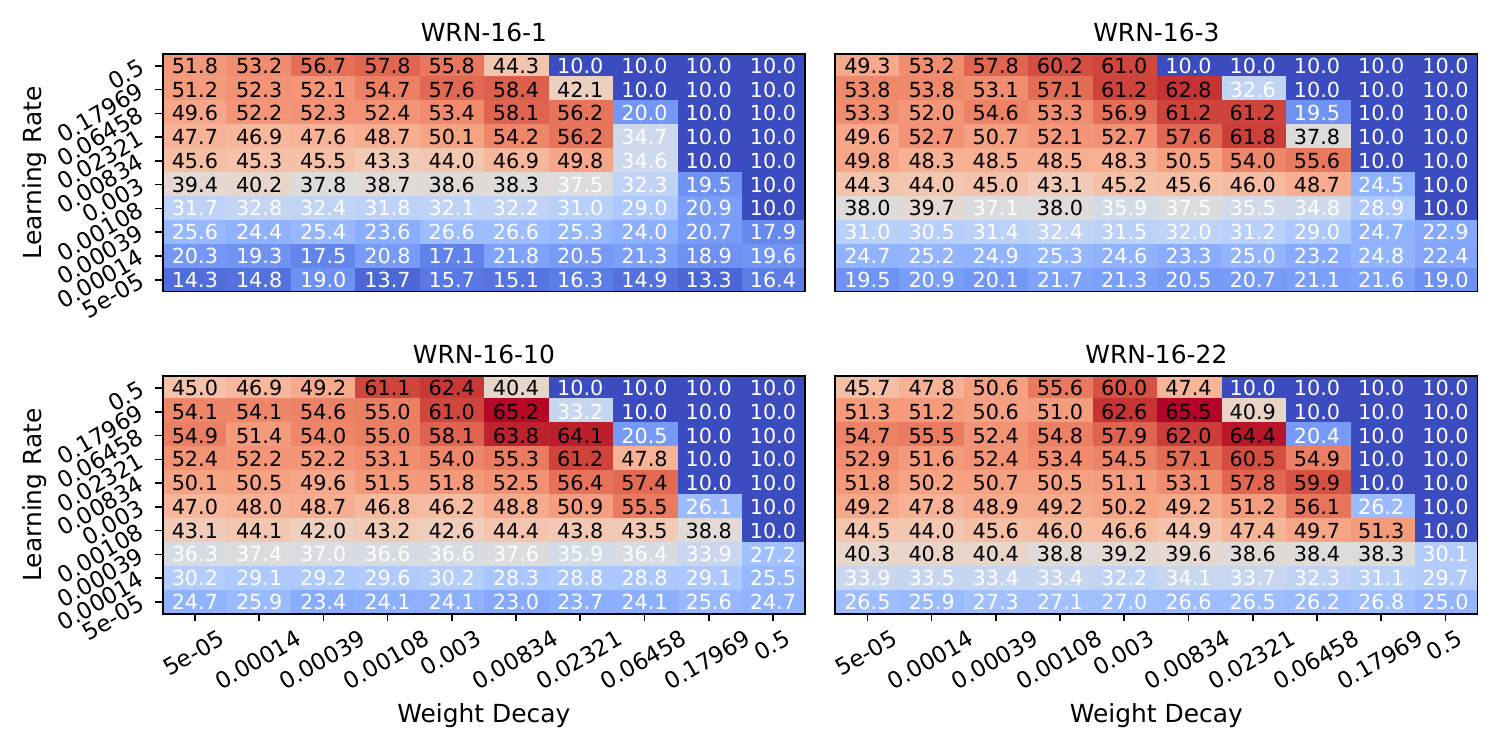}}
    \caption{\textbf{Impact of model scale.} Test accuracy over the predefined learning rate-weight decay space.
    Increased model width significantly improves maximal achievable test accuracy but plateaus when moving from WRN-16-10 to WRN-16-22.
    All networks are trained for $25k$ iterations.}
    \label{fig:scale}
\end{figure*}

Scaling up model size is a popular way to improve generalization \cite{he2016deep}.
However, with limited data, scaling the model without providing the right amount of regularization easily leads to overfitting.
To better analyze the impact of scale, we report the test accuracy of WRN-16-1, WRN-16-3, and WRN-16-10, all trained without momentum in \cref{fig:scale}.

Increasing the width by 3\(\times\) already provides a maximum increase of 4.4 percentage points.
The best achievable accuracy rises from 62.8\% with WRN-16-3 to 65.2\% with WRN-16-10.
Our HPs selection metric correctly predicts the optimal learning rate-weight decay combination, and hence we gain 7.1 percent points to reach 65.2\% test accuracy from the previous 58.2\%.  

\subsection{Increased training length}
\label{sec:train_len}
 
Prior empirical evidence indicates that extended training schedules have demonstrated comparable performance to pre-trained networks \cite{he2019rethinking}.
The limited data in small-sample scenarios bears the risk of under-training networks if the number of epochs and batch size are directly imported from the default setups with more data, as in \cite{navon2021auxiliary, kobayashi2021t}, because that would result in a lower number of actual training steps.
Indeed, previous work showed that the number of training updates plays the most important role in learning \cite{hoffer2017train}.

To this end, we test a longer training schedule that closely matches the one originally proposed in the paper that introduced the WRN architecture \cite{zagoruyko2016wrn}.
In particular, WRNs were trained on 50,000 samples for 200 epochs and mini-batches of size 128, resulting in a training schedule of \(\sim 78k\) steps.
To match this length, we triplicate the number of epochs from 500 to 1,500 while maintaining the batch size of dimension 10 to get a total of \(75k\) training steps.

At all model scales, the tested networks improve their testing accuracy (see \cref{fig:tlength}).
In particular, the smallest WRN obtained the highest gain of 4 percent points.
Not negligible improvements of 1.5, 1.6, and 2.1 percent points are scored by networks of widths 3,10 and 22, respectively.
We also tested a longer training schedule of 4,500 epochs for the WRN-16-1 in preliminary experiments.
We have not obtained significant improvements and hence stopped at 3,000.
However, we do not rule out that increased training time could provide additional but moderate gains at large model scales.

Our final architecture becomes the WRN-16-22 trained for $75k$ iterations.
The model selection strategy predicts the second-best model, which slightly underperforms the highest-scoring one (66.5\% vs 67.6\%).
Increasing the model width from 10 to 22 and tripling the training length make us gain 1.3 percent points over the previous setup.

\begin{figure*}[t]
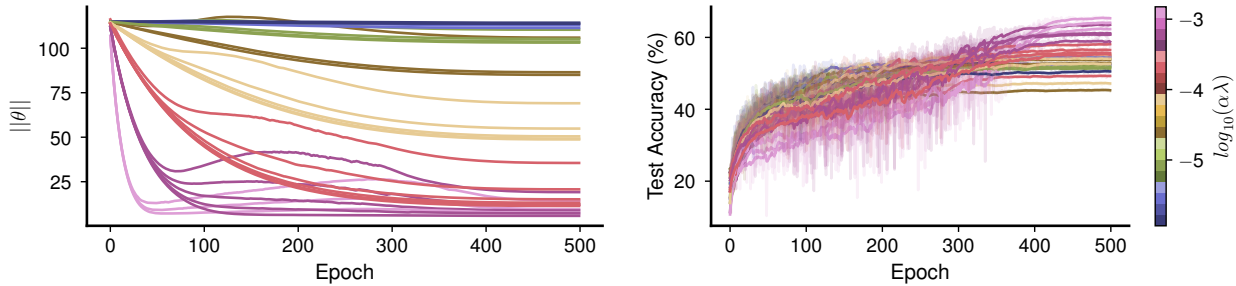

    \centering
    \resizebox{0.48\linewidth}{!}{\input{figures/fig_normw_perepoch.pgf}}
    \resizebox{0.48\linewidth}{!}{\input{figures/fig_accuracy_perepoch.pgf}}
    \caption{\textbf{Norm and test accuracy evolution.} We show the evolution of the weights norm and test accuracy for the WRN-16-10s reaching 100\% training accuracy and trained for 500 epochs ($25k$ iterations).
    The models with the highest \(\alpha \lambda\) products experience more chaotic training dynamics (noisy test accuracy profile), fast decay of parameters norm, and better generalization. 
    }
    \label{fig:norm_testacc}
\end{figure*}

\subsection{Comparison with the state of the art}

 \begin{table}
    {\begin{tabularx}{\linewidth}{cccc}
         \toprule
          Pub.   & Architecture & Augmentation & Accuracy\\
          \midrule
          \cite{brigato2021tune}         & WRN-16-8 & plain & $58.22$ \\
          \cite{gauthier2022parametric}  & Scatt. WRN & AutoAugment & ${63.13 \pm 0.29}^{*}$\\
          \cite{chimera2022}             & WRN-16-8 & MixUp &  $66.16 \pm 0.78$ \\
          \cite{chimera2022}             & WRN-16-8 & ChimeraMix$^{1}$ & $65.83 \pm 0.78$ \\
          \cite{chimera2022}             & WRN-16-8 & ChimeraMix$^{2}$ & $67.30 \pm 1.21$ \\

          Ours                           & WRN-16-10 & plain & $65.9$ \\
          Ours                           & WRN-16-22 & plain & $66.5$  \\
          \bottomrule
     \end{tabularx}
    }
     \caption{\textbf{Comparison with state-of-the-art methods.} All networks are trained on CIFAR-10 with 50 samples per class.
     $^{*}$Scattering WRN has 22.6M parameters and is evaluated on the CIFAR-10 test set rather than ciFAIR-10.
     ChimeraMix$^{1}$ employs a grid-based patch selection while ChimeraMix$^{2}$ a gradient-based methodology.
     Plain augmentation is composed of simple horizontal flipping and 4-pixel translations.}
     \label{tab:sota}
 \end{table}
 
In the preceding sections, we tested and discussed several design choices to enhance our training scheme's overall performance without relying on hand-crafted data augmentations or costly generative models.
 
To gauge the effectiveness of our approach, we now compare our WRN-16-22 against the best state-of-the-art methods.
In particular, we benchmark against WRN-16-8 architectures trained with cross-entropy loss \cite{brigato2021tune} plus basic, i.e., translation and horizontal flipping, or strong data augmentation methods such as MixUp \cite{zhang2017mixup} or ChimeraMix \cite{chimera2022}.
The hyper-parameters, i.e., learning rate and weight decay, were selected through ASHA search in the above cases.
We also report the performance of recent parametric scattering networks \cite{gauthier2022parametric} powered with AutoAugment \cite{cubuk2018autoaugment}.

We show the results in \cref{tab:sota}.
Our WRN-16-10 and WRN-16-22 architectures trained with our scheme achieve recognition performance on par with ChimeraMix and MixUp and significantly outperform the WRN-16-8 from \cite{brigato2021tune} and scattering networks coupled with AutoAugment \cite{gauthier2022parametric}.
Our reliance on plain data augmentation and implicit regularization techniques proves advantageous, as it enables our solution to generalize effectively across various domains, enhancing its practicality and transferability.
Furthermore, our scheme could be theoretically coupled with such powerful data augmentation techniques if the image domain is agnostic to the biases introduced by hand-crafted augmentations or if enough computational resources are available to train generative models properly.

\subsection{Additional Analyses}

\paragraph{Importance of HPs selection.}

We highlight that properly selecting hyper-parameters, particularly weight decay, is fundamental to providing optimal performance.
For instance, referring to \cref{fig:scale}, if the value of weight decay is set too small (\(5\cdot 10^{-5}\)), and a line search is performed over the learning rate, the maximum test accuracy improvement among WRN-16-1 and WRN-16-22 is approximately three percentage points.
On the other hand, if the search is also expanded over the weight decay direction, the gain almost doubles to 7 percentage points.

\paragraph{Chaotic train dynamics generalize better.} 

In \cref{fig:norm_testacc}, we provide additional insights regarding the evolution of the parameters norm and generalization through the test accuracy in the case of WRN-16-10 trained for \(25k\) iterations.
The largest weight decay-learning rate combinations that manage to fit the training set cause a fast decay of the parameters norm (as studied in \cref{sec:model_sel}) and also chaotic training dynamics.
The right plot of \cref{fig:norm_testacc} shows that a high \(\alpha \lambda\) combination generates noisy test accuracy profiles and late convergence.
Our findings align with previous studies \cite{li2019towards, iyer2023wide, kodryan2022training}, which suggest that training with higher learning rates leads to solutions with improved sharpening and generalization profiles.

\paragraph{HPs transfer across model sizes.}

Interestingly, it is also visible that the difference in parameter norm at the start of training due to increased model size drifts the area of better generalization towards the bottom right. 
This is partially explainable because the weight decay, as mentioned in \cref{sec:model_sel}, directly scales the weight vector by \(\alpha_t \lambda\) while the gradient update does not depend on the parameter norm but just the learning rate. 
Consequently, when the weight norm increases, the gradient step becomes smaller than the weight decay update.
However, as visible in \cref{fig:scale}, the best HPs combination remains constant across sizes, although the number of parameters has increased from 0.17M of WRN-16-1 to approximately 82.73M of WRN-16-22.
Further investigations are necessary to gain a deeper understanding of this phenomenon. 
The consistency of optimal HPs presents a promising avenue for future research, offering potential computational savings and improved efficiency.

\section{Conclusions}

In this work, we presented and ablated a simple methodology to push the limits of classifier recognition performance with small training datasets in image classification.

While approaches based on aggressive data augmentation and generative models can raise classification abilities through data synthesis, they still have limitations, such as being domain-specific or requiring extensive computational resources and careful design.
On the other hand, we explored several factors to improve the model's performance with alternative regularizations, including selecting optimal HPs more reliably and scaling the model size and training schedule.
By implementing these techniques, we achieved state-of-the-art performance on the popular ciFAIR-10 small-data benchmark, demonstrating the validity of our empirical analyses.

Although tested on a single dataset, our work provides valuable insights that can benefit practitioners and researchers interested in developing strategies to improve generalization in small-data settings.

{\small
\bibliographystyle{ieee_fullname}
\bibliography{egbib}
}

\end{document}

%% file: figures/paperdigest.pgf
\begingroup%
\makeatletter%
\begin{pgfpicture}%
\pgfpathrectangle{\pgfpointorigin}{\pgfqpoint{7.000000in}{4.000000in}}%
\pgfusepath{use as bounding box, clip}%
\begin{pgfscope}%
\pgfsetbuttcap%
\pgfsetmiterjoin%
\definecolor{currentfill}{rgb}{1.000000,1.000000,1.000000}%
\pgfsetfillcolor{currentfill}%
\pgfsetlinewidth{0.000000pt}%
\definecolor{currentstroke}{rgb}{1.000000,1.000000,1.000000}%
\pgfsetstrokecolor{currentstroke}%
\pgfsetdash{}{0pt}%
\pgfpathmoveto{\pgfqpoint{0.000000in}{0.000000in}}%
\pgfpathlineto{\pgfqpoint{7.000000in}{0.000000in}}%
\pgfpathlineto{\pgfqpoint{7.000000in}{4.000000in}}%
\pgfpathlineto{\pgfqpoint{0.000000in}{4.000000in}}%
\pgfpathlineto{\pgfqpoint{0.000000in}{0.000000in}}%
\pgfpathclose%
\pgfusepath{fill}%
\end{pgfscope}%
\begin{pgfscope}%
\pgfsetbuttcap%
\pgfsetmiterjoin%
\definecolor{currentfill}{rgb}{1.000000,1.000000,1.000000}%
\pgfsetfillcolor{currentfill}%
\pgfsetlinewidth{0.000000pt}%
\definecolor{currentstroke}{rgb}{0.000000,0.000000,0.000000}%
\pgfsetstrokecolor{currentstroke}%
\pgfsetstrokeopacity{0.000000}%
\pgfsetdash{}{0pt}%
\pgfpathmoveto{\pgfqpoint{1.750834in}{0.611927in}}%
\pgfpathlineto{\pgfqpoint{6.850000in}{0.611927in}}%
\pgfpathlineto{\pgfqpoint{6.850000in}{3.596774in}}%
\pgfpathlineto{\pgfqpoint{1.750834in}{3.596774in}}%
\pgfpathlineto{\pgfqpoint{1.750834in}{0.611927in}}%
\pgfpathclose%
\pgfusepath{fill}%
\end{pgfscope}%
\begin{pgfscope}%
\pgfpathrectangle{\pgfqpoint{1.750834in}{0.611927in}}{\pgfqpoint{5.099166in}{2.984847in}}%
\pgfusepath{clip}%
\pgfsetbuttcap%
\pgfsetmiterjoin%
\definecolor{currentfill}{rgb}{0.229806,0.298718,0.753683}%
\pgfsetfillcolor{currentfill}%
\pgfsetlinewidth{0.000000pt}%
\definecolor{currentstroke}{rgb}{0.000000,0.000000,0.000000}%
\pgfsetstrokecolor{currentstroke}%
\pgfsetstrokeopacity{0.000000}%
\pgfsetdash{}{0pt}%
\pgfpathmoveto{\pgfqpoint{-14.523748in}{3.461100in}}%
\pgfpathlineto{\pgfqpoint{2.607391in}{3.461100in}}%
\pgfpathlineto{\pgfqpoint{2.607391in}{3.141865in}}%
\pgfpathlineto{\pgfqpoint{-14.523748in}{3.141865in}}%
\pgfpathlineto{\pgfqpoint{-14.523748in}{3.461100in}}%
\pgfpathclose%
\pgfusepath{fill}%
\end{pgfscope}%
\begin{pgfscope}%
\pgfpathrectangle{\pgfqpoint{1.750834in}{0.611927in}}{\pgfqpoint{5.099166in}{2.984847in}}%
\pgfusepath{clip}%
\pgfsetbuttcap%
\pgfsetmiterjoin%
\definecolor{currentfill}{rgb}{0.435815,0.570707,0.951717}%
\pgfsetfillcolor{currentfill}%
\pgfsetlinewidth{0.000000pt}%
\definecolor{currentstroke}{rgb}{0.000000,0.000000,0.000000}%
\pgfsetstrokecolor{currentstroke}%
\pgfsetstrokeopacity{0.000000}%
\pgfsetdash{}{0pt}%
\pgfpathmoveto{\pgfqpoint{-14.523748in}{3.062056in}}%
\pgfpathlineto{\pgfqpoint{3.764466in}{3.062056in}}%
\pgfpathlineto{\pgfqpoint{3.764466in}{2.742821in}}%
\pgfpathlineto{\pgfqpoint{-14.523748in}{2.742821in}}%
\pgfpathlineto{\pgfqpoint{-14.523748in}{3.062056in}}%
\pgfpathclose%
\pgfusepath{fill}%
\end{pgfscope}%
\begin{pgfscope}%
\pgfpathrectangle{\pgfqpoint{1.750834in}{0.611927in}}{\pgfqpoint{5.099166in}{2.984847in}}%
\pgfusepath{clip}%
\pgfsetbuttcap%
\pgfsetmiterjoin%
\definecolor{currentfill}{rgb}{0.667253,0.779176,0.992959}%
\pgfsetfillcolor{currentfill}%
\pgfsetlinewidth{0.000000pt}%
\definecolor{currentstroke}{rgb}{0.000000,0.000000,0.000000}%
\pgfsetstrokecolor{currentstroke}%
\pgfsetstrokeopacity{0.000000}%
\pgfsetdash{}{0pt}%
\pgfpathmoveto{\pgfqpoint{-14.523748in}{2.663012in}}%
\pgfpathlineto{\pgfqpoint{4.150158in}{2.663012in}}%
\pgfpathlineto{\pgfqpoint{4.150158in}{2.343777in}}%
\pgfpathlineto{\pgfqpoint{-14.523748in}{2.343777in}}%
\pgfpathlineto{\pgfqpoint{-14.523748in}{2.663012in}}%
\pgfpathclose%
\pgfusepath{fill}%
\end{pgfscope}%
\begin{pgfscope}%
\pgfpathrectangle{\pgfqpoint{1.750834in}{0.611927in}}{\pgfqpoint{5.099166in}{2.984847in}}%
\pgfusepath{clip}%
\pgfsetbuttcap%
\pgfsetmiterjoin%
\definecolor{currentfill}{rgb}{0.867428,0.864377,0.862602}%
\pgfsetfillcolor{currentfill}%
\pgfsetlinewidth{0.000000pt}%
\definecolor{currentstroke}{rgb}{0.000000,0.000000,0.000000}%
\pgfsetstrokecolor{currentstroke}%
\pgfsetstrokeopacity{0.000000}%
\pgfsetdash{}{0pt}%
\pgfpathmoveto{\pgfqpoint{-14.523748in}{2.263968in}}%
\pgfpathlineto{\pgfqpoint{5.660784in}{2.263968in}}%
\pgfpathlineto{\pgfqpoint{5.660784in}{1.944733in}}%
\pgfpathlineto{\pgfqpoint{-14.523748in}{1.944733in}}%
\pgfpathlineto{\pgfqpoint{-14.523748in}{2.263968in}}%
\pgfpathclose%
\pgfusepath{fill}%
\end{pgfscope}%
\begin{pgfscope}%
\pgfpathrectangle{\pgfqpoint{1.750834in}{0.611927in}}{\pgfqpoint{5.099166in}{2.984847in}}%
\pgfusepath{clip}%
\pgfsetbuttcap%
\pgfsetmiterjoin%
\definecolor{currentfill}{rgb}{0.968203,0.720844,0.612293}%
\pgfsetfillcolor{currentfill}%
\pgfsetlinewidth{0.000000pt}%
\definecolor{currentstroke}{rgb}{0.000000,0.000000,0.000000}%
\pgfsetstrokecolor{currentstroke}%
\pgfsetstrokeopacity{0.000000}%
\pgfsetdash{}{0pt}%
\pgfpathmoveto{\pgfqpoint{-14.523748in}{1.864925in}}%
\pgfpathlineto{\pgfqpoint{6.432167in}{1.864925in}}%
\pgfpathlineto{\pgfqpoint{6.432167in}{1.545690in}}%
\pgfpathlineto{\pgfqpoint{-14.523748in}{1.545690in}}%
\pgfpathlineto{\pgfqpoint{-14.523748in}{1.864925in}}%
\pgfpathclose%
\pgfusepath{fill}%
\end{pgfscope}%
\begin{pgfscope}%
\pgfpathrectangle{\pgfqpoint{1.750834in}{0.611927in}}{\pgfqpoint{5.099166in}{2.984847in}}%
\pgfusepath{clip}%
\pgfsetbuttcap%
\pgfsetmiterjoin%
\definecolor{currentfill}{rgb}{0.905783,0.455186,0.355336}%
\pgfsetfillcolor{currentfill}%
\pgfsetlinewidth{0.000000pt}%
\definecolor{currentstroke}{rgb}{0.000000,0.000000,0.000000}%
\pgfsetstrokecolor{currentstroke}%
\pgfsetstrokeopacity{0.000000}%
\pgfsetdash{}{0pt}%
\pgfpathmoveto{\pgfqpoint{-14.523748in}{1.465881in}}%
\pgfpathlineto{\pgfqpoint{6.657154in}{1.465881in}}%
\pgfpathlineto{\pgfqpoint{6.657154in}{1.146646in}}%
\pgfpathlineto{\pgfqpoint{-14.523748in}{1.146646in}}%
\pgfpathlineto{\pgfqpoint{-14.523748in}{1.465881in}}%
\pgfpathclose%
\pgfusepath{fill}%
\end{pgfscope}%
\begin{pgfscope}%
\pgfpathrectangle{\pgfqpoint{1.750834in}{0.611927in}}{\pgfqpoint{5.099166in}{2.984847in}}%
\pgfusepath{clip}%
\pgfsetbuttcap%
\pgfsetmiterjoin%
\definecolor{currentfill}{rgb}{0.705673,0.015556,0.150233}%
\pgfsetfillcolor{currentfill}%
\pgfsetlinewidth{0.000000pt}%
\definecolor{currentstroke}{rgb}{0.000000,0.000000,0.000000}%
\pgfsetstrokecolor{currentstroke}%
\pgfsetstrokeopacity{0.000000}%
\pgfsetdash{}{0pt}%
\pgfpathmoveto{\pgfqpoint{-14.523748in}{1.066837in}}%
\pgfpathlineto{\pgfqpoint{6.850000in}{1.066837in}}%
\pgfpathlineto{\pgfqpoint{6.850000in}{0.747602in}}%
\pgfpathlineto{\pgfqpoint{-14.523748in}{0.747602in}}%
\pgfpathlineto{\pgfqpoint{-14.523748in}{1.066837in}}%
\pgfpathclose%
\pgfusepath{fill}%
\end{pgfscope}%
\begin{pgfscope}%
\pgfsetbuttcap%
\pgfsetroundjoin%
\definecolor{currentfill}{rgb}{0.000000,0.000000,0.000000}%
\pgfsetfillcolor{currentfill}%
\pgfsetlinewidth{0.803000pt}%
\definecolor{currentstroke}{rgb}{0.000000,0.000000,0.000000}%
\pgfsetstrokecolor{currentstroke}%
\pgfsetdash{}{0pt}%
\pgfsys@defobject{currentmarker}{\pgfqpoint{0.000000in}{-0.048611in}}{\pgfqpoint{0.000000in}{0.000000in}}{%
\pgfpathmoveto{\pgfqpoint{0.000000in}{0.000000in}}%
\pgfpathlineto{\pgfqpoint{0.000000in}{-0.048611in}}%
\pgfusepath{stroke,fill}%
}%
\begin{pgfscope}%
\pgfsys@transformshift{2.510968in}{0.611927in}%
\pgfsys@useobject{currentmarker}{}%
\end{pgfscope}%
\end{pgfscope}%
\begin{pgfscope}%
\definecolor{textcolor}{rgb}{0.000000,0.000000,0.000000}%
\pgfsetstrokecolor{textcolor}%
\pgfsetfillcolor{textcolor}%
\pgftext[x=2.510968in,y=0.514705in,,top]{\color{textcolor}\sffamily\fontsize{11.000000}{13.200000}\selectfont 53}%
\end{pgfscope}%
\begin{pgfscope}%
\pgfsetbuttcap%
\pgfsetroundjoin%
\definecolor{currentfill}{rgb}{0.000000,0.000000,0.000000}%
\pgfsetfillcolor{currentfill}%
\pgfsetlinewidth{0.803000pt}%
\definecolor{currentstroke}{rgb}{0.000000,0.000000,0.000000}%
\pgfsetstrokecolor{currentstroke}%
\pgfsetdash{}{0pt}%
\pgfsys@defobject{currentmarker}{\pgfqpoint{0.000000in}{-0.048611in}}{\pgfqpoint{0.000000in}{0.000000in}}{%
\pgfpathmoveto{\pgfqpoint{0.000000in}{0.000000in}}%
\pgfpathlineto{\pgfqpoint{0.000000in}{-0.048611in}}%
\pgfusepath{stroke,fill}%
}%
\begin{pgfscope}%
\pgfsys@transformshift{3.796607in}{0.611927in}%
\pgfsys@useobject{currentmarker}{}%
\end{pgfscope}%
\end{pgfscope}%
\begin{pgfscope}%
\definecolor{textcolor}{rgb}{0.000000,0.000000,0.000000}%
\pgfsetstrokecolor{textcolor}%
\pgfsetfillcolor{textcolor}%
\pgftext[x=3.796607in,y=0.514705in,,top]{\color{textcolor}\sffamily\fontsize{11.000000}{13.200000}\selectfont 57}%
\end{pgfscope}%
\begin{pgfscope}%
\pgfsetbuttcap%
\pgfsetroundjoin%
\definecolor{currentfill}{rgb}{0.000000,0.000000,0.000000}%
\pgfsetfillcolor{currentfill}%
\pgfsetlinewidth{0.803000pt}%
\definecolor{currentstroke}{rgb}{0.000000,0.000000,0.000000}%
\pgfsetstrokecolor{currentstroke}%
\pgfsetdash{}{0pt}%
\pgfsys@defobject{currentmarker}{\pgfqpoint{0.000000in}{-0.048611in}}{\pgfqpoint{0.000000in}{0.000000in}}{%
\pgfpathmoveto{\pgfqpoint{0.000000in}{0.000000in}}%
\pgfpathlineto{\pgfqpoint{0.000000in}{-0.048611in}}%
\pgfusepath{stroke,fill}%
}%
\begin{pgfscope}%
\pgfsys@transformshift{4.760837in}{0.611927in}%
\pgfsys@useobject{currentmarker}{}%
\end{pgfscope}%
\end{pgfscope}%
\begin{pgfscope}%
\definecolor{textcolor}{rgb}{0.000000,0.000000,0.000000}%
\pgfsetstrokecolor{textcolor}%
\pgfsetfillcolor{textcolor}%
\pgftext[x=4.760837in,y=0.514705in,,top]{\color{textcolor}\sffamily\fontsize{11.000000}{13.200000}\selectfont 60}%
\end{pgfscope}%
\begin{pgfscope}%
\pgfsetbuttcap%
\pgfsetroundjoin%
\definecolor{currentfill}{rgb}{0.000000,0.000000,0.000000}%
\pgfsetfillcolor{currentfill}%
\pgfsetlinewidth{0.803000pt}%
\definecolor{currentstroke}{rgb}{0.000000,0.000000,0.000000}%
\pgfsetstrokecolor{currentstroke}%
\pgfsetdash{}{0pt}%
\pgfsys@defobject{currentmarker}{\pgfqpoint{0.000000in}{-0.048611in}}{\pgfqpoint{0.000000in}{0.000000in}}{%
\pgfpathmoveto{\pgfqpoint{0.000000in}{0.000000in}}%
\pgfpathlineto{\pgfqpoint{0.000000in}{-0.048611in}}%
\pgfusepath{stroke,fill}%
}%
\begin{pgfscope}%
\pgfsys@transformshift{5.725066in}{0.611927in}%
\pgfsys@useobject{currentmarker}{}%
\end{pgfscope}%
\end{pgfscope}%
\begin{pgfscope}%
\definecolor{textcolor}{rgb}{0.000000,0.000000,0.000000}%
\pgfsetstrokecolor{textcolor}%
\pgfsetfillcolor{textcolor}%
\pgftext[x=5.725066in,y=0.514705in,,top]{\color{textcolor}\sffamily\fontsize{11.000000}{13.200000}\selectfont 63}%
\end{pgfscope}%
\begin{pgfscope}%
\pgfsetbuttcap%
\pgfsetroundjoin%
\definecolor{currentfill}{rgb}{0.000000,0.000000,0.000000}%
\pgfsetfillcolor{currentfill}%
\pgfsetlinewidth{0.803000pt}%
\definecolor{currentstroke}{rgb}{0.000000,0.000000,0.000000}%
\pgfsetstrokecolor{currentstroke}%
\pgfsetdash{}{0pt}%
\pgfsys@defobject{currentmarker}{\pgfqpoint{0.000000in}{-0.048611in}}{\pgfqpoint{0.000000in}{0.000000in}}{%
\pgfpathmoveto{\pgfqpoint{0.000000in}{0.000000in}}%
\pgfpathlineto{\pgfqpoint{0.000000in}{-0.048611in}}%
\pgfusepath{stroke,fill}%
}%
\begin{pgfscope}%
\pgfsys@transformshift{6.689295in}{0.611927in}%
\pgfsys@useobject{currentmarker}{}%
\end{pgfscope}%
\end{pgfscope}%
\begin{pgfscope}%
\definecolor{textcolor}{rgb}{0.000000,0.000000,0.000000}%
\pgfsetstrokecolor{textcolor}%
\pgfsetfillcolor{textcolor}%
\pgftext[x=6.689295in,y=0.514705in,,top]{\color{textcolor}\sffamily\fontsize{11.000000}{13.200000}\selectfont 66}%
\end{pgfscope}%
\begin{pgfscope}%
\definecolor{textcolor}{rgb}{0.000000,0.000000,0.000000}%
\pgfsetstrokecolor{textcolor}%
\pgfsetfillcolor{textcolor}%
\pgftext[x=4.300417in,y=0.311295in,,top]{\color{textcolor}\sffamily\fontsize{12.000000}{14.400000}\selectfont Test Accuracy (\%)}%
\end{pgfscope}%
\begin{pgfscope}%
\pgfsetbuttcap%
\pgfsetroundjoin%
\definecolor{currentfill}{rgb}{0.000000,0.000000,0.000000}%
\pgfsetfillcolor{currentfill}%
\pgfsetlinewidth{0.803000pt}%
\definecolor{currentstroke}{rgb}{0.000000,0.000000,0.000000}%
\pgfsetstrokecolor{currentstroke}%
\pgfsetdash{}{0pt}%
\pgfsys@defobject{currentmarker}{\pgfqpoint{-0.048611in}{0.000000in}}{\pgfqpoint{-0.000000in}{0.000000in}}{%
\pgfpathmoveto{\pgfqpoint{-0.000000in}{0.000000in}}%
\pgfpathlineto{\pgfqpoint{-0.048611in}{0.000000in}}%
\pgfusepath{stroke,fill}%
}%
\begin{pgfscope}%
\pgfsys@transformshift{1.750834in}{3.301482in}%
\pgfsys@useobject{currentmarker}{}%
\end{pgfscope}%
\end{pgfscope}%
\begin{pgfscope}%
\definecolor{textcolor}{rgb}{0.000000,0.000000,0.000000}%
\pgfsetstrokecolor{textcolor}%
\pgfsetfillcolor{textcolor}%
\pgftext[x=0.783359in, y=3.227616in, left, base]{\color{textcolor}\sffamily\fontsize{14.000000}{16.800000}\selectfont wrn-16-1}%
\end{pgfscope}%
\begin{pgfscope}%
\pgfsetbuttcap%
\pgfsetroundjoin%
\definecolor{currentfill}{rgb}{0.000000,0.000000,0.000000}%
\pgfsetfillcolor{currentfill}%
\pgfsetlinewidth{0.803000pt}%
\definecolor{currentstroke}{rgb}{0.000000,0.000000,0.000000}%
\pgfsetstrokecolor{currentstroke}%
\pgfsetdash{}{0pt}%
\pgfsys@defobject{currentmarker}{\pgfqpoint{-0.048611in}{0.000000in}}{\pgfqpoint{-0.000000in}{0.000000in}}{%
\pgfpathmoveto{\pgfqpoint{-0.000000in}{0.000000in}}%
\pgfpathlineto{\pgfqpoint{-0.048611in}{0.000000in}}%
\pgfusepath{stroke,fill}%
}%
\begin{pgfscope}%
\pgfsys@transformshift{1.750834in}{2.902438in}%
\pgfsys@useobject{currentmarker}{}%
\end{pgfscope}%
\end{pgfscope}%
\begin{pgfscope}%
\definecolor{textcolor}{rgb}{0.000000,0.000000,0.000000}%
\pgfsetstrokecolor{textcolor}%
\pgfsetfillcolor{textcolor}%
\pgftext[x=0.356493in, y=2.922661in, left, base]{\color{textcolor}\sffamily\fontsize{14.000000}{16.800000}\selectfont HPs selection}%
\end{pgfscope}%
\begin{pgfscope}%
\definecolor{textcolor}{rgb}{0.000000,0.000000,0.000000}%
\pgfsetstrokecolor{textcolor}%
\pgfsetfillcolor{textcolor}%
\pgftext[x=0.319180in, y=2.704937in, left, base]{\color{textcolor}\sffamily\fontsize{14.000000}{16.800000}\selectfont w/out val. set}%
\end{pgfscope}%
\begin{pgfscope}%
\pgfsetbuttcap%
\pgfsetroundjoin%
\definecolor{currentfill}{rgb}{0.000000,0.000000,0.000000}%
\pgfsetfillcolor{currentfill}%
\pgfsetlinewidth{0.803000pt}%
\definecolor{currentstroke}{rgb}{0.000000,0.000000,0.000000}%
\pgfsetstrokecolor{currentstroke}%
\pgfsetdash{}{0pt}%
\pgfsys@defobject{currentmarker}{\pgfqpoint{-0.048611in}{0.000000in}}{\pgfqpoint{-0.000000in}{0.000000in}}{%
\pgfpathmoveto{\pgfqpoint{-0.000000in}{0.000000in}}%
\pgfpathlineto{\pgfqpoint{-0.048611in}{0.000000in}}%
\pgfusepath{stroke,fill}%
}%
\begin{pgfscope}%
\pgfsys@transformshift{1.750834in}{2.503395in}%
\pgfsys@useobject{currentmarker}{}%
\end{pgfscope}%
\end{pgfscope}%
\begin{pgfscope}%
\definecolor{textcolor}{rgb}{0.000000,0.000000,0.000000}%
\pgfsetstrokecolor{textcolor}%
\pgfsetfillcolor{textcolor}%
\pgftext[x=0.197842in, y=2.429528in, left, base]{\color{textcolor}\sffamily\fontsize{14.000000}{16.800000}\selectfont No momentum}%
\end{pgfscope}%
\begin{pgfscope}%
\pgfsetbuttcap%
\pgfsetroundjoin%
\definecolor{currentfill}{rgb}{0.000000,0.000000,0.000000}%
\pgfsetfillcolor{currentfill}%
\pgfsetlinewidth{0.803000pt}%
\definecolor{currentstroke}{rgb}{0.000000,0.000000,0.000000}%
\pgfsetstrokecolor{currentstroke}%
\pgfsetdash{}{0pt}%
\pgfsys@defobject{currentmarker}{\pgfqpoint{-0.048611in}{0.000000in}}{\pgfqpoint{-0.000000in}{0.000000in}}{%
\pgfpathmoveto{\pgfqpoint{-0.000000in}{0.000000in}}%
\pgfpathlineto{\pgfqpoint{-0.048611in}{0.000000in}}%
\pgfusepath{stroke,fill}%
}%
\begin{pgfscope}%
\pgfsys@transformshift{1.750834in}{2.104351in}%
\pgfsys@useobject{currentmarker}{}%
\end{pgfscope}%
\end{pgfscope}%
\begin{pgfscope}%
\definecolor{textcolor}{rgb}{0.000000,0.000000,0.000000}%
\pgfsetstrokecolor{textcolor}%
\pgfsetfillcolor{textcolor}%
\pgftext[x=0.802376in, y=2.030485in, left, base]{\color{textcolor}\sffamily\fontsize{14.000000}{16.800000}\selectfont \(\displaystyle 3\times\) width}%
\end{pgfscope}%
\begin{pgfscope}%
\pgfsetbuttcap%
\pgfsetroundjoin%
\definecolor{currentfill}{rgb}{0.000000,0.000000,0.000000}%
\pgfsetfillcolor{currentfill}%
\pgfsetlinewidth{0.803000pt}%
\definecolor{currentstroke}{rgb}{0.000000,0.000000,0.000000}%
\pgfsetstrokecolor{currentstroke}%
\pgfsetdash{}{0pt}%
\pgfsys@defobject{currentmarker}{\pgfqpoint{-0.048611in}{0.000000in}}{\pgfqpoint{-0.000000in}{0.000000in}}{%
\pgfpathmoveto{\pgfqpoint{-0.000000in}{0.000000in}}%
\pgfpathlineto{\pgfqpoint{-0.048611in}{0.000000in}}%
\pgfusepath{stroke,fill}%
}%
\begin{pgfscope}%
\pgfsys@transformshift{1.750834in}{1.705307in}%
\pgfsys@useobject{currentmarker}{}%
\end{pgfscope}%
\end{pgfscope}%
\begin{pgfscope}%
\definecolor{textcolor}{rgb}{0.000000,0.000000,0.000000}%
\pgfsetstrokecolor{textcolor}%
\pgfsetfillcolor{textcolor}%
\pgftext[x=0.704461in, y=1.631441in, left, base]{\color{textcolor}\sffamily\fontsize{14.000000}{16.800000}\selectfont \(\displaystyle 10\times\) width}%
\end{pgfscope}%
\begin{pgfscope}%
\pgfsetbuttcap%
\pgfsetroundjoin%
\definecolor{currentfill}{rgb}{0.000000,0.000000,0.000000}%
\pgfsetfillcolor{currentfill}%
\pgfsetlinewidth{0.803000pt}%
\definecolor{currentstroke}{rgb}{0.000000,0.000000,0.000000}%
\pgfsetstrokecolor{currentstroke}%
\pgfsetdash{}{0pt}%
\pgfsys@defobject{currentmarker}{\pgfqpoint{-0.048611in}{0.000000in}}{\pgfqpoint{-0.000000in}{0.000000in}}{%
\pgfpathmoveto{\pgfqpoint{-0.000000in}{0.000000in}}%
\pgfpathlineto{\pgfqpoint{-0.048611in}{0.000000in}}%
\pgfusepath{stroke,fill}%
}%
\begin{pgfscope}%
\pgfsys@transformshift{1.750834in}{1.306263in}%
\pgfsys@useobject{currentmarker}{}%
\end{pgfscope}%
\end{pgfscope}%
\begin{pgfscope}%
\definecolor{textcolor}{rgb}{0.000000,0.000000,0.000000}%
\pgfsetstrokecolor{textcolor}%
\pgfsetfillcolor{textcolor}%
\pgftext[x=0.150000in, y=1.232397in, left, base]{\color{textcolor}\sffamily\fontsize{14.000000}{16.800000}\selectfont \(\displaystyle 25k \rightarrow 75k\) steps}%
\end{pgfscope}%
\begin{pgfscope}%
\pgfsetbuttcap%
\pgfsetroundjoin%
\definecolor{currentfill}{rgb}{0.000000,0.000000,0.000000}%
\pgfsetfillcolor{currentfill}%
\pgfsetlinewidth{0.803000pt}%
\definecolor{currentstroke}{rgb}{0.000000,0.000000,0.000000}%
\pgfsetstrokecolor{currentstroke}%
\pgfsetdash{}{0pt}%
\pgfsys@defobject{currentmarker}{\pgfqpoint{-0.048611in}{0.000000in}}{\pgfqpoint{-0.000000in}{0.000000in}}{%
\pgfpathmoveto{\pgfqpoint{-0.000000in}{0.000000in}}%
\pgfpathlineto{\pgfqpoint{-0.048611in}{0.000000in}}%
\pgfusepath{stroke,fill}%
}%
\begin{pgfscope}%
\pgfsys@transformshift{1.750834in}{0.907220in}%
\pgfsys@useobject{currentmarker}{}%
\end{pgfscope}%
\end{pgfscope}%
\begin{pgfscope}%
\definecolor{textcolor}{rgb}{0.000000,0.000000,0.000000}%
\pgfsetstrokecolor{textcolor}%
\pgfsetfillcolor{textcolor}%
\pgftext[x=0.704461in, y=0.833353in, left, base]{\color{textcolor}\sffamily\fontsize{14.000000}{16.800000}\selectfont \(\displaystyle 22\times\) width}%
\end{pgfscope}%
\begin{pgfscope}%
\pgfsetrectcap%
\pgfsetmiterjoin%
\pgfsetlinewidth{1.505625pt}%
\definecolor{currentstroke}{rgb}{0.000000,0.000000,0.000000}%
\pgfsetstrokecolor{currentstroke}%
\pgfsetdash{}{0pt}%
\pgfpathmoveto{\pgfqpoint{1.750834in}{0.611927in}}%
\pgfpathlineto{\pgfqpoint{1.750834in}{3.596774in}}%
\pgfusepath{stroke}%
\end{pgfscope}%
\begin{pgfscope}%
\pgfsetrectcap%
\pgfsetmiterjoin%
\pgfsetlinewidth{1.505625pt}%
\definecolor{currentstroke}{rgb}{0.000000,0.000000,0.000000}%
\pgfsetstrokecolor{currentstroke}%
\pgfsetdash{}{0pt}%
\pgfpathmoveto{\pgfqpoint{1.750834in}{0.611927in}}%
\pgfpathlineto{\pgfqpoint{6.850000in}{0.611927in}}%
\pgfusepath{stroke}%
\end{pgfscope}%
\begin{pgfscope}%
\definecolor{textcolor}{rgb}{1.000000,1.000000,1.000000}%
\pgfsetstrokecolor{textcolor}%
\pgfsetfillcolor{textcolor}%
\pgftext[x=2.527039in,y=3.301482in,right,]{\color{textcolor}\sffamily\fontsize{14.000000}{16.800000}\selectfont 53.3}%
\end{pgfscope}%
\begin{pgfscope}%
\definecolor{textcolor}{rgb}{1.000000,1.000000,1.000000}%
\pgfsetstrokecolor{textcolor}%
\pgfsetfillcolor{textcolor}%
\pgftext[x=3.684114in,y=2.902438in,right,]{\color{textcolor}\sffamily\fontsize{14.000000}{16.800000}\selectfont 56.9}%
\end{pgfscope}%
\begin{pgfscope}%
\definecolor{textcolor}{rgb}{1.000000,1.000000,1.000000}%
\pgfsetstrokecolor{textcolor}%
\pgfsetfillcolor{textcolor}%
\pgftext[x=4.069806in,y=2.503395in,right,]{\color{textcolor}\sffamily\fontsize{14.000000}{16.800000}\selectfont 58.1}%
\end{pgfscope}%
\begin{pgfscope}%
\definecolor{textcolor}{rgb}{1.000000,1.000000,1.000000}%
\pgfsetstrokecolor{textcolor}%
\pgfsetfillcolor{textcolor}%
\pgftext[x=5.580431in,y=2.104351in,right,]{\color{textcolor}\sffamily\fontsize{14.000000}{16.800000}\selectfont 62.8}%
\end{pgfscope}%
\begin{pgfscope}%
\definecolor{textcolor}{rgb}{1.000000,1.000000,1.000000}%
\pgfsetstrokecolor{textcolor}%
\pgfsetfillcolor{textcolor}%
\pgftext[x=6.351815in,y=1.705307in,right,]{\color{textcolor}\sffamily\fontsize{14.000000}{16.800000}\selectfont 65.2}%
\end{pgfscope}%
\begin{pgfscope}%
\definecolor{textcolor}{rgb}{1.000000,1.000000,1.000000}%
\pgfsetstrokecolor{textcolor}%
\pgfsetfillcolor{textcolor}%
\pgftext[x=6.576802in,y=1.306263in,right,]{\color{textcolor}\sffamily\fontsize{14.000000}{16.800000}\selectfont 65.9}%
\end{pgfscope}%
\begin{pgfscope}%
\definecolor{textcolor}{rgb}{1.000000,1.000000,1.000000}%
\pgfsetstrokecolor{textcolor}%
\pgfsetfillcolor{textcolor}%
\pgftext[x=6.769648in,y=0.907220in,right,]{\color{textcolor}\sffamily\fontsize{14.000000}{16.800000}\selectfont 66.5}%
\end{pgfscope}%
\begin{pgfscope}%
\definecolor{textcolor}{rgb}{0.000000,0.000000,0.000000}%
\pgfsetstrokecolor{textcolor}%
\pgfsetfillcolor{textcolor}%
\pgftext[x=4.300417in,y=3.680108in,,base]{\color{textcolor}\sffamily\fontsize{16.100000}{19.320000}\selectfont ciFAIR-10 (1\% training set)}%
\end{pgfscope}%
\end{pgfpicture}%
\makeatother%
\endgroup%

%% file: figures/fig_correlation_wrn-16-1_mu-09.pgf
\begingroup%
\makeatletter%
\begin{pgfpicture}%
\pgfpathrectangle{\pgfpointorigin}{\pgfqpoint{5.000000in}{2.000000in}}%
\pgfusepath{use as bounding box, clip}%
\begin{pgfscope}%
\pgfsetbuttcap%
\pgfsetmiterjoin%
\definecolor{currentfill}{rgb}{1.000000,1.000000,1.000000}%
\pgfsetfillcolor{currentfill}%
\pgfsetlinewidth{0.000000pt}%
\definecolor{currentstroke}{rgb}{1.000000,1.000000,1.000000}%
\pgfsetstrokecolor{currentstroke}%
\pgfsetdash{}{0pt}%
\pgfpathmoveto{\pgfqpoint{0.000000in}{0.000000in}}%
\pgfpathlineto{\pgfqpoint{5.000000in}{0.000000in}}%
\pgfpathlineto{\pgfqpoint{5.000000in}{2.000000in}}%
\pgfpathlineto{\pgfqpoint{0.000000in}{2.000000in}}%
\pgfpathlineto{\pgfqpoint{0.000000in}{0.000000in}}%
\pgfpathclose%
\pgfusepath{fill}%
\end{pgfscope}%
\begin{pgfscope}%
\pgfsetbuttcap%
\pgfsetmiterjoin%
\definecolor{currentfill}{rgb}{1.000000,1.000000,1.000000}%
\pgfsetfillcolor{currentfill}%
\pgfsetlinewidth{0.000000pt}%
\definecolor{currentstroke}{rgb}{0.000000,0.000000,0.000000}%
\pgfsetstrokecolor{currentstroke}%
\pgfsetstrokeopacity{0.000000}%
\pgfsetdash{}{0pt}%
\pgfpathmoveto{\pgfqpoint{0.707041in}{0.395076in}}%
\pgfpathlineto{\pgfqpoint{4.021408in}{0.395076in}}%
\pgfpathlineto{\pgfqpoint{4.021408in}{1.640039in}}%
\pgfpathlineto{\pgfqpoint{0.707041in}{1.640039in}}%
\pgfpathlineto{\pgfqpoint{0.707041in}{0.395076in}}%
\pgfpathclose%
\pgfusepath{fill}%
\end{pgfscope}%
\begin{pgfscope}%
\pgfpathrectangle{\pgfqpoint{0.707041in}{0.395076in}}{\pgfqpoint{3.314368in}{1.244963in}}%
\pgfusepath{clip}%
\pgfsetbuttcap%
\pgfsetroundjoin%
\definecolor{currentfill}{rgb}{0.839216,0.380392,0.419608}%
\pgfsetfillcolor{currentfill}%
\pgfsetlinewidth{1.003750pt}%
\definecolor{currentstroke}{rgb}{0.839216,0.380392,0.419608}%
\pgfsetstrokecolor{currentstroke}%
\pgfsetdash{}{0pt}%
\pgfpathmoveto{\pgfqpoint{3.870755in}{1.145705in}}%
\pgfpathcurveto{\pgfqpoint{3.881805in}{1.145705in}}{\pgfqpoint{3.892404in}{1.150095in}}{\pgfqpoint{3.900218in}{1.157908in}}%
\pgfpathcurveto{\pgfqpoint{3.908031in}{1.165722in}}{\pgfqpoint{3.912422in}{1.176321in}}{\pgfqpoint{3.912422in}{1.187371in}}%
\pgfpathcurveto{\pgfqpoint{3.912422in}{1.198421in}}{\pgfqpoint{3.908031in}{1.209020in}}{\pgfqpoint{3.900218in}{1.216834in}}%
\pgfpathcurveto{\pgfqpoint{3.892404in}{1.224648in}}{\pgfqpoint{3.881805in}{1.229038in}}{\pgfqpoint{3.870755in}{1.229038in}}%
\pgfpathcurveto{\pgfqpoint{3.859705in}{1.229038in}}{\pgfqpoint{3.849106in}{1.224648in}}{\pgfqpoint{3.841292in}{1.216834in}}%
\pgfpathcurveto{\pgfqpoint{3.833479in}{1.209020in}}{\pgfqpoint{3.829088in}{1.198421in}}{\pgfqpoint{3.829088in}{1.187371in}}%
\pgfpathcurveto{\pgfqpoint{3.829088in}{1.176321in}}{\pgfqpoint{3.833479in}{1.165722in}}{\pgfqpoint{3.841292in}{1.157908in}}%
\pgfpathcurveto{\pgfqpoint{3.849106in}{1.150095in}}{\pgfqpoint{3.859705in}{1.145705in}}{\pgfqpoint{3.870755in}{1.145705in}}%
\pgfpathlineto{\pgfqpoint{3.870755in}{1.145705in}}%
\pgfpathclose%
\pgfusepath{stroke,fill}%
\end{pgfscope}%
\begin{pgfscope}%
\pgfpathrectangle{\pgfqpoint{0.707041in}{0.395076in}}{\pgfqpoint{3.314368in}{1.244963in}}%
\pgfusepath{clip}%
\pgfsetbuttcap%
\pgfsetroundjoin%
\definecolor{currentfill}{rgb}{0.870588,0.619608,0.839216}%
\pgfsetfillcolor{currentfill}%
\pgfsetlinewidth{1.003750pt}%
\definecolor{currentstroke}{rgb}{0.870588,0.619608,0.839216}%
\pgfsetstrokecolor{currentstroke}%
\pgfsetdash{}{0pt}%
\pgfpathmoveto{\pgfqpoint{3.164033in}{0.940235in}}%
\pgfpathcurveto{\pgfqpoint{3.175083in}{0.940235in}}{\pgfqpoint{3.185682in}{0.944626in}}{\pgfqpoint{3.193495in}{0.952439in}}%
\pgfpathcurveto{\pgfqpoint{3.201309in}{0.960253in}}{\pgfqpoint{3.205699in}{0.970852in}}{\pgfqpoint{3.205699in}{0.981902in}}%
\pgfpathcurveto{\pgfqpoint{3.205699in}{0.992952in}}{\pgfqpoint{3.201309in}{1.003551in}}{\pgfqpoint{3.193495in}{1.011365in}}%
\pgfpathcurveto{\pgfqpoint{3.185682in}{1.019178in}}{\pgfqpoint{3.175083in}{1.023569in}}{\pgfqpoint{3.164033in}{1.023569in}}%
\pgfpathcurveto{\pgfqpoint{3.152982in}{1.023569in}}{\pgfqpoint{3.142383in}{1.019178in}}{\pgfqpoint{3.134570in}{1.011365in}}%
\pgfpathcurveto{\pgfqpoint{3.126756in}{1.003551in}}{\pgfqpoint{3.122366in}{0.992952in}}{\pgfqpoint{3.122366in}{0.981902in}}%
\pgfpathcurveto{\pgfqpoint{3.122366in}{0.970852in}}{\pgfqpoint{3.126756in}{0.960253in}}{\pgfqpoint{3.134570in}{0.952439in}}%
\pgfpathcurveto{\pgfqpoint{3.142383in}{0.944626in}}{\pgfqpoint{3.152982in}{0.940235in}}{\pgfqpoint{3.164033in}{0.940235in}}%
\pgfpathlineto{\pgfqpoint{3.164033in}{0.940235in}}%
\pgfpathclose%
\pgfusepath{stroke,fill}%
\end{pgfscope}%
\begin{pgfscope}%
\pgfpathrectangle{\pgfqpoint{0.707041in}{0.395076in}}{\pgfqpoint{3.314368in}{1.244963in}}%
\pgfusepath{clip}%
\pgfsetbuttcap%
\pgfsetroundjoin%
\definecolor{currentfill}{rgb}{0.388235,0.474510,0.223529}%
\pgfsetfillcolor{currentfill}%
\pgfsetlinewidth{1.003750pt}%
\definecolor{currentstroke}{rgb}{0.388235,0.474510,0.223529}%
\pgfsetstrokecolor{currentstroke}%
\pgfsetdash{}{0pt}%
\pgfpathmoveto{\pgfqpoint{2.512476in}{1.421787in}}%
\pgfpathcurveto{\pgfqpoint{2.523526in}{1.421787in}}{\pgfqpoint{2.534125in}{1.426177in}}{\pgfqpoint{2.541939in}{1.433991in}}%
\pgfpathcurveto{\pgfqpoint{2.549752in}{1.441804in}}{\pgfqpoint{2.554142in}{1.452403in}}{\pgfqpoint{2.554142in}{1.463454in}}%
\pgfpathcurveto{\pgfqpoint{2.554142in}{1.474504in}}{\pgfqpoint{2.549752in}{1.485103in}}{\pgfqpoint{2.541939in}{1.492916in}}%
\pgfpathcurveto{\pgfqpoint{2.534125in}{1.500730in}}{\pgfqpoint{2.523526in}{1.505120in}}{\pgfqpoint{2.512476in}{1.505120in}}%
\pgfpathcurveto{\pgfqpoint{2.501426in}{1.505120in}}{\pgfqpoint{2.490827in}{1.500730in}}{\pgfqpoint{2.483013in}{1.492916in}}%
\pgfpathcurveto{\pgfqpoint{2.475199in}{1.485103in}}{\pgfqpoint{2.470809in}{1.474504in}}{\pgfqpoint{2.470809in}{1.463454in}}%
\pgfpathcurveto{\pgfqpoint{2.470809in}{1.452403in}}{\pgfqpoint{2.475199in}{1.441804in}}{\pgfqpoint{2.483013in}{1.433991in}}%
\pgfpathcurveto{\pgfqpoint{2.490827in}{1.426177in}}{\pgfqpoint{2.501426in}{1.421787in}}{\pgfqpoint{2.512476in}{1.421787in}}%
\pgfpathlineto{\pgfqpoint{2.512476in}{1.421787in}}%
\pgfpathclose%
\pgfusepath{stroke,fill}%
\end{pgfscope}%
\begin{pgfscope}%
\pgfpathrectangle{\pgfqpoint{0.707041in}{0.395076in}}{\pgfqpoint{3.314368in}{1.244963in}}%
\pgfusepath{clip}%
\pgfsetbuttcap%
\pgfsetroundjoin%
\definecolor{currentfill}{rgb}{0.741176,0.619608,0.223529}%
\pgfsetfillcolor{currentfill}%
\pgfsetlinewidth{1.003750pt}%
\definecolor{currentstroke}{rgb}{0.741176,0.619608,0.223529}%
\pgfsetstrokecolor{currentstroke}%
\pgfsetdash{}{0pt}%
\pgfpathmoveto{\pgfqpoint{2.633403in}{1.280412in}}%
\pgfpathcurveto{\pgfqpoint{2.644453in}{1.280412in}}{\pgfqpoint{2.655052in}{1.284802in}}{\pgfqpoint{2.662866in}{1.292616in}}%
\pgfpathcurveto{\pgfqpoint{2.670680in}{1.300429in}}{\pgfqpoint{2.675070in}{1.311028in}}{\pgfqpoint{2.675070in}{1.322079in}}%
\pgfpathcurveto{\pgfqpoint{2.675070in}{1.333129in}}{\pgfqpoint{2.670680in}{1.343728in}}{\pgfqpoint{2.662866in}{1.351541in}}%
\pgfpathcurveto{\pgfqpoint{2.655052in}{1.359355in}}{\pgfqpoint{2.644453in}{1.363745in}}{\pgfqpoint{2.633403in}{1.363745in}}%
\pgfpathcurveto{\pgfqpoint{2.622353in}{1.363745in}}{\pgfqpoint{2.611754in}{1.359355in}}{\pgfqpoint{2.603941in}{1.351541in}}%
\pgfpathcurveto{\pgfqpoint{2.596127in}{1.343728in}}{\pgfqpoint{2.591737in}{1.333129in}}{\pgfqpoint{2.591737in}{1.322079in}}%
\pgfpathcurveto{\pgfqpoint{2.591737in}{1.311028in}}{\pgfqpoint{2.596127in}{1.300429in}}{\pgfqpoint{2.603941in}{1.292616in}}%
\pgfpathcurveto{\pgfqpoint{2.611754in}{1.284802in}}{\pgfqpoint{2.622353in}{1.280412in}}{\pgfqpoint{2.633403in}{1.280412in}}%
\pgfpathlineto{\pgfqpoint{2.633403in}{1.280412in}}%
\pgfpathclose%
\pgfusepath{stroke,fill}%
\end{pgfscope}%
\begin{pgfscope}%
\pgfpathrectangle{\pgfqpoint{0.707041in}{0.395076in}}{\pgfqpoint{3.314368in}{1.244963in}}%
\pgfusepath{clip}%
\pgfsetbuttcap%
\pgfsetroundjoin%
\definecolor{currentfill}{rgb}{0.741176,0.619608,0.223529}%
\pgfsetfillcolor{currentfill}%
\pgfsetlinewidth{1.003750pt}%
\definecolor{currentstroke}{rgb}{0.741176,0.619608,0.223529}%
\pgfsetstrokecolor{currentstroke}%
\pgfsetdash{}{0pt}%
\pgfpathmoveto{\pgfqpoint{2.047710in}{1.001793in}}%
\pgfpathcurveto{\pgfqpoint{2.058760in}{1.001793in}}{\pgfqpoint{2.069359in}{1.006183in}}{\pgfqpoint{2.077173in}{1.013996in}}%
\pgfpathcurveto{\pgfqpoint{2.084986in}{1.021810in}}{\pgfqpoint{2.089377in}{1.032409in}}{\pgfqpoint{2.089377in}{1.043459in}}%
\pgfpathcurveto{\pgfqpoint{2.089377in}{1.054509in}}{\pgfqpoint{2.084986in}{1.065108in}}{\pgfqpoint{2.077173in}{1.072922in}}%
\pgfpathcurveto{\pgfqpoint{2.069359in}{1.080736in}}{\pgfqpoint{2.058760in}{1.085126in}}{\pgfqpoint{2.047710in}{1.085126in}}%
\pgfpathcurveto{\pgfqpoint{2.036660in}{1.085126in}}{\pgfqpoint{2.026061in}{1.080736in}}{\pgfqpoint{2.018247in}{1.072922in}}%
\pgfpathcurveto{\pgfqpoint{2.010434in}{1.065108in}}{\pgfqpoint{2.006043in}{1.054509in}}{\pgfqpoint{2.006043in}{1.043459in}}%
\pgfpathcurveto{\pgfqpoint{2.006043in}{1.032409in}}{\pgfqpoint{2.010434in}{1.021810in}}{\pgfqpoint{2.018247in}{1.013996in}}%
\pgfpathcurveto{\pgfqpoint{2.026061in}{1.006183in}}{\pgfqpoint{2.036660in}{1.001793in}}{\pgfqpoint{2.047710in}{1.001793in}}%
\pgfpathlineto{\pgfqpoint{2.047710in}{1.001793in}}%
\pgfpathclose%
\pgfusepath{stroke,fill}%
\end{pgfscope}%
\begin{pgfscope}%
\pgfpathrectangle{\pgfqpoint{0.707041in}{0.395076in}}{\pgfqpoint{3.314368in}{1.244963in}}%
\pgfusepath{clip}%
\pgfsetbuttcap%
\pgfsetroundjoin%
\definecolor{currentfill}{rgb}{0.839216,0.380392,0.419608}%
\pgfsetfillcolor{currentfill}%
\pgfsetlinewidth{1.003750pt}%
\definecolor{currentstroke}{rgb}{0.839216,0.380392,0.419608}%
\pgfsetstrokecolor{currentstroke}%
\pgfsetdash{}{0pt}%
\pgfpathmoveto{\pgfqpoint{1.894007in}{0.683157in}}%
\pgfpathcurveto{\pgfqpoint{1.905057in}{0.683157in}}{\pgfqpoint{1.915656in}{0.687547in}}{\pgfqpoint{1.923470in}{0.695361in}}%
\pgfpathcurveto{\pgfqpoint{1.931284in}{0.703175in}}{\pgfqpoint{1.935674in}{0.713774in}}{\pgfqpoint{1.935674in}{0.724824in}}%
\pgfpathcurveto{\pgfqpoint{1.935674in}{0.735874in}}{\pgfqpoint{1.931284in}{0.746473in}}{\pgfqpoint{1.923470in}{0.754287in}}%
\pgfpathcurveto{\pgfqpoint{1.915656in}{0.762100in}}{\pgfqpoint{1.905057in}{0.766490in}}{\pgfqpoint{1.894007in}{0.766490in}}%
\pgfpathcurveto{\pgfqpoint{1.882957in}{0.766490in}}{\pgfqpoint{1.872358in}{0.762100in}}{\pgfqpoint{1.864545in}{0.754287in}}%
\pgfpathcurveto{\pgfqpoint{1.856731in}{0.746473in}}{\pgfqpoint{1.852341in}{0.735874in}}{\pgfqpoint{1.852341in}{0.724824in}}%
\pgfpathcurveto{\pgfqpoint{1.852341in}{0.713774in}}{\pgfqpoint{1.856731in}{0.703175in}}{\pgfqpoint{1.864545in}{0.695361in}}%
\pgfpathcurveto{\pgfqpoint{1.872358in}{0.687547in}}{\pgfqpoint{1.882957in}{0.683157in}}{\pgfqpoint{1.894007in}{0.683157in}}%
\pgfpathlineto{\pgfqpoint{1.894007in}{0.683157in}}%
\pgfpathclose%
\pgfusepath{stroke,fill}%
\end{pgfscope}%
\begin{pgfscope}%
\pgfpathrectangle{\pgfqpoint{0.707041in}{0.395076in}}{\pgfqpoint{3.314368in}{1.244963in}}%
\pgfusepath{clip}%
\pgfsetbuttcap%
\pgfsetroundjoin%
\definecolor{currentfill}{rgb}{0.870588,0.619608,0.839216}%
\pgfsetfillcolor{currentfill}%
\pgfsetlinewidth{1.003750pt}%
\definecolor{currentstroke}{rgb}{0.870588,0.619608,0.839216}%
\pgfsetstrokecolor{currentstroke}%
\pgfsetdash{}{0pt}%
\pgfpathmoveto{\pgfqpoint{0.952267in}{0.446121in}}%
\pgfpathcurveto{\pgfqpoint{0.963317in}{0.446121in}}{\pgfqpoint{0.973916in}{0.450512in}}{\pgfqpoint{0.981730in}{0.458325in}}%
\pgfpathcurveto{\pgfqpoint{0.989544in}{0.466139in}}{\pgfqpoint{0.993934in}{0.476738in}}{\pgfqpoint{0.993934in}{0.487788in}}%
\pgfpathcurveto{\pgfqpoint{0.993934in}{0.498838in}}{\pgfqpoint{0.989544in}{0.509437in}}{\pgfqpoint{0.981730in}{0.517251in}}%
\pgfpathcurveto{\pgfqpoint{0.973916in}{0.525064in}}{\pgfqpoint{0.963317in}{0.529455in}}{\pgfqpoint{0.952267in}{0.529455in}}%
\pgfpathcurveto{\pgfqpoint{0.941217in}{0.529455in}}{\pgfqpoint{0.930618in}{0.525064in}}{\pgfqpoint{0.922804in}{0.517251in}}%
\pgfpathcurveto{\pgfqpoint{0.914991in}{0.509437in}}{\pgfqpoint{0.910600in}{0.498838in}}{\pgfqpoint{0.910600in}{0.487788in}}%
\pgfpathcurveto{\pgfqpoint{0.910600in}{0.476738in}}{\pgfqpoint{0.914991in}{0.466139in}}{\pgfqpoint{0.922804in}{0.458325in}}%
\pgfpathcurveto{\pgfqpoint{0.930618in}{0.450512in}}{\pgfqpoint{0.941217in}{0.446121in}}{\pgfqpoint{0.952267in}{0.446121in}}%
\pgfpathlineto{\pgfqpoint{0.952267in}{0.446121in}}%
\pgfpathclose%
\pgfusepath{stroke,fill}%
\end{pgfscope}%
\begin{pgfscope}%
\pgfpathrectangle{\pgfqpoint{0.707041in}{0.395076in}}{\pgfqpoint{3.314368in}{1.244963in}}%
\pgfusepath{clip}%
\pgfsetbuttcap%
\pgfsetroundjoin%
\definecolor{currentfill}{rgb}{0.223529,0.231373,0.474510}%
\pgfsetfillcolor{currentfill}%
\pgfsetlinewidth{1.003750pt}%
\definecolor{currentstroke}{rgb}{0.223529,0.231373,0.474510}%
\pgfsetstrokecolor{currentstroke}%
\pgfsetdash{}{0pt}%
\pgfpathmoveto{\pgfqpoint{2.162073in}{1.227545in}}%
\pgfpathcurveto{\pgfqpoint{2.173123in}{1.227545in}}{\pgfqpoint{2.183722in}{1.231935in}}{\pgfqpoint{2.191536in}{1.239749in}}%
\pgfpathcurveto{\pgfqpoint{2.199349in}{1.247562in}}{\pgfqpoint{2.203740in}{1.258161in}}{\pgfqpoint{2.203740in}{1.269211in}}%
\pgfpathcurveto{\pgfqpoint{2.203740in}{1.280262in}}{\pgfqpoint{2.199349in}{1.290861in}}{\pgfqpoint{2.191536in}{1.298674in}}%
\pgfpathcurveto{\pgfqpoint{2.183722in}{1.306488in}}{\pgfqpoint{2.173123in}{1.310878in}}{\pgfqpoint{2.162073in}{1.310878in}}%
\pgfpathcurveto{\pgfqpoint{2.151023in}{1.310878in}}{\pgfqpoint{2.140424in}{1.306488in}}{\pgfqpoint{2.132610in}{1.298674in}}%
\pgfpathcurveto{\pgfqpoint{2.124797in}{1.290861in}}{\pgfqpoint{2.120406in}{1.280262in}}{\pgfqpoint{2.120406in}{1.269211in}}%
\pgfpathcurveto{\pgfqpoint{2.120406in}{1.258161in}}{\pgfqpoint{2.124797in}{1.247562in}}{\pgfqpoint{2.132610in}{1.239749in}}%
\pgfpathcurveto{\pgfqpoint{2.140424in}{1.231935in}}{\pgfqpoint{2.151023in}{1.227545in}}{\pgfqpoint{2.162073in}{1.227545in}}%
\pgfpathlineto{\pgfqpoint{2.162073in}{1.227545in}}%
\pgfpathclose%
\pgfusepath{stroke,fill}%
\end{pgfscope}%
\begin{pgfscope}%
\pgfpathrectangle{\pgfqpoint{0.707041in}{0.395076in}}{\pgfqpoint{3.314368in}{1.244963in}}%
\pgfusepath{clip}%
\pgfsetbuttcap%
\pgfsetroundjoin%
\definecolor{currentfill}{rgb}{0.741176,0.619608,0.223529}%
\pgfsetfillcolor{currentfill}%
\pgfsetlinewidth{1.003750pt}%
\definecolor{currentstroke}{rgb}{0.741176,0.619608,0.223529}%
\pgfsetstrokecolor{currentstroke}%
\pgfsetdash{}{0pt}%
\pgfpathmoveto{\pgfqpoint{2.132964in}{0.836872in}}%
\pgfpathcurveto{\pgfqpoint{2.144014in}{0.836872in}}{\pgfqpoint{2.154613in}{0.841262in}}{\pgfqpoint{2.162427in}{0.849076in}}%
\pgfpathcurveto{\pgfqpoint{2.170240in}{0.856890in}}{\pgfqpoint{2.174631in}{0.867489in}}{\pgfqpoint{2.174631in}{0.878539in}}%
\pgfpathcurveto{\pgfqpoint{2.174631in}{0.889589in}}{\pgfqpoint{2.170240in}{0.900188in}}{\pgfqpoint{2.162427in}{0.908002in}}%
\pgfpathcurveto{\pgfqpoint{2.154613in}{0.915815in}}{\pgfqpoint{2.144014in}{0.920205in}}{\pgfqpoint{2.132964in}{0.920205in}}%
\pgfpathcurveto{\pgfqpoint{2.121914in}{0.920205in}}{\pgfqpoint{2.111315in}{0.915815in}}{\pgfqpoint{2.103501in}{0.908002in}}%
\pgfpathcurveto{\pgfqpoint{2.095688in}{0.900188in}}{\pgfqpoint{2.091297in}{0.889589in}}{\pgfqpoint{2.091297in}{0.878539in}}%
\pgfpathcurveto{\pgfqpoint{2.091297in}{0.867489in}}{\pgfqpoint{2.095688in}{0.856890in}}{\pgfqpoint{2.103501in}{0.849076in}}%
\pgfpathcurveto{\pgfqpoint{2.111315in}{0.841262in}}{\pgfqpoint{2.121914in}{0.836872in}}{\pgfqpoint{2.132964in}{0.836872in}}%
\pgfpathlineto{\pgfqpoint{2.132964in}{0.836872in}}%
\pgfpathclose%
\pgfusepath{stroke,fill}%
\end{pgfscope}%
\begin{pgfscope}%
\pgfpathrectangle{\pgfqpoint{0.707041in}{0.395076in}}{\pgfqpoint{3.314368in}{1.244963in}}%
\pgfusepath{clip}%
\pgfsetbuttcap%
\pgfsetroundjoin%
\definecolor{currentfill}{rgb}{0.839216,0.380392,0.419608}%
\pgfsetfillcolor{currentfill}%
\pgfsetlinewidth{1.003750pt}%
\definecolor{currentstroke}{rgb}{0.839216,0.380392,0.419608}%
\pgfsetstrokecolor{currentstroke}%
\pgfsetdash{}{0pt}%
\pgfpathmoveto{\pgfqpoint{1.809622in}{0.525354in}}%
\pgfpathcurveto{\pgfqpoint{1.820672in}{0.525354in}}{\pgfqpoint{1.831271in}{0.529744in}}{\pgfqpoint{1.839085in}{0.537558in}}%
\pgfpathcurveto{\pgfqpoint{1.846898in}{0.545372in}}{\pgfqpoint{1.851288in}{0.555971in}}{\pgfqpoint{1.851288in}{0.567021in}}%
\pgfpathcurveto{\pgfqpoint{1.851288in}{0.578071in}}{\pgfqpoint{1.846898in}{0.588670in}}{\pgfqpoint{1.839085in}{0.596483in}}%
\pgfpathcurveto{\pgfqpoint{1.831271in}{0.604297in}}{\pgfqpoint{1.820672in}{0.608687in}}{\pgfqpoint{1.809622in}{0.608687in}}%
\pgfpathcurveto{\pgfqpoint{1.798572in}{0.608687in}}{\pgfqpoint{1.787973in}{0.604297in}}{\pgfqpoint{1.780159in}{0.596483in}}%
\pgfpathcurveto{\pgfqpoint{1.772345in}{0.588670in}}{\pgfqpoint{1.767955in}{0.578071in}}{\pgfqpoint{1.767955in}{0.567021in}}%
\pgfpathcurveto{\pgfqpoint{1.767955in}{0.555971in}}{\pgfqpoint{1.772345in}{0.545372in}}{\pgfqpoint{1.780159in}{0.537558in}}%
\pgfpathcurveto{\pgfqpoint{1.787973in}{0.529744in}}{\pgfqpoint{1.798572in}{0.525354in}}{\pgfqpoint{1.809622in}{0.525354in}}%
\pgfpathlineto{\pgfqpoint{1.809622in}{0.525354in}}%
\pgfpathclose%
\pgfusepath{stroke,fill}%
\end{pgfscope}%
\begin{pgfscope}%
\pgfpathrectangle{\pgfqpoint{0.707041in}{0.395076in}}{\pgfqpoint{3.314368in}{1.244963in}}%
\pgfusepath{clip}%
\pgfsetbuttcap%
\pgfsetroundjoin%
\definecolor{currentfill}{rgb}{0.870588,0.619608,0.839216}%
\pgfsetfillcolor{currentfill}%
\pgfsetlinewidth{1.003750pt}%
\definecolor{currentstroke}{rgb}{0.870588,0.619608,0.839216}%
\pgfsetstrokecolor{currentstroke}%
\pgfsetdash{}{0pt}%
\pgfpathmoveto{\pgfqpoint{0.857694in}{0.430907in}}%
\pgfpathcurveto{\pgfqpoint{0.868744in}{0.430907in}}{\pgfqpoint{0.879343in}{0.435297in}}{\pgfqpoint{0.887156in}{0.443111in}}%
\pgfpathcurveto{\pgfqpoint{0.894970in}{0.450925in}}{\pgfqpoint{0.899360in}{0.461524in}}{\pgfqpoint{0.899360in}{0.472574in}}%
\pgfpathcurveto{\pgfqpoint{0.899360in}{0.483624in}}{\pgfqpoint{0.894970in}{0.494223in}}{\pgfqpoint{0.887156in}{0.502037in}}%
\pgfpathcurveto{\pgfqpoint{0.879343in}{0.509850in}}{\pgfqpoint{0.868744in}{0.514240in}}{\pgfqpoint{0.857694in}{0.514240in}}%
\pgfpathcurveto{\pgfqpoint{0.846643in}{0.514240in}}{\pgfqpoint{0.836044in}{0.509850in}}{\pgfqpoint{0.828231in}{0.502037in}}%
\pgfpathcurveto{\pgfqpoint{0.820417in}{0.494223in}}{\pgfqpoint{0.816027in}{0.483624in}}{\pgfqpoint{0.816027in}{0.472574in}}%
\pgfpathcurveto{\pgfqpoint{0.816027in}{0.461524in}}{\pgfqpoint{0.820417in}{0.450925in}}{\pgfqpoint{0.828231in}{0.443111in}}%
\pgfpathcurveto{\pgfqpoint{0.836044in}{0.435297in}}{\pgfqpoint{0.846643in}{0.430907in}}{\pgfqpoint{0.857694in}{0.430907in}}%
\pgfpathlineto{\pgfqpoint{0.857694in}{0.430907in}}%
\pgfpathclose%
\pgfusepath{stroke,fill}%
\end{pgfscope}%
\begin{pgfscope}%
\pgfpathrectangle{\pgfqpoint{0.707041in}{0.395076in}}{\pgfqpoint{3.314368in}{1.244963in}}%
\pgfusepath{clip}%
\pgfsetbuttcap%
\pgfsetroundjoin%
\definecolor{currentfill}{rgb}{0.839216,0.380392,0.419608}%
\pgfsetfillcolor{currentfill}%
\pgfsetlinewidth{1.003750pt}%
\definecolor{currentstroke}{rgb}{0.839216,0.380392,0.419608}%
\pgfsetstrokecolor{currentstroke}%
\pgfsetdash{}{0pt}%
\pgfpathmoveto{\pgfqpoint{1.888900in}{0.450014in}}%
\pgfpathcurveto{\pgfqpoint{1.899951in}{0.450014in}}{\pgfqpoint{1.910550in}{0.454405in}}{\pgfqpoint{1.918363in}{0.462218in}}%
\pgfpathcurveto{\pgfqpoint{1.926177in}{0.470032in}}{\pgfqpoint{1.930567in}{0.480631in}}{\pgfqpoint{1.930567in}{0.491681in}}%
\pgfpathcurveto{\pgfqpoint{1.930567in}{0.502731in}}{\pgfqpoint{1.926177in}{0.513330in}}{\pgfqpoint{1.918363in}{0.521144in}}%
\pgfpathcurveto{\pgfqpoint{1.910550in}{0.528957in}}{\pgfqpoint{1.899951in}{0.533348in}}{\pgfqpoint{1.888900in}{0.533348in}}%
\pgfpathcurveto{\pgfqpoint{1.877850in}{0.533348in}}{\pgfqpoint{1.867251in}{0.528957in}}{\pgfqpoint{1.859438in}{0.521144in}}%
\pgfpathcurveto{\pgfqpoint{1.851624in}{0.513330in}}{\pgfqpoint{1.847234in}{0.502731in}}{\pgfqpoint{1.847234in}{0.491681in}}%
\pgfpathcurveto{\pgfqpoint{1.847234in}{0.480631in}}{\pgfqpoint{1.851624in}{0.470032in}}{\pgfqpoint{1.859438in}{0.462218in}}%
\pgfpathcurveto{\pgfqpoint{1.867251in}{0.454405in}}{\pgfqpoint{1.877850in}{0.450014in}}{\pgfqpoint{1.888900in}{0.450014in}}%
\pgfpathlineto{\pgfqpoint{1.888900in}{0.450014in}}%
\pgfpathclose%
\pgfusepath{stroke,fill}%
\end{pgfscope}%
\begin{pgfscope}%
\pgfsetbuttcap%
\pgfsetroundjoin%
\definecolor{currentfill}{rgb}{0.000000,0.000000,0.000000}%
\pgfsetfillcolor{currentfill}%
\pgfsetlinewidth{0.803000pt}%
\definecolor{currentstroke}{rgb}{0.000000,0.000000,0.000000}%
\pgfsetstrokecolor{currentstroke}%
\pgfsetdash{}{0pt}%
\pgfsys@defobject{currentmarker}{\pgfqpoint{0.000000in}{-0.048611in}}{\pgfqpoint{0.000000in}{0.000000in}}{%
\pgfpathmoveto{\pgfqpoint{0.000000in}{0.000000in}}%
\pgfpathlineto{\pgfqpoint{0.000000in}{-0.048611in}}%
\pgfusepath{stroke,fill}%
}%
\begin{pgfscope}%
\pgfsys@transformshift{0.992898in}{0.395076in}%
\pgfsys@useobject{currentmarker}{}%
\end{pgfscope}%
\end{pgfscope}%
\begin{pgfscope}%
\definecolor{textcolor}{rgb}{0.000000,0.000000,0.000000}%
\pgfsetstrokecolor{textcolor}%
\pgfsetfillcolor{textcolor}%
\pgftext[x=0.992898in,y=0.297854in,,top]{\color{textcolor}\sffamily\fontsize{11.000000}{13.200000}\selectfont 36.0}%
\end{pgfscope}%
\begin{pgfscope}%
\pgfsetbuttcap%
\pgfsetroundjoin%
\definecolor{currentfill}{rgb}{0.000000,0.000000,0.000000}%
\pgfsetfillcolor{currentfill}%
\pgfsetlinewidth{0.803000pt}%
\definecolor{currentstroke}{rgb}{0.000000,0.000000,0.000000}%
\pgfsetstrokecolor{currentstroke}%
\pgfsetdash{}{0pt}%
\pgfsys@defobject{currentmarker}{\pgfqpoint{0.000000in}{-0.048611in}}{\pgfqpoint{0.000000in}{0.000000in}}{%
\pgfpathmoveto{\pgfqpoint{0.000000in}{0.000000in}}%
\pgfpathlineto{\pgfqpoint{0.000000in}{-0.048611in}}%
\pgfusepath{stroke,fill}%
}%
\begin{pgfscope}%
\pgfsys@transformshift{1.772313in}{0.395076in}%
\pgfsys@useobject{currentmarker}{}%
\end{pgfscope}%
\end{pgfscope}%
\begin{pgfscope}%
\definecolor{textcolor}{rgb}{0.000000,0.000000,0.000000}%
\pgfsetstrokecolor{textcolor}%
\pgfsetfillcolor{textcolor}%
\pgftext[x=1.772313in,y=0.297854in,,top]{\color{textcolor}\sffamily\fontsize{11.000000}{13.200000}\selectfont 36.6}%
\end{pgfscope}%
\begin{pgfscope}%
\pgfsetbuttcap%
\pgfsetroundjoin%
\definecolor{currentfill}{rgb}{0.000000,0.000000,0.000000}%
\pgfsetfillcolor{currentfill}%
\pgfsetlinewidth{0.803000pt}%
\definecolor{currentstroke}{rgb}{0.000000,0.000000,0.000000}%
\pgfsetstrokecolor{currentstroke}%
\pgfsetdash{}{0pt}%
\pgfsys@defobject{currentmarker}{\pgfqpoint{0.000000in}{-0.048611in}}{\pgfqpoint{0.000000in}{0.000000in}}{%
\pgfpathmoveto{\pgfqpoint{0.000000in}{0.000000in}}%
\pgfpathlineto{\pgfqpoint{0.000000in}{-0.048611in}}%
\pgfusepath{stroke,fill}%
}%
\begin{pgfscope}%
\pgfsys@transformshift{2.551729in}{0.395076in}%
\pgfsys@useobject{currentmarker}{}%
\end{pgfscope}%
\end{pgfscope}%
\begin{pgfscope}%
\definecolor{textcolor}{rgb}{0.000000,0.000000,0.000000}%
\pgfsetstrokecolor{textcolor}%
\pgfsetfillcolor{textcolor}%
\pgftext[x=2.551729in,y=0.297854in,,top]{\color{textcolor}\sffamily\fontsize{11.000000}{13.200000}\selectfont 37.2}%
\end{pgfscope}%
\begin{pgfscope}%
\pgfsetbuttcap%
\pgfsetroundjoin%
\definecolor{currentfill}{rgb}{0.000000,0.000000,0.000000}%
\pgfsetfillcolor{currentfill}%
\pgfsetlinewidth{0.803000pt}%
\definecolor{currentstroke}{rgb}{0.000000,0.000000,0.000000}%
\pgfsetstrokecolor{currentstroke}%
\pgfsetdash{}{0pt}%
\pgfsys@defobject{currentmarker}{\pgfqpoint{0.000000in}{-0.048611in}}{\pgfqpoint{0.000000in}{0.000000in}}{%
\pgfpathmoveto{\pgfqpoint{0.000000in}{0.000000in}}%
\pgfpathlineto{\pgfqpoint{0.000000in}{-0.048611in}}%
\pgfusepath{stroke,fill}%
}%
\begin{pgfscope}%
\pgfsys@transformshift{3.331144in}{0.395076in}%
\pgfsys@useobject{currentmarker}{}%
\end{pgfscope}%
\end{pgfscope}%
\begin{pgfscope}%
\definecolor{textcolor}{rgb}{0.000000,0.000000,0.000000}%
\pgfsetstrokecolor{textcolor}%
\pgfsetfillcolor{textcolor}%
\pgftext[x=3.331144in,y=0.297854in,,top]{\color{textcolor}\sffamily\fontsize{11.000000}{13.200000}\selectfont 37.8}%
\end{pgfscope}%
\begin{pgfscope}%
\pgfsetbuttcap%
\pgfsetroundjoin%
\definecolor{currentfill}{rgb}{0.000000,0.000000,0.000000}%
\pgfsetfillcolor{currentfill}%
\pgfsetlinewidth{0.803000pt}%
\definecolor{currentstroke}{rgb}{0.000000,0.000000,0.000000}%
\pgfsetstrokecolor{currentstroke}%
\pgfsetdash{}{0pt}%
\pgfsys@defobject{currentmarker}{\pgfqpoint{-0.048611in}{0.000000in}}{\pgfqpoint{-0.000000in}{0.000000in}}{%
\pgfpathmoveto{\pgfqpoint{-0.000000in}{0.000000in}}%
\pgfpathlineto{\pgfqpoint{-0.048611in}{0.000000in}}%
\pgfusepath{stroke,fill}%
}%
\begin{pgfscope}%
\pgfsys@transformshift{0.707041in}{0.577951in}%
\pgfsys@useobject{currentmarker}{}%
\end{pgfscope}%
\end{pgfscope}%
\begin{pgfscope}%
\definecolor{textcolor}{rgb}{0.000000,0.000000,0.000000}%
\pgfsetstrokecolor{textcolor}%
\pgfsetfillcolor{textcolor}%
\pgftext[x=0.366851in, y=0.519913in, left, base]{\color{textcolor}\sffamily\fontsize{11.000000}{13.200000}\selectfont 1.5}%
\end{pgfscope}%
\begin{pgfscope}%
\pgfsetbuttcap%
\pgfsetroundjoin%
\definecolor{currentfill}{rgb}{0.000000,0.000000,0.000000}%
\pgfsetfillcolor{currentfill}%
\pgfsetlinewidth{0.803000pt}%
\definecolor{currentstroke}{rgb}{0.000000,0.000000,0.000000}%
\pgfsetstrokecolor{currentstroke}%
\pgfsetdash{}{0pt}%
\pgfsys@defobject{currentmarker}{\pgfqpoint{-0.048611in}{0.000000in}}{\pgfqpoint{-0.000000in}{0.000000in}}{%
\pgfpathmoveto{\pgfqpoint{-0.000000in}{0.000000in}}%
\pgfpathlineto{\pgfqpoint{-0.048611in}{0.000000in}}%
\pgfusepath{stroke,fill}%
}%
\begin{pgfscope}%
\pgfsys@transformshift{0.707041in}{0.871401in}%
\pgfsys@useobject{currentmarker}{}%
\end{pgfscope}%
\end{pgfscope}%
\begin{pgfscope}%
\definecolor{textcolor}{rgb}{0.000000,0.000000,0.000000}%
\pgfsetstrokecolor{textcolor}%
\pgfsetfillcolor{textcolor}%
\pgftext[x=0.366851in, y=0.813364in, left, base]{\color{textcolor}\sffamily\fontsize{11.000000}{13.200000}\selectfont 2.0}%
\end{pgfscope}%
\begin{pgfscope}%
\pgfsetbuttcap%
\pgfsetroundjoin%
\definecolor{currentfill}{rgb}{0.000000,0.000000,0.000000}%
\pgfsetfillcolor{currentfill}%
\pgfsetlinewidth{0.803000pt}%
\definecolor{currentstroke}{rgb}{0.000000,0.000000,0.000000}%
\pgfsetstrokecolor{currentstroke}%
\pgfsetdash{}{0pt}%
\pgfsys@defobject{currentmarker}{\pgfqpoint{-0.048611in}{0.000000in}}{\pgfqpoint{-0.000000in}{0.000000in}}{%
\pgfpathmoveto{\pgfqpoint{-0.000000in}{0.000000in}}%
\pgfpathlineto{\pgfqpoint{-0.048611in}{0.000000in}}%
\pgfusepath{stroke,fill}%
}%
\begin{pgfscope}%
\pgfsys@transformshift{0.707041in}{1.164852in}%
\pgfsys@useobject{currentmarker}{}%
\end{pgfscope}%
\end{pgfscope}%
\begin{pgfscope}%
\definecolor{textcolor}{rgb}{0.000000,0.000000,0.000000}%
\pgfsetstrokecolor{textcolor}%
\pgfsetfillcolor{textcolor}%
\pgftext[x=0.366851in, y=1.106814in, left, base]{\color{textcolor}\sffamily\fontsize{11.000000}{13.200000}\selectfont 2.5}%
\end{pgfscope}%
\begin{pgfscope}%
\pgfsetbuttcap%
\pgfsetroundjoin%
\definecolor{currentfill}{rgb}{0.000000,0.000000,0.000000}%
\pgfsetfillcolor{currentfill}%
\pgfsetlinewidth{0.803000pt}%
\definecolor{currentstroke}{rgb}{0.000000,0.000000,0.000000}%
\pgfsetstrokecolor{currentstroke}%
\pgfsetdash{}{0pt}%
\pgfsys@defobject{currentmarker}{\pgfqpoint{-0.048611in}{0.000000in}}{\pgfqpoint{-0.000000in}{0.000000in}}{%
\pgfpathmoveto{\pgfqpoint{-0.000000in}{0.000000in}}%
\pgfpathlineto{\pgfqpoint{-0.048611in}{0.000000in}}%
\pgfusepath{stroke,fill}%
}%
\begin{pgfscope}%
\pgfsys@transformshift{0.707041in}{1.458303in}%
\pgfsys@useobject{currentmarker}{}%
\end{pgfscope}%
\end{pgfscope}%
\begin{pgfscope}%
\definecolor{textcolor}{rgb}{0.000000,0.000000,0.000000}%
\pgfsetstrokecolor{textcolor}%
\pgfsetfillcolor{textcolor}%
\pgftext[x=0.366851in, y=1.400265in, left, base]{\color{textcolor}\sffamily\fontsize{11.000000}{13.200000}\selectfont 3.0}%
\end{pgfscope}%
\begin{pgfscope}%
\definecolor{textcolor}{rgb}{0.000000,0.000000,0.000000}%
\pgfsetstrokecolor{textcolor}%
\pgfsetfillcolor{textcolor}%
\pgftext[x=0.311295in,y=1.017558in,,bottom,rotate=90.000000]{\color{textcolor}\sffamily\fontsize{12.000000}{14.400000}\selectfont Test Loss}%
\end{pgfscope}%
\begin{pgfscope}%
\pgfsetrectcap%
\pgfsetmiterjoin%
\pgfsetlinewidth{1.505625pt}%
\definecolor{currentstroke}{rgb}{0.000000,0.000000,0.000000}%
\pgfsetstrokecolor{currentstroke}%
\pgfsetdash{}{0pt}%
\pgfpathmoveto{\pgfqpoint{0.707041in}{0.395076in}}%
\pgfpathlineto{\pgfqpoint{0.707041in}{1.640039in}}%
\pgfusepath{stroke}%
\end{pgfscope}%
\begin{pgfscope}%
\pgfsetrectcap%
\pgfsetmiterjoin%
\pgfsetlinewidth{1.505625pt}%
\definecolor{currentstroke}{rgb}{0.000000,0.000000,0.000000}%
\pgfsetstrokecolor{currentstroke}%
\pgfsetdash{}{0pt}%
\pgfpathmoveto{\pgfqpoint{0.707041in}{0.395076in}}%
\pgfpathlineto{\pgfqpoint{4.021408in}{0.395076in}}%
\pgfusepath{stroke}%
\end{pgfscope}%
\begin{pgfscope}%
\definecolor{textcolor}{rgb}{0.000000,0.000000,0.000000}%
\pgfsetstrokecolor{textcolor}%
\pgfsetfillcolor{textcolor}%
\pgftext[x=0.872759in, y=1.500962in, left, base]{\color{textcolor}\sffamily\fontsize{12.000000}{14.400000}\selectfont \(\displaystyle \rho: 0.69\)}%
\end{pgfscope}%
\begin{pgfscope}%
\definecolor{textcolor}{rgb}{0.000000,0.000000,0.000000}%
\pgfsetstrokecolor{textcolor}%
\pgfsetfillcolor{textcolor}%
\pgftext[x=0.872759in, y=1.314341in, left, base]{\color{textcolor}\sffamily\fontsize{12.000000}{14.400000}\selectfont \(\displaystyle r: 0.84\)}%
\end{pgfscope}%
\begin{pgfscope}%
\definecolor{textcolor}{rgb}{0.000000,0.000000,0.000000}%
\pgfsetstrokecolor{textcolor}%
\pgfsetfillcolor{textcolor}%
\pgftext[x=2.364224in,y=1.723372in,,base]{\color{textcolor}\sffamily\fontsize{12.000000}{14.400000}\selectfont WRN-16-1, \(\displaystyle \mu=0.9\), \(\displaystyle T = 25k \)}%
\end{pgfscope}%
\begin{pgfscope}%
\pgfsetbuttcap%
\pgfsetmiterjoin%
\definecolor{currentfill}{rgb}{1.000000,1.000000,1.000000}%
\pgfsetfillcolor{currentfill}%
\pgfsetlinewidth{0.000000pt}%
\definecolor{currentstroke}{rgb}{0.000000,0.000000,0.000000}%
\pgfsetstrokecolor{currentstroke}%
\pgfsetstrokeopacity{0.000000}%
\pgfsetdash{}{0pt}%
\pgfpathmoveto{\pgfqpoint{4.228556in}{0.395076in}}%
\pgfpathlineto{\pgfqpoint{4.290804in}{0.395076in}}%
\pgfpathlineto{\pgfqpoint{4.290804in}{1.640039in}}%
\pgfpathlineto{\pgfqpoint{4.228556in}{1.640039in}}%
\pgfpathlineto{\pgfqpoint{4.228556in}{0.395076in}}%
\pgfpathclose%
\pgfusepath{fill}%
\end{pgfscope}%
\begin{pgfscope}%
\pgfpathrectangle{\pgfqpoint{4.228556in}{0.395076in}}{\pgfqpoint{0.062248in}{1.244963in}}%
\pgfusepath{clip}%
\pgfsetbuttcap%
\pgfsetmiterjoin%
\definecolor{currentfill}{rgb}{1.000000,1.000000,1.000000}%
\pgfsetfillcolor{currentfill}%
\pgfsetlinewidth{0.010037pt}%
\definecolor{currentstroke}{rgb}{1.000000,1.000000,1.000000}%
\pgfsetstrokecolor{currentstroke}%
\pgfsetdash{}{0pt}%
\pgfusepath{stroke,fill}%
\end{pgfscope}%
\begin{pgfscope}%
\pgfpathrectangle{\pgfqpoint{4.228556in}{0.395076in}}{\pgfqpoint{0.062248in}{1.244963in}}%
\pgfusepath{clip}%
\pgfsetbuttcap%
\pgfsetroundjoin%
\definecolor{currentfill}{rgb}{0.223529,0.231373,0.474510}%
\pgfsetfillcolor{currentfill}%
\pgfsetlinewidth{0.000000pt}%
\definecolor{currentstroke}{rgb}{0.000000,0.000000,0.000000}%
\pgfsetstrokecolor{currentstroke}%
\pgfsetdash{}{0pt}%
\pgfpathmoveto{\pgfqpoint{4.228556in}{0.395076in}}%
\pgfpathlineto{\pgfqpoint{4.290804in}{0.395076in}}%
\pgfpathlineto{\pgfqpoint{4.290804in}{0.457324in}}%
\pgfpathlineto{\pgfqpoint{4.228556in}{0.457324in}}%
\pgfpathlineto{\pgfqpoint{4.228556in}{0.395076in}}%
\pgfusepath{fill}%
\end{pgfscope}%
\begin{pgfscope}%
\pgfpathrectangle{\pgfqpoint{4.228556in}{0.395076in}}{\pgfqpoint{0.062248in}{1.244963in}}%
\pgfusepath{clip}%
\pgfsetbuttcap%
\pgfsetroundjoin%
\definecolor{currentfill}{rgb}{0.321569,0.329412,0.639216}%
\pgfsetfillcolor{currentfill}%
\pgfsetlinewidth{0.000000pt}%
\definecolor{currentstroke}{rgb}{0.000000,0.000000,0.000000}%
\pgfsetstrokecolor{currentstroke}%
\pgfsetdash{}{0pt}%
\pgfpathmoveto{\pgfqpoint{4.228556in}{0.457324in}}%
\pgfpathlineto{\pgfqpoint{4.290804in}{0.457324in}}%
\pgfpathlineto{\pgfqpoint{4.290804in}{0.519573in}}%
\pgfpathlineto{\pgfqpoint{4.228556in}{0.519573in}}%
\pgfpathlineto{\pgfqpoint{4.228556in}{0.457324in}}%
\pgfusepath{fill}%
\end{pgfscope}%
\begin{pgfscope}%
\pgfpathrectangle{\pgfqpoint{4.228556in}{0.395076in}}{\pgfqpoint{0.062248in}{1.244963in}}%
\pgfusepath{clip}%
\pgfsetbuttcap%
\pgfsetroundjoin%
\definecolor{currentfill}{rgb}{0.419608,0.431373,0.811765}%
\pgfsetfillcolor{currentfill}%
\pgfsetlinewidth{0.000000pt}%
\definecolor{currentstroke}{rgb}{0.000000,0.000000,0.000000}%
\pgfsetstrokecolor{currentstroke}%
\pgfsetdash{}{0pt}%
\pgfpathmoveto{\pgfqpoint{4.228556in}{0.519573in}}%
\pgfpathlineto{\pgfqpoint{4.290804in}{0.519573in}}%
\pgfpathlineto{\pgfqpoint{4.290804in}{0.581821in}}%
\pgfpathlineto{\pgfqpoint{4.228556in}{0.581821in}}%
\pgfpathlineto{\pgfqpoint{4.228556in}{0.519573in}}%
\pgfusepath{fill}%
\end{pgfscope}%
\begin{pgfscope}%
\pgfpathrectangle{\pgfqpoint{4.228556in}{0.395076in}}{\pgfqpoint{0.062248in}{1.244963in}}%
\pgfusepath{clip}%
\pgfsetbuttcap%
\pgfsetroundjoin%
\definecolor{currentfill}{rgb}{0.611765,0.619608,0.870588}%
\pgfsetfillcolor{currentfill}%
\pgfsetlinewidth{0.000000pt}%
\definecolor{currentstroke}{rgb}{0.000000,0.000000,0.000000}%
\pgfsetstrokecolor{currentstroke}%
\pgfsetdash{}{0pt}%
\pgfpathmoveto{\pgfqpoint{4.228556in}{0.581821in}}%
\pgfpathlineto{\pgfqpoint{4.290804in}{0.581821in}}%
\pgfpathlineto{\pgfqpoint{4.290804in}{0.644069in}}%
\pgfpathlineto{\pgfqpoint{4.228556in}{0.644069in}}%
\pgfpathlineto{\pgfqpoint{4.228556in}{0.581821in}}%
\pgfusepath{fill}%
\end{pgfscope}%
\begin{pgfscope}%
\pgfpathrectangle{\pgfqpoint{4.228556in}{0.395076in}}{\pgfqpoint{0.062248in}{1.244963in}}%
\pgfusepath{clip}%
\pgfsetbuttcap%
\pgfsetroundjoin%
\definecolor{currentfill}{rgb}{0.388235,0.474510,0.223529}%
\pgfsetfillcolor{currentfill}%
\pgfsetlinewidth{0.000000pt}%
\definecolor{currentstroke}{rgb}{0.000000,0.000000,0.000000}%
\pgfsetstrokecolor{currentstroke}%
\pgfsetdash{}{0pt}%
\pgfpathmoveto{\pgfqpoint{4.228556in}{0.644069in}}%
\pgfpathlineto{\pgfqpoint{4.290804in}{0.644069in}}%
\pgfpathlineto{\pgfqpoint{4.290804in}{0.706317in}}%
\pgfpathlineto{\pgfqpoint{4.228556in}{0.706317in}}%
\pgfpathlineto{\pgfqpoint{4.228556in}{0.644069in}}%
\pgfusepath{fill}%
\end{pgfscope}%
\begin{pgfscope}%
\pgfpathrectangle{\pgfqpoint{4.228556in}{0.395076in}}{\pgfqpoint{0.062248in}{1.244963in}}%
\pgfusepath{clip}%
\pgfsetbuttcap%
\pgfsetroundjoin%
\definecolor{currentfill}{rgb}{0.549020,0.635294,0.321569}%
\pgfsetfillcolor{currentfill}%
\pgfsetlinewidth{0.000000pt}%
\definecolor{currentstroke}{rgb}{0.000000,0.000000,0.000000}%
\pgfsetstrokecolor{currentstroke}%
\pgfsetdash{}{0pt}%
\pgfpathmoveto{\pgfqpoint{4.228556in}{0.706317in}}%
\pgfpathlineto{\pgfqpoint{4.290804in}{0.706317in}}%
\pgfpathlineto{\pgfqpoint{4.290804in}{0.768565in}}%
\pgfpathlineto{\pgfqpoint{4.228556in}{0.768565in}}%
\pgfpathlineto{\pgfqpoint{4.228556in}{0.706317in}}%
\pgfusepath{fill}%
\end{pgfscope}%
\begin{pgfscope}%
\pgfpathrectangle{\pgfqpoint{4.228556in}{0.395076in}}{\pgfqpoint{0.062248in}{1.244963in}}%
\pgfusepath{clip}%
\pgfsetbuttcap%
\pgfsetroundjoin%
\definecolor{currentfill}{rgb}{0.709804,0.811765,0.419608}%
\pgfsetfillcolor{currentfill}%
\pgfsetlinewidth{0.000000pt}%
\definecolor{currentstroke}{rgb}{0.000000,0.000000,0.000000}%
\pgfsetstrokecolor{currentstroke}%
\pgfsetdash{}{0pt}%
\pgfpathmoveto{\pgfqpoint{4.228556in}{0.768565in}}%
\pgfpathlineto{\pgfqpoint{4.290804in}{0.768565in}}%
\pgfpathlineto{\pgfqpoint{4.290804in}{0.830813in}}%
\pgfpathlineto{\pgfqpoint{4.228556in}{0.830813in}}%
\pgfpathlineto{\pgfqpoint{4.228556in}{0.768565in}}%
\pgfusepath{fill}%
\end{pgfscope}%
\begin{pgfscope}%
\pgfpathrectangle{\pgfqpoint{4.228556in}{0.395076in}}{\pgfqpoint{0.062248in}{1.244963in}}%
\pgfusepath{clip}%
\pgfsetbuttcap%
\pgfsetroundjoin%
\definecolor{currentfill}{rgb}{0.807843,0.858824,0.611765}%
\pgfsetfillcolor{currentfill}%
\pgfsetlinewidth{0.000000pt}%
\definecolor{currentstroke}{rgb}{0.000000,0.000000,0.000000}%
\pgfsetstrokecolor{currentstroke}%
\pgfsetdash{}{0pt}%
\pgfpathmoveto{\pgfqpoint{4.228556in}{0.830813in}}%
\pgfpathlineto{\pgfqpoint{4.290804in}{0.830813in}}%
\pgfpathlineto{\pgfqpoint{4.290804in}{0.893061in}}%
\pgfpathlineto{\pgfqpoint{4.228556in}{0.893061in}}%
\pgfpathlineto{\pgfqpoint{4.228556in}{0.830813in}}%
\pgfusepath{fill}%
\end{pgfscope}%
\begin{pgfscope}%
\pgfpathrectangle{\pgfqpoint{4.228556in}{0.395076in}}{\pgfqpoint{0.062248in}{1.244963in}}%
\pgfusepath{clip}%
\pgfsetbuttcap%
\pgfsetroundjoin%
\definecolor{currentfill}{rgb}{0.549020,0.427451,0.192157}%
\pgfsetfillcolor{currentfill}%
\pgfsetlinewidth{0.000000pt}%
\definecolor{currentstroke}{rgb}{0.000000,0.000000,0.000000}%
\pgfsetstrokecolor{currentstroke}%
\pgfsetdash{}{0pt}%
\pgfpathmoveto{\pgfqpoint{4.228556in}{0.893061in}}%
\pgfpathlineto{\pgfqpoint{4.290804in}{0.893061in}}%
\pgfpathlineto{\pgfqpoint{4.290804in}{0.955310in}}%
\pgfpathlineto{\pgfqpoint{4.228556in}{0.955310in}}%
\pgfpathlineto{\pgfqpoint{4.228556in}{0.893061in}}%
\pgfusepath{fill}%
\end{pgfscope}%
\begin{pgfscope}%
\pgfpathrectangle{\pgfqpoint{4.228556in}{0.395076in}}{\pgfqpoint{0.062248in}{1.244963in}}%
\pgfusepath{clip}%
\pgfsetbuttcap%
\pgfsetroundjoin%
\definecolor{currentfill}{rgb}{0.741176,0.619608,0.223529}%
\pgfsetfillcolor{currentfill}%
\pgfsetlinewidth{0.000000pt}%
\definecolor{currentstroke}{rgb}{0.000000,0.000000,0.000000}%
\pgfsetstrokecolor{currentstroke}%
\pgfsetdash{}{0pt}%
\pgfpathmoveto{\pgfqpoint{4.228556in}{0.955310in}}%
\pgfpathlineto{\pgfqpoint{4.290804in}{0.955310in}}%
\pgfpathlineto{\pgfqpoint{4.290804in}{1.017558in}}%
\pgfpathlineto{\pgfqpoint{4.228556in}{1.017558in}}%
\pgfpathlineto{\pgfqpoint{4.228556in}{0.955310in}}%
\pgfusepath{fill}%
\end{pgfscope}%
\begin{pgfscope}%
\pgfpathrectangle{\pgfqpoint{4.228556in}{0.395076in}}{\pgfqpoint{0.062248in}{1.244963in}}%
\pgfusepath{clip}%
\pgfsetbuttcap%
\pgfsetroundjoin%
\definecolor{currentfill}{rgb}{0.905882,0.729412,0.321569}%
\pgfsetfillcolor{currentfill}%
\pgfsetlinewidth{0.000000pt}%
\definecolor{currentstroke}{rgb}{0.000000,0.000000,0.000000}%
\pgfsetstrokecolor{currentstroke}%
\pgfsetdash{}{0pt}%
\pgfpathmoveto{\pgfqpoint{4.228556in}{1.017558in}}%
\pgfpathlineto{\pgfqpoint{4.290804in}{1.017558in}}%
\pgfpathlineto{\pgfqpoint{4.290804in}{1.079806in}}%
\pgfpathlineto{\pgfqpoint{4.228556in}{1.079806in}}%
\pgfpathlineto{\pgfqpoint{4.228556in}{1.017558in}}%
\pgfusepath{fill}%
\end{pgfscope}%
\begin{pgfscope}%
\pgfpathrectangle{\pgfqpoint{4.228556in}{0.395076in}}{\pgfqpoint{0.062248in}{1.244963in}}%
\pgfusepath{clip}%
\pgfsetbuttcap%
\pgfsetroundjoin%
\definecolor{currentfill}{rgb}{0.905882,0.796078,0.580392}%
\pgfsetfillcolor{currentfill}%
\pgfsetlinewidth{0.000000pt}%
\definecolor{currentstroke}{rgb}{0.000000,0.000000,0.000000}%
\pgfsetstrokecolor{currentstroke}%
\pgfsetdash{}{0pt}%
\pgfpathmoveto{\pgfqpoint{4.228556in}{1.079806in}}%
\pgfpathlineto{\pgfqpoint{4.290804in}{1.079806in}}%
\pgfpathlineto{\pgfqpoint{4.290804in}{1.142054in}}%
\pgfpathlineto{\pgfqpoint{4.228556in}{1.142054in}}%
\pgfpathlineto{\pgfqpoint{4.228556in}{1.079806in}}%
\pgfusepath{fill}%
\end{pgfscope}%
\begin{pgfscope}%
\pgfpathrectangle{\pgfqpoint{4.228556in}{0.395076in}}{\pgfqpoint{0.062248in}{1.244963in}}%
\pgfusepath{clip}%
\pgfsetbuttcap%
\pgfsetroundjoin%
\definecolor{currentfill}{rgb}{0.517647,0.235294,0.223529}%
\pgfsetfillcolor{currentfill}%
\pgfsetlinewidth{0.000000pt}%
\definecolor{currentstroke}{rgb}{0.000000,0.000000,0.000000}%
\pgfsetstrokecolor{currentstroke}%
\pgfsetdash{}{0pt}%
\pgfpathmoveto{\pgfqpoint{4.228556in}{1.142054in}}%
\pgfpathlineto{\pgfqpoint{4.290804in}{1.142054in}}%
\pgfpathlineto{\pgfqpoint{4.290804in}{1.204302in}}%
\pgfpathlineto{\pgfqpoint{4.228556in}{1.204302in}}%
\pgfpathlineto{\pgfqpoint{4.228556in}{1.142054in}}%
\pgfusepath{fill}%
\end{pgfscope}%
\begin{pgfscope}%
\pgfpathrectangle{\pgfqpoint{4.228556in}{0.395076in}}{\pgfqpoint{0.062248in}{1.244963in}}%
\pgfusepath{clip}%
\pgfsetbuttcap%
\pgfsetroundjoin%
\definecolor{currentfill}{rgb}{0.678431,0.286275,0.290196}%
\pgfsetfillcolor{currentfill}%
\pgfsetlinewidth{0.000000pt}%
\definecolor{currentstroke}{rgb}{0.000000,0.000000,0.000000}%
\pgfsetstrokecolor{currentstroke}%
\pgfsetdash{}{0pt}%
\pgfpathmoveto{\pgfqpoint{4.228556in}{1.204302in}}%
\pgfpathlineto{\pgfqpoint{4.290804in}{1.204302in}}%
\pgfpathlineto{\pgfqpoint{4.290804in}{1.266550in}}%
\pgfpathlineto{\pgfqpoint{4.228556in}{1.266550in}}%
\pgfpathlineto{\pgfqpoint{4.228556in}{1.204302in}}%
\pgfusepath{fill}%
\end{pgfscope}%
\begin{pgfscope}%
\pgfpathrectangle{\pgfqpoint{4.228556in}{0.395076in}}{\pgfqpoint{0.062248in}{1.244963in}}%
\pgfusepath{clip}%
\pgfsetbuttcap%
\pgfsetroundjoin%
\definecolor{currentfill}{rgb}{0.839216,0.380392,0.419608}%
\pgfsetfillcolor{currentfill}%
\pgfsetlinewidth{0.000000pt}%
\definecolor{currentstroke}{rgb}{0.000000,0.000000,0.000000}%
\pgfsetstrokecolor{currentstroke}%
\pgfsetdash{}{0pt}%
\pgfpathmoveto{\pgfqpoint{4.228556in}{1.266550in}}%
\pgfpathlineto{\pgfqpoint{4.290804in}{1.266550in}}%
\pgfpathlineto{\pgfqpoint{4.290804in}{1.328798in}}%
\pgfpathlineto{\pgfqpoint{4.228556in}{1.328798in}}%
\pgfpathlineto{\pgfqpoint{4.228556in}{1.266550in}}%
\pgfusepath{fill}%
\end{pgfscope}%
\begin{pgfscope}%
\pgfpathrectangle{\pgfqpoint{4.228556in}{0.395076in}}{\pgfqpoint{0.062248in}{1.244963in}}%
\pgfusepath{clip}%
\pgfsetbuttcap%
\pgfsetroundjoin%
\definecolor{currentfill}{rgb}{0.905882,0.588235,0.611765}%
\pgfsetfillcolor{currentfill}%
\pgfsetlinewidth{0.000000pt}%
\definecolor{currentstroke}{rgb}{0.000000,0.000000,0.000000}%
\pgfsetstrokecolor{currentstroke}%
\pgfsetdash{}{0pt}%
\pgfpathmoveto{\pgfqpoint{4.228556in}{1.328798in}}%
\pgfpathlineto{\pgfqpoint{4.290804in}{1.328798in}}%
\pgfpathlineto{\pgfqpoint{4.290804in}{1.391046in}}%
\pgfpathlineto{\pgfqpoint{4.228556in}{1.391046in}}%
\pgfpathlineto{\pgfqpoint{4.228556in}{1.328798in}}%
\pgfusepath{fill}%
\end{pgfscope}%
\begin{pgfscope}%
\pgfpathrectangle{\pgfqpoint{4.228556in}{0.395076in}}{\pgfqpoint{0.062248in}{1.244963in}}%
\pgfusepath{clip}%
\pgfsetbuttcap%
\pgfsetroundjoin%
\definecolor{currentfill}{rgb}{0.482353,0.254902,0.450980}%
\pgfsetfillcolor{currentfill}%
\pgfsetlinewidth{0.000000pt}%
\definecolor{currentstroke}{rgb}{0.000000,0.000000,0.000000}%
\pgfsetstrokecolor{currentstroke}%
\pgfsetdash{}{0pt}%
\pgfpathmoveto{\pgfqpoint{4.228556in}{1.391046in}}%
\pgfpathlineto{\pgfqpoint{4.290804in}{1.391046in}}%
\pgfpathlineto{\pgfqpoint{4.290804in}{1.453295in}}%
\pgfpathlineto{\pgfqpoint{4.228556in}{1.453295in}}%
\pgfpathlineto{\pgfqpoint{4.228556in}{1.391046in}}%
\pgfusepath{fill}%
\end{pgfscope}%
\begin{pgfscope}%
\pgfpathrectangle{\pgfqpoint{4.228556in}{0.395076in}}{\pgfqpoint{0.062248in}{1.244963in}}%
\pgfusepath{clip}%
\pgfsetbuttcap%
\pgfsetroundjoin%
\definecolor{currentfill}{rgb}{0.647059,0.317647,0.580392}%
\pgfsetfillcolor{currentfill}%
\pgfsetlinewidth{0.000000pt}%
\definecolor{currentstroke}{rgb}{0.000000,0.000000,0.000000}%
\pgfsetstrokecolor{currentstroke}%
\pgfsetdash{}{0pt}%
\pgfpathmoveto{\pgfqpoint{4.228556in}{1.453295in}}%
\pgfpathlineto{\pgfqpoint{4.290804in}{1.453295in}}%
\pgfpathlineto{\pgfqpoint{4.290804in}{1.515543in}}%
\pgfpathlineto{\pgfqpoint{4.228556in}{1.515543in}}%
\pgfpathlineto{\pgfqpoint{4.228556in}{1.453295in}}%
\pgfusepath{fill}%
\end{pgfscope}%
\begin{pgfscope}%
\pgfpathrectangle{\pgfqpoint{4.228556in}{0.395076in}}{\pgfqpoint{0.062248in}{1.244963in}}%
\pgfusepath{clip}%
\pgfsetbuttcap%
\pgfsetroundjoin%
\definecolor{currentfill}{rgb}{0.807843,0.427451,0.741176}%
\pgfsetfillcolor{currentfill}%
\pgfsetlinewidth{0.000000pt}%
\definecolor{currentstroke}{rgb}{0.000000,0.000000,0.000000}%
\pgfsetstrokecolor{currentstroke}%
\pgfsetdash{}{0pt}%
\pgfpathmoveto{\pgfqpoint{4.228556in}{1.515543in}}%
\pgfpathlineto{\pgfqpoint{4.290804in}{1.515543in}}%
\pgfpathlineto{\pgfqpoint{4.290804in}{1.577791in}}%
\pgfpathlineto{\pgfqpoint{4.228556in}{1.577791in}}%
\pgfpathlineto{\pgfqpoint{4.228556in}{1.515543in}}%
\pgfusepath{fill}%
\end{pgfscope}%
\begin{pgfscope}%
\pgfpathrectangle{\pgfqpoint{4.228556in}{0.395076in}}{\pgfqpoint{0.062248in}{1.244963in}}%
\pgfusepath{clip}%
\pgfsetbuttcap%
\pgfsetroundjoin%
\definecolor{currentfill}{rgb}{0.870588,0.619608,0.839216}%
\pgfsetfillcolor{currentfill}%
\pgfsetlinewidth{0.000000pt}%
\definecolor{currentstroke}{rgb}{0.000000,0.000000,0.000000}%
\pgfsetstrokecolor{currentstroke}%
\pgfsetdash{}{0pt}%
\pgfpathmoveto{\pgfqpoint{4.228556in}{1.577791in}}%
\pgfpathlineto{\pgfqpoint{4.290804in}{1.577791in}}%
\pgfpathlineto{\pgfqpoint{4.290804in}{1.640039in}}%
\pgfpathlineto{\pgfqpoint{4.228556in}{1.640039in}}%
\pgfpathlineto{\pgfqpoint{4.228556in}{1.577791in}}%
\pgfusepath{fill}%
\end{pgfscope}%
\begin{pgfscope}%
\pgfsetbuttcap%
\pgfsetroundjoin%
\definecolor{currentfill}{rgb}{0.000000,0.000000,0.000000}%
\pgfsetfillcolor{currentfill}%
\pgfsetlinewidth{0.803000pt}%
\definecolor{currentstroke}{rgb}{0.000000,0.000000,0.000000}%
\pgfsetstrokecolor{currentstroke}%
\pgfsetdash{}{0pt}%
\pgfsys@defobject{currentmarker}{\pgfqpoint{0.000000in}{0.000000in}}{\pgfqpoint{0.048611in}{0.000000in}}{%
\pgfpathmoveto{\pgfqpoint{0.000000in}{0.000000in}}%
\pgfpathlineto{\pgfqpoint{0.048611in}{0.000000in}}%
\pgfusepath{stroke,fill}%
}%
\begin{pgfscope}%
\pgfsys@transformshift{4.290804in}{0.699979in}%
\pgfsys@useobject{currentmarker}{}%
\end{pgfscope}%
\end{pgfscope}%
\begin{pgfscope}%
\definecolor{textcolor}{rgb}{0.000000,0.000000,0.000000}%
\pgfsetstrokecolor{textcolor}%
\pgfsetfillcolor{textcolor}%
\pgftext[x=4.388026in, y=0.641941in, left, base]{\color{textcolor}\sffamily\fontsize{11.000000}{13.200000}\selectfont \ensuremath{-}5.5}%
\end{pgfscope}%
\begin{pgfscope}%
\pgfsetbuttcap%
\pgfsetroundjoin%
\definecolor{currentfill}{rgb}{0.000000,0.000000,0.000000}%
\pgfsetfillcolor{currentfill}%
\pgfsetlinewidth{0.803000pt}%
\definecolor{currentstroke}{rgb}{0.000000,0.000000,0.000000}%
\pgfsetstrokecolor{currentstroke}%
\pgfsetdash{}{0pt}%
\pgfsys@defobject{currentmarker}{\pgfqpoint{0.000000in}{0.000000in}}{\pgfqpoint{0.048611in}{0.000000in}}{%
\pgfpathmoveto{\pgfqpoint{0.000000in}{0.000000in}}%
\pgfpathlineto{\pgfqpoint{0.048611in}{0.000000in}}%
\pgfusepath{stroke,fill}%
}%
\begin{pgfscope}%
\pgfsys@transformshift{4.290804in}{1.050124in}%
\pgfsys@useobject{currentmarker}{}%
\end{pgfscope}%
\end{pgfscope}%
\begin{pgfscope}%
\definecolor{textcolor}{rgb}{0.000000,0.000000,0.000000}%
\pgfsetstrokecolor{textcolor}%
\pgfsetfillcolor{textcolor}%
\pgftext[x=4.388026in, y=0.992087in, left, base]{\color{textcolor}\sffamily\fontsize{11.000000}{13.200000}\selectfont \ensuremath{-}5.0}%
\end{pgfscope}%
\begin{pgfscope}%
\pgfsetbuttcap%
\pgfsetroundjoin%
\definecolor{currentfill}{rgb}{0.000000,0.000000,0.000000}%
\pgfsetfillcolor{currentfill}%
\pgfsetlinewidth{0.803000pt}%
\definecolor{currentstroke}{rgb}{0.000000,0.000000,0.000000}%
\pgfsetstrokecolor{currentstroke}%
\pgfsetdash{}{0pt}%
\pgfsys@defobject{currentmarker}{\pgfqpoint{0.000000in}{0.000000in}}{\pgfqpoint{0.048611in}{0.000000in}}{%
\pgfpathmoveto{\pgfqpoint{0.000000in}{0.000000in}}%
\pgfpathlineto{\pgfqpoint{0.048611in}{0.000000in}}%
\pgfusepath{stroke,fill}%
}%
\begin{pgfscope}%
\pgfsys@transformshift{4.290804in}{1.400270in}%
\pgfsys@useobject{currentmarker}{}%
\end{pgfscope}%
\end{pgfscope}%
\begin{pgfscope}%
\definecolor{textcolor}{rgb}{0.000000,0.000000,0.000000}%
\pgfsetstrokecolor{textcolor}%
\pgfsetfillcolor{textcolor}%
\pgftext[x=4.388026in, y=1.342232in, left, base]{\color{textcolor}\sffamily\fontsize{11.000000}{13.200000}\selectfont \ensuremath{-}4.5}%
\end{pgfscope}%
\begin{pgfscope}%
\pgfsetrectcap%
\pgfsetmiterjoin%
\pgfsetlinewidth{0.803000pt}%
\definecolor{currentstroke}{rgb}{0.000000,0.000000,0.000000}%
\pgfsetstrokecolor{currentstroke}%
\pgfsetdash{}{0pt}%
\pgfpathmoveto{\pgfqpoint{4.228556in}{0.395076in}}%
\pgfpathlineto{\pgfqpoint{4.259680in}{0.395076in}}%
\pgfpathlineto{\pgfqpoint{4.290804in}{0.395076in}}%
\pgfpathlineto{\pgfqpoint{4.290804in}{1.640039in}}%
\pgfpathlineto{\pgfqpoint{4.259680in}{1.640039in}}%
\pgfpathlineto{\pgfqpoint{4.228556in}{1.640039in}}%
\pgfpathlineto{\pgfqpoint{4.228556in}{0.395076in}}%
\pgfpathclose%
\pgfusepath{stroke}%
\end{pgfscope}%
\end{pgfpicture}%
\makeatother%
\endgroup%

%% file: figures/fig_correlation_wrn-16-1_mu-00.pgf
\begingroup%
\makeatletter%
\begin{pgfpicture}%
\pgfpathrectangle{\pgfpointorigin}{\pgfqpoint{5.000000in}{2.000000in}}%
\pgfusepath{use as bounding box, clip}%
\begin{pgfscope}%
\pgfsetbuttcap%
\pgfsetmiterjoin%
\definecolor{currentfill}{rgb}{1.000000,1.000000,1.000000}%
\pgfsetfillcolor{currentfill}%
\pgfsetlinewidth{0.000000pt}%
\definecolor{currentstroke}{rgb}{1.000000,1.000000,1.000000}%
\pgfsetstrokecolor{currentstroke}%
\pgfsetdash{}{0pt}%
\pgfpathmoveto{\pgfqpoint{0.000000in}{0.000000in}}%
\pgfpathlineto{\pgfqpoint{5.000000in}{0.000000in}}%
\pgfpathlineto{\pgfqpoint{5.000000in}{2.000000in}}%
\pgfpathlineto{\pgfqpoint{0.000000in}{2.000000in}}%
\pgfpathlineto{\pgfqpoint{0.000000in}{0.000000in}}%
\pgfpathclose%
\pgfusepath{fill}%
\end{pgfscope}%
\begin{pgfscope}%
\pgfsetbuttcap%
\pgfsetmiterjoin%
\definecolor{currentfill}{rgb}{1.000000,1.000000,1.000000}%
\pgfsetfillcolor{currentfill}%
\pgfsetlinewidth{0.000000pt}%
\definecolor{currentstroke}{rgb}{0.000000,0.000000,0.000000}%
\pgfsetstrokecolor{currentstroke}%
\pgfsetstrokeopacity{0.000000}%
\pgfsetdash{}{0pt}%
\pgfpathmoveto{\pgfqpoint{0.490190in}{0.395076in}}%
\pgfpathlineto{\pgfqpoint{3.954691in}{0.395076in}}%
\pgfpathlineto{\pgfqpoint{3.954691in}{1.640039in}}%
\pgfpathlineto{\pgfqpoint{0.490190in}{1.640039in}}%
\pgfpathlineto{\pgfqpoint{0.490190in}{0.395076in}}%
\pgfpathclose%
\pgfusepath{fill}%
\end{pgfscope}%
\begin{pgfscope}%
\pgfpathrectangle{\pgfqpoint{0.490190in}{0.395076in}}{\pgfqpoint{3.464501in}{1.244963in}}%
\pgfusepath{clip}%
\pgfsetbuttcap%
\pgfsetroundjoin%
\definecolor{currentfill}{rgb}{0.549020,0.427451,0.192157}%
\pgfsetfillcolor{currentfill}%
\pgfsetlinewidth{1.003750pt}%
\definecolor{currentstroke}{rgb}{0.549020,0.427451,0.192157}%
\pgfsetstrokecolor{currentstroke}%
\pgfsetdash{}{0pt}%
\pgfpathmoveto{\pgfqpoint{3.797214in}{1.423203in}}%
\pgfpathcurveto{\pgfqpoint{3.808264in}{1.423203in}}{\pgfqpoint{3.818863in}{1.427594in}}{\pgfqpoint{3.826676in}{1.435407in}}%
\pgfpathcurveto{\pgfqpoint{3.834490in}{1.443221in}}{\pgfqpoint{3.838880in}{1.453820in}}{\pgfqpoint{3.838880in}{1.464870in}}%
\pgfpathcurveto{\pgfqpoint{3.838880in}{1.475920in}}{\pgfqpoint{3.834490in}{1.486519in}}{\pgfqpoint{3.826676in}{1.494333in}}%
\pgfpathcurveto{\pgfqpoint{3.818863in}{1.502147in}}{\pgfqpoint{3.808264in}{1.506537in}}{\pgfqpoint{3.797214in}{1.506537in}}%
\pgfpathcurveto{\pgfqpoint{3.786163in}{1.506537in}}{\pgfqpoint{3.775564in}{1.502147in}}{\pgfqpoint{3.767751in}{1.494333in}}%
\pgfpathcurveto{\pgfqpoint{3.759937in}{1.486519in}}{\pgfqpoint{3.755547in}{1.475920in}}{\pgfqpoint{3.755547in}{1.464870in}}%
\pgfpathcurveto{\pgfqpoint{3.755547in}{1.453820in}}{\pgfqpoint{3.759937in}{1.443221in}}{\pgfqpoint{3.767751in}{1.435407in}}%
\pgfpathcurveto{\pgfqpoint{3.775564in}{1.427594in}}{\pgfqpoint{3.786163in}{1.423203in}}{\pgfqpoint{3.797214in}{1.423203in}}%
\pgfpathlineto{\pgfqpoint{3.797214in}{1.423203in}}%
\pgfpathclose%
\pgfusepath{stroke,fill}%
\end{pgfscope}%
\begin{pgfscope}%
\pgfpathrectangle{\pgfqpoint{0.490190in}{0.395076in}}{\pgfqpoint{3.464501in}{1.244963in}}%
\pgfusepath{clip}%
\pgfsetbuttcap%
\pgfsetroundjoin%
\definecolor{currentfill}{rgb}{0.517647,0.235294,0.223529}%
\pgfsetfillcolor{currentfill}%
\pgfsetlinewidth{1.003750pt}%
\definecolor{currentstroke}{rgb}{0.517647,0.235294,0.223529}%
\pgfsetstrokecolor{currentstroke}%
\pgfsetdash{}{0pt}%
\pgfpathmoveto{\pgfqpoint{3.549693in}{1.180745in}}%
\pgfpathcurveto{\pgfqpoint{3.560743in}{1.180745in}}{\pgfqpoint{3.571342in}{1.185135in}}{\pgfqpoint{3.579155in}{1.192949in}}%
\pgfpathcurveto{\pgfqpoint{3.586969in}{1.200762in}}{\pgfqpoint{3.591359in}{1.211361in}}{\pgfqpoint{3.591359in}{1.222411in}}%
\pgfpathcurveto{\pgfqpoint{3.591359in}{1.233462in}}{\pgfqpoint{3.586969in}{1.244061in}}{\pgfqpoint{3.579155in}{1.251874in}}%
\pgfpathcurveto{\pgfqpoint{3.571342in}{1.259688in}}{\pgfqpoint{3.560743in}{1.264078in}}{\pgfqpoint{3.549693in}{1.264078in}}%
\pgfpathcurveto{\pgfqpoint{3.538642in}{1.264078in}}{\pgfqpoint{3.528043in}{1.259688in}}{\pgfqpoint{3.520230in}{1.251874in}}%
\pgfpathcurveto{\pgfqpoint{3.512416in}{1.244061in}}{\pgfqpoint{3.508026in}{1.233462in}}{\pgfqpoint{3.508026in}{1.222411in}}%
\pgfpathcurveto{\pgfqpoint{3.508026in}{1.211361in}}{\pgfqpoint{3.512416in}{1.200762in}}{\pgfqpoint{3.520230in}{1.192949in}}%
\pgfpathcurveto{\pgfqpoint{3.528043in}{1.185135in}}{\pgfqpoint{3.538642in}{1.180745in}}{\pgfqpoint{3.549693in}{1.180745in}}%
\pgfpathlineto{\pgfqpoint{3.549693in}{1.180745in}}%
\pgfpathclose%
\pgfusepath{stroke,fill}%
\end{pgfscope}%
\begin{pgfscope}%
\pgfpathrectangle{\pgfqpoint{0.490190in}{0.395076in}}{\pgfqpoint{3.464501in}{1.244963in}}%
\pgfusepath{clip}%
\pgfsetbuttcap%
\pgfsetroundjoin%
\definecolor{currentfill}{rgb}{0.870588,0.619608,0.839216}%
\pgfsetfillcolor{currentfill}%
\pgfsetlinewidth{1.003750pt}%
\definecolor{currentstroke}{rgb}{0.870588,0.619608,0.839216}%
\pgfsetstrokecolor{currentstroke}%
\pgfsetdash{}{0pt}%
\pgfpathmoveto{\pgfqpoint{1.257461in}{0.624768in}}%
\pgfpathcurveto{\pgfqpoint{1.268511in}{0.624768in}}{\pgfqpoint{1.279110in}{0.629158in}}{\pgfqpoint{1.286924in}{0.636972in}}%
\pgfpathcurveto{\pgfqpoint{1.294737in}{0.644786in}}{\pgfqpoint{1.299127in}{0.655385in}}{\pgfqpoint{1.299127in}{0.666435in}}%
\pgfpathcurveto{\pgfqpoint{1.299127in}{0.677485in}}{\pgfqpoint{1.294737in}{0.688084in}}{\pgfqpoint{1.286924in}{0.695898in}}%
\pgfpathcurveto{\pgfqpoint{1.279110in}{0.703711in}}{\pgfqpoint{1.268511in}{0.708101in}}{\pgfqpoint{1.257461in}{0.708101in}}%
\pgfpathcurveto{\pgfqpoint{1.246411in}{0.708101in}}{\pgfqpoint{1.235812in}{0.703711in}}{\pgfqpoint{1.227998in}{0.695898in}}%
\pgfpathcurveto{\pgfqpoint{1.220184in}{0.688084in}}{\pgfqpoint{1.215794in}{0.677485in}}{\pgfqpoint{1.215794in}{0.666435in}}%
\pgfpathcurveto{\pgfqpoint{1.215794in}{0.655385in}}{\pgfqpoint{1.220184in}{0.644786in}}{\pgfqpoint{1.227998in}{0.636972in}}%
\pgfpathcurveto{\pgfqpoint{1.235812in}{0.629158in}}{\pgfqpoint{1.246411in}{0.624768in}}{\pgfqpoint{1.257461in}{0.624768in}}%
\pgfpathlineto{\pgfqpoint{1.257461in}{0.624768in}}%
\pgfpathclose%
\pgfusepath{stroke,fill}%
\end{pgfscope}%
\begin{pgfscope}%
\pgfpathrectangle{\pgfqpoint{0.490190in}{0.395076in}}{\pgfqpoint{3.464501in}{1.244963in}}%
\pgfusepath{clip}%
\pgfsetbuttcap%
\pgfsetroundjoin%
\definecolor{currentfill}{rgb}{0.388235,0.474510,0.223529}%
\pgfsetfillcolor{currentfill}%
\pgfsetlinewidth{1.003750pt}%
\definecolor{currentstroke}{rgb}{0.388235,0.474510,0.223529}%
\pgfsetstrokecolor{currentstroke}%
\pgfsetdash{}{0pt}%
\pgfpathmoveto{\pgfqpoint{3.301365in}{1.366304in}}%
\pgfpathcurveto{\pgfqpoint{3.312415in}{1.366304in}}{\pgfqpoint{3.323014in}{1.370694in}}{\pgfqpoint{3.330828in}{1.378508in}}%
\pgfpathcurveto{\pgfqpoint{3.338641in}{1.386322in}}{\pgfqpoint{3.343032in}{1.396921in}}{\pgfqpoint{3.343032in}{1.407971in}}%
\pgfpathcurveto{\pgfqpoint{3.343032in}{1.419021in}}{\pgfqpoint{3.338641in}{1.429620in}}{\pgfqpoint{3.330828in}{1.437434in}}%
\pgfpathcurveto{\pgfqpoint{3.323014in}{1.445247in}}{\pgfqpoint{3.312415in}{1.449637in}}{\pgfqpoint{3.301365in}{1.449637in}}%
\pgfpathcurveto{\pgfqpoint{3.290315in}{1.449637in}}{\pgfqpoint{3.279716in}{1.445247in}}{\pgfqpoint{3.271902in}{1.437434in}}%
\pgfpathcurveto{\pgfqpoint{3.264089in}{1.429620in}}{\pgfqpoint{3.259698in}{1.419021in}}{\pgfqpoint{3.259698in}{1.407971in}}%
\pgfpathcurveto{\pgfqpoint{3.259698in}{1.396921in}}{\pgfqpoint{3.264089in}{1.386322in}}{\pgfqpoint{3.271902in}{1.378508in}}%
\pgfpathcurveto{\pgfqpoint{3.279716in}{1.370694in}}{\pgfqpoint{3.290315in}{1.366304in}}{\pgfqpoint{3.301365in}{1.366304in}}%
\pgfpathlineto{\pgfqpoint{3.301365in}{1.366304in}}%
\pgfpathclose%
\pgfusepath{stroke,fill}%
\end{pgfscope}%
\begin{pgfscope}%
\pgfpathrectangle{\pgfqpoint{0.490190in}{0.395076in}}{\pgfqpoint{3.464501in}{1.244963in}}%
\pgfusepath{clip}%
\pgfsetbuttcap%
\pgfsetroundjoin%
\definecolor{currentfill}{rgb}{0.549020,0.427451,0.192157}%
\pgfsetfillcolor{currentfill}%
\pgfsetlinewidth{1.003750pt}%
\definecolor{currentstroke}{rgb}{0.549020,0.427451,0.192157}%
\pgfsetstrokecolor{currentstroke}%
\pgfsetdash{}{0pt}%
\pgfpathmoveto{\pgfqpoint{3.585584in}{1.227423in}}%
\pgfpathcurveto{\pgfqpoint{3.596635in}{1.227423in}}{\pgfqpoint{3.607234in}{1.231813in}}{\pgfqpoint{3.615047in}{1.239626in}}%
\pgfpathcurveto{\pgfqpoint{3.622861in}{1.247440in}}{\pgfqpoint{3.627251in}{1.258039in}}{\pgfqpoint{3.627251in}{1.269089in}}%
\pgfpathcurveto{\pgfqpoint{3.627251in}{1.280139in}}{\pgfqpoint{3.622861in}{1.290738in}}{\pgfqpoint{3.615047in}{1.298552in}}%
\pgfpathcurveto{\pgfqpoint{3.607234in}{1.306366in}}{\pgfqpoint{3.596635in}{1.310756in}}{\pgfqpoint{3.585584in}{1.310756in}}%
\pgfpathcurveto{\pgfqpoint{3.574534in}{1.310756in}}{\pgfqpoint{3.563935in}{1.306366in}}{\pgfqpoint{3.556122in}{1.298552in}}%
\pgfpathcurveto{\pgfqpoint{3.548308in}{1.290738in}}{\pgfqpoint{3.543918in}{1.280139in}}{\pgfqpoint{3.543918in}{1.269089in}}%
\pgfpathcurveto{\pgfqpoint{3.543918in}{1.258039in}}{\pgfqpoint{3.548308in}{1.247440in}}{\pgfqpoint{3.556122in}{1.239626in}}%
\pgfpathcurveto{\pgfqpoint{3.563935in}{1.231813in}}{\pgfqpoint{3.574534in}{1.227423in}}{\pgfqpoint{3.585584in}{1.227423in}}%
\pgfpathlineto{\pgfqpoint{3.585584in}{1.227423in}}%
\pgfpathclose%
\pgfusepath{stroke,fill}%
\end{pgfscope}%
\begin{pgfscope}%
\pgfpathrectangle{\pgfqpoint{0.490190in}{0.395076in}}{\pgfqpoint{3.464501in}{1.244963in}}%
\pgfusepath{clip}%
\pgfsetbuttcap%
\pgfsetroundjoin%
\definecolor{currentfill}{rgb}{0.517647,0.235294,0.223529}%
\pgfsetfillcolor{currentfill}%
\pgfsetlinewidth{1.003750pt}%
\definecolor{currentstroke}{rgb}{0.517647,0.235294,0.223529}%
\pgfsetstrokecolor{currentstroke}%
\pgfsetdash{}{0pt}%
\pgfpathmoveto{\pgfqpoint{3.605746in}{1.007098in}}%
\pgfpathcurveto{\pgfqpoint{3.616796in}{1.007098in}}{\pgfqpoint{3.627395in}{1.011488in}}{\pgfqpoint{3.635209in}{1.019302in}}%
\pgfpathcurveto{\pgfqpoint{3.643022in}{1.027116in}}{\pgfqpoint{3.647412in}{1.037715in}}{\pgfqpoint{3.647412in}{1.048765in}}%
\pgfpathcurveto{\pgfqpoint{3.647412in}{1.059815in}}{\pgfqpoint{3.643022in}{1.070414in}}{\pgfqpoint{3.635209in}{1.078228in}}%
\pgfpathcurveto{\pgfqpoint{3.627395in}{1.086041in}}{\pgfqpoint{3.616796in}{1.090432in}}{\pgfqpoint{3.605746in}{1.090432in}}%
\pgfpathcurveto{\pgfqpoint{3.594696in}{1.090432in}}{\pgfqpoint{3.584097in}{1.086041in}}{\pgfqpoint{3.576283in}{1.078228in}}%
\pgfpathcurveto{\pgfqpoint{3.568469in}{1.070414in}}{\pgfqpoint{3.564079in}{1.059815in}}{\pgfqpoint{3.564079in}{1.048765in}}%
\pgfpathcurveto{\pgfqpoint{3.564079in}{1.037715in}}{\pgfqpoint{3.568469in}{1.027116in}}{\pgfqpoint{3.576283in}{1.019302in}}%
\pgfpathcurveto{\pgfqpoint{3.584097in}{1.011488in}}{\pgfqpoint{3.594696in}{1.007098in}}{\pgfqpoint{3.605746in}{1.007098in}}%
\pgfpathlineto{\pgfqpoint{3.605746in}{1.007098in}}%
\pgfpathclose%
\pgfusepath{stroke,fill}%
\end{pgfscope}%
\begin{pgfscope}%
\pgfpathrectangle{\pgfqpoint{0.490190in}{0.395076in}}{\pgfqpoint{3.464501in}{1.244963in}}%
\pgfusepath{clip}%
\pgfsetbuttcap%
\pgfsetroundjoin%
\definecolor{currentfill}{rgb}{0.870588,0.619608,0.839216}%
\pgfsetfillcolor{currentfill}%
\pgfsetlinewidth{1.003750pt}%
\definecolor{currentstroke}{rgb}{0.870588,0.619608,0.839216}%
\pgfsetstrokecolor{currentstroke}%
\pgfsetdash{}{0pt}%
\pgfpathmoveto{\pgfqpoint{0.821994in}{0.498631in}}%
\pgfpathcurveto{\pgfqpoint{0.833044in}{0.498631in}}{\pgfqpoint{0.843643in}{0.503022in}}{\pgfqpoint{0.851457in}{0.510835in}}%
\pgfpathcurveto{\pgfqpoint{0.859270in}{0.518649in}}{\pgfqpoint{0.863661in}{0.529248in}}{\pgfqpoint{0.863661in}{0.540298in}}%
\pgfpathcurveto{\pgfqpoint{0.863661in}{0.551348in}}{\pgfqpoint{0.859270in}{0.561947in}}{\pgfqpoint{0.851457in}{0.569761in}}%
\pgfpathcurveto{\pgfqpoint{0.843643in}{0.577575in}}{\pgfqpoint{0.833044in}{0.581965in}}{\pgfqpoint{0.821994in}{0.581965in}}%
\pgfpathcurveto{\pgfqpoint{0.810944in}{0.581965in}}{\pgfqpoint{0.800345in}{0.577575in}}{\pgfqpoint{0.792531in}{0.569761in}}%
\pgfpathcurveto{\pgfqpoint{0.784717in}{0.561947in}}{\pgfqpoint{0.780327in}{0.551348in}}{\pgfqpoint{0.780327in}{0.540298in}}%
\pgfpathcurveto{\pgfqpoint{0.780327in}{0.529248in}}{\pgfqpoint{0.784717in}{0.518649in}}{\pgfqpoint{0.792531in}{0.510835in}}%
\pgfpathcurveto{\pgfqpoint{0.800345in}{0.503022in}}{\pgfqpoint{0.810944in}{0.498631in}}{\pgfqpoint{0.821994in}{0.498631in}}%
\pgfpathlineto{\pgfqpoint{0.821994in}{0.498631in}}%
\pgfpathclose%
\pgfusepath{stroke,fill}%
\end{pgfscope}%
\begin{pgfscope}%
\pgfpathrectangle{\pgfqpoint{0.490190in}{0.395076in}}{\pgfqpoint{3.464501in}{1.244963in}}%
\pgfusepath{clip}%
\pgfsetbuttcap%
\pgfsetroundjoin%
\definecolor{currentfill}{rgb}{0.223529,0.231373,0.474510}%
\pgfsetfillcolor{currentfill}%
\pgfsetlinewidth{1.003750pt}%
\definecolor{currentstroke}{rgb}{0.223529,0.231373,0.474510}%
\pgfsetstrokecolor{currentstroke}%
\pgfsetdash{}{0pt}%
\pgfpathmoveto{\pgfqpoint{3.611109in}{1.296410in}}%
\pgfpathcurveto{\pgfqpoint{3.622159in}{1.296410in}}{\pgfqpoint{3.632758in}{1.300801in}}{\pgfqpoint{3.640572in}{1.308614in}}%
\pgfpathcurveto{\pgfqpoint{3.648385in}{1.316428in}}{\pgfqpoint{3.652775in}{1.327027in}}{\pgfqpoint{3.652775in}{1.338077in}}%
\pgfpathcurveto{\pgfqpoint{3.652775in}{1.349127in}}{\pgfqpoint{3.648385in}{1.359726in}}{\pgfqpoint{3.640572in}{1.367540in}}%
\pgfpathcurveto{\pgfqpoint{3.632758in}{1.375353in}}{\pgfqpoint{3.622159in}{1.379744in}}{\pgfqpoint{3.611109in}{1.379744in}}%
\pgfpathcurveto{\pgfqpoint{3.600059in}{1.379744in}}{\pgfqpoint{3.589460in}{1.375353in}}{\pgfqpoint{3.581646in}{1.367540in}}%
\pgfpathcurveto{\pgfqpoint{3.573832in}{1.359726in}}{\pgfqpoint{3.569442in}{1.349127in}}{\pgfqpoint{3.569442in}{1.338077in}}%
\pgfpathcurveto{\pgfqpoint{3.569442in}{1.327027in}}{\pgfqpoint{3.573832in}{1.316428in}}{\pgfqpoint{3.581646in}{1.308614in}}%
\pgfpathcurveto{\pgfqpoint{3.589460in}{1.300801in}}{\pgfqpoint{3.600059in}{1.296410in}}{\pgfqpoint{3.611109in}{1.296410in}}%
\pgfpathlineto{\pgfqpoint{3.611109in}{1.296410in}}%
\pgfpathclose%
\pgfusepath{stroke,fill}%
\end{pgfscope}%
\begin{pgfscope}%
\pgfpathrectangle{\pgfqpoint{0.490190in}{0.395076in}}{\pgfqpoint{3.464501in}{1.244963in}}%
\pgfusepath{clip}%
\pgfsetbuttcap%
\pgfsetroundjoin%
\definecolor{currentfill}{rgb}{0.388235,0.474510,0.223529}%
\pgfsetfillcolor{currentfill}%
\pgfsetlinewidth{1.003750pt}%
\definecolor{currentstroke}{rgb}{0.388235,0.474510,0.223529}%
\pgfsetstrokecolor{currentstroke}%
\pgfsetdash{}{0pt}%
\pgfpathmoveto{\pgfqpoint{3.514862in}{1.140769in}}%
\pgfpathcurveto{\pgfqpoint{3.525912in}{1.140769in}}{\pgfqpoint{3.536511in}{1.145159in}}{\pgfqpoint{3.544324in}{1.152973in}}%
\pgfpathcurveto{\pgfqpoint{3.552138in}{1.160787in}}{\pgfqpoint{3.556528in}{1.171386in}}{\pgfqpoint{3.556528in}{1.182436in}}%
\pgfpathcurveto{\pgfqpoint{3.556528in}{1.193486in}}{\pgfqpoint{3.552138in}{1.204085in}}{\pgfqpoint{3.544324in}{1.211899in}}%
\pgfpathcurveto{\pgfqpoint{3.536511in}{1.219712in}}{\pgfqpoint{3.525912in}{1.224102in}}{\pgfqpoint{3.514862in}{1.224102in}}%
\pgfpathcurveto{\pgfqpoint{3.503812in}{1.224102in}}{\pgfqpoint{3.493212in}{1.219712in}}{\pgfqpoint{3.485399in}{1.211899in}}%
\pgfpathcurveto{\pgfqpoint{3.477585in}{1.204085in}}{\pgfqpoint{3.473195in}{1.193486in}}{\pgfqpoint{3.473195in}{1.182436in}}%
\pgfpathcurveto{\pgfqpoint{3.473195in}{1.171386in}}{\pgfqpoint{3.477585in}{1.160787in}}{\pgfqpoint{3.485399in}{1.152973in}}%
\pgfpathcurveto{\pgfqpoint{3.493212in}{1.145159in}}{\pgfqpoint{3.503812in}{1.140769in}}{\pgfqpoint{3.514862in}{1.140769in}}%
\pgfpathlineto{\pgfqpoint{3.514862in}{1.140769in}}%
\pgfpathclose%
\pgfusepath{stroke,fill}%
\end{pgfscope}%
\begin{pgfscope}%
\pgfpathrectangle{\pgfqpoint{0.490190in}{0.395076in}}{\pgfqpoint{3.464501in}{1.244963in}}%
\pgfusepath{clip}%
\pgfsetbuttcap%
\pgfsetroundjoin%
\definecolor{currentfill}{rgb}{0.807843,0.858824,0.611765}%
\pgfsetfillcolor{currentfill}%
\pgfsetlinewidth{1.003750pt}%
\definecolor{currentstroke}{rgb}{0.807843,0.858824,0.611765}%
\pgfsetstrokecolor{currentstroke}%
\pgfsetdash{}{0pt}%
\pgfpathmoveto{\pgfqpoint{3.356419in}{1.031059in}}%
\pgfpathcurveto{\pgfqpoint{3.367469in}{1.031059in}}{\pgfqpoint{3.378068in}{1.035450in}}{\pgfqpoint{3.385881in}{1.043263in}}%
\pgfpathcurveto{\pgfqpoint{3.393695in}{1.051077in}}{\pgfqpoint{3.398085in}{1.061676in}}{\pgfqpoint{3.398085in}{1.072726in}}%
\pgfpathcurveto{\pgfqpoint{3.398085in}{1.083776in}}{\pgfqpoint{3.393695in}{1.094375in}}{\pgfqpoint{3.385881in}{1.102189in}}%
\pgfpathcurveto{\pgfqpoint{3.378068in}{1.110002in}}{\pgfqpoint{3.367469in}{1.114393in}}{\pgfqpoint{3.356419in}{1.114393in}}%
\pgfpathcurveto{\pgfqpoint{3.345369in}{1.114393in}}{\pgfqpoint{3.334770in}{1.110002in}}{\pgfqpoint{3.326956in}{1.102189in}}%
\pgfpathcurveto{\pgfqpoint{3.319142in}{1.094375in}}{\pgfqpoint{3.314752in}{1.083776in}}{\pgfqpoint{3.314752in}{1.072726in}}%
\pgfpathcurveto{\pgfqpoint{3.314752in}{1.061676in}}{\pgfqpoint{3.319142in}{1.051077in}}{\pgfqpoint{3.326956in}{1.043263in}}%
\pgfpathcurveto{\pgfqpoint{3.334770in}{1.035450in}}{\pgfqpoint{3.345369in}{1.031059in}}{\pgfqpoint{3.356419in}{1.031059in}}%
\pgfpathlineto{\pgfqpoint{3.356419in}{1.031059in}}%
\pgfpathclose%
\pgfusepath{stroke,fill}%
\end{pgfscope}%
\begin{pgfscope}%
\pgfpathrectangle{\pgfqpoint{0.490190in}{0.395076in}}{\pgfqpoint{3.464501in}{1.244963in}}%
\pgfusepath{clip}%
\pgfsetbuttcap%
\pgfsetroundjoin%
\definecolor{currentfill}{rgb}{0.905882,0.796078,0.580392}%
\pgfsetfillcolor{currentfill}%
\pgfsetlinewidth{1.003750pt}%
\definecolor{currentstroke}{rgb}{0.905882,0.796078,0.580392}%
\pgfsetstrokecolor{currentstroke}%
\pgfsetdash{}{0pt}%
\pgfpathmoveto{\pgfqpoint{3.311928in}{0.852688in}}%
\pgfpathcurveto{\pgfqpoint{3.322979in}{0.852688in}}{\pgfqpoint{3.333578in}{0.857078in}}{\pgfqpoint{3.341391in}{0.864892in}}%
\pgfpathcurveto{\pgfqpoint{3.349205in}{0.872705in}}{\pgfqpoint{3.353595in}{0.883304in}}{\pgfqpoint{3.353595in}{0.894354in}}%
\pgfpathcurveto{\pgfqpoint{3.353595in}{0.905404in}}{\pgfqpoint{3.349205in}{0.916004in}}{\pgfqpoint{3.341391in}{0.923817in}}%
\pgfpathcurveto{\pgfqpoint{3.333578in}{0.931631in}}{\pgfqpoint{3.322979in}{0.936021in}}{\pgfqpoint{3.311928in}{0.936021in}}%
\pgfpathcurveto{\pgfqpoint{3.300878in}{0.936021in}}{\pgfqpoint{3.290279in}{0.931631in}}{\pgfqpoint{3.282466in}{0.923817in}}%
\pgfpathcurveto{\pgfqpoint{3.274652in}{0.916004in}}{\pgfqpoint{3.270262in}{0.905404in}}{\pgfqpoint{3.270262in}{0.894354in}}%
\pgfpathcurveto{\pgfqpoint{3.270262in}{0.883304in}}{\pgfqpoint{3.274652in}{0.872705in}}{\pgfqpoint{3.282466in}{0.864892in}}%
\pgfpathcurveto{\pgfqpoint{3.290279in}{0.857078in}}{\pgfqpoint{3.300878in}{0.852688in}}{\pgfqpoint{3.311928in}{0.852688in}}%
\pgfpathlineto{\pgfqpoint{3.311928in}{0.852688in}}%
\pgfpathclose%
\pgfusepath{stroke,fill}%
\end{pgfscope}%
\begin{pgfscope}%
\pgfpathrectangle{\pgfqpoint{0.490190in}{0.395076in}}{\pgfqpoint{3.464501in}{1.244963in}}%
\pgfusepath{clip}%
\pgfsetbuttcap%
\pgfsetroundjoin%
\definecolor{currentfill}{rgb}{0.870588,0.619608,0.839216}%
\pgfsetfillcolor{currentfill}%
\pgfsetlinewidth{1.003750pt}%
\definecolor{currentstroke}{rgb}{0.870588,0.619608,0.839216}%
\pgfsetstrokecolor{currentstroke}%
\pgfsetdash{}{0pt}%
\pgfpathmoveto{\pgfqpoint{0.647667in}{0.429176in}}%
\pgfpathcurveto{\pgfqpoint{0.658717in}{0.429176in}}{\pgfqpoint{0.669316in}{0.433566in}}{\pgfqpoint{0.677130in}{0.441380in}}%
\pgfpathcurveto{\pgfqpoint{0.684943in}{0.449193in}}{\pgfqpoint{0.689334in}{0.459792in}}{\pgfqpoint{0.689334in}{0.470842in}}%
\pgfpathcurveto{\pgfqpoint{0.689334in}{0.481892in}}{\pgfqpoint{0.684943in}{0.492492in}}{\pgfqpoint{0.677130in}{0.500305in}}%
\pgfpathcurveto{\pgfqpoint{0.669316in}{0.508119in}}{\pgfqpoint{0.658717in}{0.512509in}}{\pgfqpoint{0.647667in}{0.512509in}}%
\pgfpathcurveto{\pgfqpoint{0.636617in}{0.512509in}}{\pgfqpoint{0.626018in}{0.508119in}}{\pgfqpoint{0.618204in}{0.500305in}}%
\pgfpathcurveto{\pgfqpoint{0.610391in}{0.492492in}}{\pgfqpoint{0.606000in}{0.481892in}}{\pgfqpoint{0.606000in}{0.470842in}}%
\pgfpathcurveto{\pgfqpoint{0.606000in}{0.459792in}}{\pgfqpoint{0.610391in}{0.449193in}}{\pgfqpoint{0.618204in}{0.441380in}}%
\pgfpathcurveto{\pgfqpoint{0.626018in}{0.433566in}}{\pgfqpoint{0.636617in}{0.429176in}}{\pgfqpoint{0.647667in}{0.429176in}}%
\pgfpathlineto{\pgfqpoint{0.647667in}{0.429176in}}%
\pgfpathclose%
\pgfusepath{stroke,fill}%
\end{pgfscope}%
\begin{pgfscope}%
\pgfsetbuttcap%
\pgfsetroundjoin%
\definecolor{currentfill}{rgb}{0.000000,0.000000,0.000000}%
\pgfsetfillcolor{currentfill}%
\pgfsetlinewidth{0.803000pt}%
\definecolor{currentstroke}{rgb}{0.000000,0.000000,0.000000}%
\pgfsetstrokecolor{currentstroke}%
\pgfsetdash{}{0pt}%
\pgfsys@defobject{currentmarker}{\pgfqpoint{0.000000in}{-0.048611in}}{\pgfqpoint{0.000000in}{0.000000in}}{%
\pgfpathmoveto{\pgfqpoint{0.000000in}{0.000000in}}%
\pgfpathlineto{\pgfqpoint{0.000000in}{-0.048611in}}%
\pgfusepath{stroke,fill}%
}%
\begin{pgfscope}%
\pgfsys@transformshift{0.849281in}{0.395076in}%
\pgfsys@useobject{currentmarker}{}%
\end{pgfscope}%
\end{pgfscope}%
\begin{pgfscope}%
\definecolor{textcolor}{rgb}{0.000000,0.000000,0.000000}%
\pgfsetstrokecolor{textcolor}%
\pgfsetfillcolor{textcolor}%
\pgftext[x=0.849281in,y=0.297854in,,top]{\color{textcolor}\sffamily\fontsize{11.000000}{13.200000}\selectfont 36.0}%
\end{pgfscope}%
\begin{pgfscope}%
\pgfsetbuttcap%
\pgfsetroundjoin%
\definecolor{currentfill}{rgb}{0.000000,0.000000,0.000000}%
\pgfsetfillcolor{currentfill}%
\pgfsetlinewidth{0.803000pt}%
\definecolor{currentstroke}{rgb}{0.000000,0.000000,0.000000}%
\pgfsetstrokecolor{currentstroke}%
\pgfsetdash{}{0pt}%
\pgfsys@defobject{currentmarker}{\pgfqpoint{0.000000in}{-0.048611in}}{\pgfqpoint{0.000000in}{0.000000in}}{%
\pgfpathmoveto{\pgfqpoint{0.000000in}{0.000000in}}%
\pgfpathlineto{\pgfqpoint{0.000000in}{-0.048611in}}%
\pgfusepath{stroke,fill}%
}%
\begin{pgfscope}%
\pgfsys@transformshift{1.647170in}{0.395076in}%
\pgfsys@useobject{currentmarker}{}%
\end{pgfscope}%
\end{pgfscope}%
\begin{pgfscope}%
\definecolor{textcolor}{rgb}{0.000000,0.000000,0.000000}%
\pgfsetstrokecolor{textcolor}%
\pgfsetfillcolor{textcolor}%
\pgftext[x=1.647170in,y=0.297854in,,top]{\color{textcolor}\sffamily\fontsize{11.000000}{13.200000}\selectfont 36.3}%
\end{pgfscope}%
\begin{pgfscope}%
\pgfsetbuttcap%
\pgfsetroundjoin%
\definecolor{currentfill}{rgb}{0.000000,0.000000,0.000000}%
\pgfsetfillcolor{currentfill}%
\pgfsetlinewidth{0.803000pt}%
\definecolor{currentstroke}{rgb}{0.000000,0.000000,0.000000}%
\pgfsetstrokecolor{currentstroke}%
\pgfsetdash{}{0pt}%
\pgfsys@defobject{currentmarker}{\pgfqpoint{0.000000in}{-0.048611in}}{\pgfqpoint{0.000000in}{0.000000in}}{%
\pgfpathmoveto{\pgfqpoint{0.000000in}{0.000000in}}%
\pgfpathlineto{\pgfqpoint{0.000000in}{-0.048611in}}%
\pgfusepath{stroke,fill}%
}%
\begin{pgfscope}%
\pgfsys@transformshift{2.445060in}{0.395076in}%
\pgfsys@useobject{currentmarker}{}%
\end{pgfscope}%
\end{pgfscope}%
\begin{pgfscope}%
\definecolor{textcolor}{rgb}{0.000000,0.000000,0.000000}%
\pgfsetstrokecolor{textcolor}%
\pgfsetfillcolor{textcolor}%
\pgftext[x=2.445060in,y=0.297854in,,top]{\color{textcolor}\sffamily\fontsize{11.000000}{13.200000}\selectfont 36.6}%
\end{pgfscope}%
\begin{pgfscope}%
\pgfsetbuttcap%
\pgfsetroundjoin%
\definecolor{currentfill}{rgb}{0.000000,0.000000,0.000000}%
\pgfsetfillcolor{currentfill}%
\pgfsetlinewidth{0.803000pt}%
\definecolor{currentstroke}{rgb}{0.000000,0.000000,0.000000}%
\pgfsetstrokecolor{currentstroke}%
\pgfsetdash{}{0pt}%
\pgfsys@defobject{currentmarker}{\pgfqpoint{0.000000in}{-0.048611in}}{\pgfqpoint{0.000000in}{0.000000in}}{%
\pgfpathmoveto{\pgfqpoint{0.000000in}{0.000000in}}%
\pgfpathlineto{\pgfqpoint{0.000000in}{-0.048611in}}%
\pgfusepath{stroke,fill}%
}%
\begin{pgfscope}%
\pgfsys@transformshift{3.242949in}{0.395076in}%
\pgfsys@useobject{currentmarker}{}%
\end{pgfscope}%
\end{pgfscope}%
\begin{pgfscope}%
\definecolor{textcolor}{rgb}{0.000000,0.000000,0.000000}%
\pgfsetstrokecolor{textcolor}%
\pgfsetfillcolor{textcolor}%
\pgftext[x=3.242949in,y=0.297854in,,top]{\color{textcolor}\sffamily\fontsize{11.000000}{13.200000}\selectfont 36.9}%
\end{pgfscope}%
\begin{pgfscope}%
\pgfsetbuttcap%
\pgfsetroundjoin%
\definecolor{currentfill}{rgb}{0.000000,0.000000,0.000000}%
\pgfsetfillcolor{currentfill}%
\pgfsetlinewidth{0.803000pt}%
\definecolor{currentstroke}{rgb}{0.000000,0.000000,0.000000}%
\pgfsetstrokecolor{currentstroke}%
\pgfsetdash{}{0pt}%
\pgfsys@defobject{currentmarker}{\pgfqpoint{-0.048611in}{0.000000in}}{\pgfqpoint{-0.000000in}{0.000000in}}{%
\pgfpathmoveto{\pgfqpoint{-0.000000in}{0.000000in}}%
\pgfpathlineto{\pgfqpoint{-0.048611in}{0.000000in}}%
\pgfusepath{stroke,fill}%
}%
\begin{pgfscope}%
\pgfsys@transformshift{0.490190in}{0.595644in}%
\pgfsys@useobject{currentmarker}{}%
\end{pgfscope}%
\end{pgfscope}%
\begin{pgfscope}%
\definecolor{textcolor}{rgb}{0.000000,0.000000,0.000000}%
\pgfsetstrokecolor{textcolor}%
\pgfsetfillcolor{textcolor}%
\pgftext[x=0.150000in, y=0.537606in, left, base]{\color{textcolor}\sffamily\fontsize{11.000000}{13.200000}\selectfont 1.5}%
\end{pgfscope}%
\begin{pgfscope}%
\pgfsetbuttcap%
\pgfsetroundjoin%
\definecolor{currentfill}{rgb}{0.000000,0.000000,0.000000}%
\pgfsetfillcolor{currentfill}%
\pgfsetlinewidth{0.803000pt}%
\definecolor{currentstroke}{rgb}{0.000000,0.000000,0.000000}%
\pgfsetstrokecolor{currentstroke}%
\pgfsetdash{}{0pt}%
\pgfsys@defobject{currentmarker}{\pgfqpoint{-0.048611in}{0.000000in}}{\pgfqpoint{-0.000000in}{0.000000in}}{%
\pgfpathmoveto{\pgfqpoint{-0.000000in}{0.000000in}}%
\pgfpathlineto{\pgfqpoint{-0.048611in}{0.000000in}}%
\pgfusepath{stroke,fill}%
}%
\begin{pgfscope}%
\pgfsys@transformshift{0.490190in}{0.889798in}%
\pgfsys@useobject{currentmarker}{}%
\end{pgfscope}%
\end{pgfscope}%
\begin{pgfscope}%
\definecolor{textcolor}{rgb}{0.000000,0.000000,0.000000}%
\pgfsetstrokecolor{textcolor}%
\pgfsetfillcolor{textcolor}%
\pgftext[x=0.150000in, y=0.831761in, left, base]{\color{textcolor}\sffamily\fontsize{11.000000}{13.200000}\selectfont 2.0}%
\end{pgfscope}%
\begin{pgfscope}%
\pgfsetbuttcap%
\pgfsetroundjoin%
\definecolor{currentfill}{rgb}{0.000000,0.000000,0.000000}%
\pgfsetfillcolor{currentfill}%
\pgfsetlinewidth{0.803000pt}%
\definecolor{currentstroke}{rgb}{0.000000,0.000000,0.000000}%
\pgfsetstrokecolor{currentstroke}%
\pgfsetdash{}{0pt}%
\pgfsys@defobject{currentmarker}{\pgfqpoint{-0.048611in}{0.000000in}}{\pgfqpoint{-0.000000in}{0.000000in}}{%
\pgfpathmoveto{\pgfqpoint{-0.000000in}{0.000000in}}%
\pgfpathlineto{\pgfqpoint{-0.048611in}{0.000000in}}%
\pgfusepath{stroke,fill}%
}%
\begin{pgfscope}%
\pgfsys@transformshift{0.490190in}{1.183953in}%
\pgfsys@useobject{currentmarker}{}%
\end{pgfscope}%
\end{pgfscope}%
\begin{pgfscope}%
\definecolor{textcolor}{rgb}{0.000000,0.000000,0.000000}%
\pgfsetstrokecolor{textcolor}%
\pgfsetfillcolor{textcolor}%
\pgftext[x=0.150000in, y=1.125915in, left, base]{\color{textcolor}\sffamily\fontsize{11.000000}{13.200000}\selectfont 2.5}%
\end{pgfscope}%
\begin{pgfscope}%
\pgfsetbuttcap%
\pgfsetroundjoin%
\definecolor{currentfill}{rgb}{0.000000,0.000000,0.000000}%
\pgfsetfillcolor{currentfill}%
\pgfsetlinewidth{0.803000pt}%
\definecolor{currentstroke}{rgb}{0.000000,0.000000,0.000000}%
\pgfsetstrokecolor{currentstroke}%
\pgfsetdash{}{0pt}%
\pgfsys@defobject{currentmarker}{\pgfqpoint{-0.048611in}{0.000000in}}{\pgfqpoint{-0.000000in}{0.000000in}}{%
\pgfpathmoveto{\pgfqpoint{-0.000000in}{0.000000in}}%
\pgfpathlineto{\pgfqpoint{-0.048611in}{0.000000in}}%
\pgfusepath{stroke,fill}%
}%
\begin{pgfscope}%
\pgfsys@transformshift{0.490190in}{1.478107in}%
\pgfsys@useobject{currentmarker}{}%
\end{pgfscope}%
\end{pgfscope}%
\begin{pgfscope}%
\definecolor{textcolor}{rgb}{0.000000,0.000000,0.000000}%
\pgfsetstrokecolor{textcolor}%
\pgfsetfillcolor{textcolor}%
\pgftext[x=0.150000in, y=1.420069in, left, base]{\color{textcolor}\sffamily\fontsize{11.000000}{13.200000}\selectfont 3.0}%
\end{pgfscope}%
\begin{pgfscope}%
\pgfsetrectcap%
\pgfsetmiterjoin%
\pgfsetlinewidth{1.505625pt}%
\definecolor{currentstroke}{rgb}{0.000000,0.000000,0.000000}%
\pgfsetstrokecolor{currentstroke}%
\pgfsetdash{}{0pt}%
\pgfpathmoveto{\pgfqpoint{0.490190in}{0.395076in}}%
\pgfpathlineto{\pgfqpoint{0.490190in}{1.640039in}}%
\pgfusepath{stroke}%
\end{pgfscope}%
\begin{pgfscope}%
\pgfsetrectcap%
\pgfsetmiterjoin%
\pgfsetlinewidth{1.505625pt}%
\definecolor{currentstroke}{rgb}{0.000000,0.000000,0.000000}%
\pgfsetstrokecolor{currentstroke}%
\pgfsetdash{}{0pt}%
\pgfpathmoveto{\pgfqpoint{0.490190in}{0.395076in}}%
\pgfpathlineto{\pgfqpoint{3.954691in}{0.395076in}}%
\pgfusepath{stroke}%
\end{pgfscope}%
\begin{pgfscope}%
\definecolor{textcolor}{rgb}{0.000000,0.000000,0.000000}%
\pgfsetstrokecolor{textcolor}%
\pgfsetfillcolor{textcolor}%
\pgftext[x=0.663415in, y=1.500962in, left, base]{\color{textcolor}\sffamily\fontsize{12.000000}{14.400000}\selectfont \(\displaystyle \rho: 0.91\)}%
\end{pgfscope}%
\begin{pgfscope}%
\definecolor{textcolor}{rgb}{0.000000,0.000000,0.000000}%
\pgfsetstrokecolor{textcolor}%
\pgfsetfillcolor{textcolor}%
\pgftext[x=0.663415in, y=1.314341in, left, base]{\color{textcolor}\sffamily\fontsize{12.000000}{14.400000}\selectfont \(\displaystyle r: 0.73\)}%
\end{pgfscope}%
\begin{pgfscope}%
\definecolor{textcolor}{rgb}{0.000000,0.000000,0.000000}%
\pgfsetstrokecolor{textcolor}%
\pgfsetfillcolor{textcolor}%
\pgftext[x=2.222440in,y=1.723372in,,base]{\color{textcolor}\sffamily\fontsize{12.000000}{14.400000}\selectfont WRN-16-1, \(\displaystyle \mu=0.0\), \(\displaystyle T = 25k \)}%
\end{pgfscope}%
\begin{pgfscope}%
\pgfsetbuttcap%
\pgfsetmiterjoin%
\definecolor{currentfill}{rgb}{1.000000,1.000000,1.000000}%
\pgfsetfillcolor{currentfill}%
\pgfsetlinewidth{0.000000pt}%
\definecolor{currentstroke}{rgb}{0.000000,0.000000,0.000000}%
\pgfsetstrokecolor{currentstroke}%
\pgfsetstrokeopacity{0.000000}%
\pgfsetdash{}{0pt}%
\pgfpathmoveto{\pgfqpoint{4.171222in}{0.395076in}}%
\pgfpathlineto{\pgfqpoint{4.233470in}{0.395076in}}%
\pgfpathlineto{\pgfqpoint{4.233470in}{1.640039in}}%
\pgfpathlineto{\pgfqpoint{4.171222in}{1.640039in}}%
\pgfpathlineto{\pgfqpoint{4.171222in}{0.395076in}}%
\pgfpathclose%
\pgfusepath{fill}%
\end{pgfscope}%
\begin{pgfscope}%
\pgfpathrectangle{\pgfqpoint{4.171222in}{0.395076in}}{\pgfqpoint{0.062248in}{1.244963in}}%
\pgfusepath{clip}%
\pgfsetbuttcap%
\pgfsetmiterjoin%
\definecolor{currentfill}{rgb}{1.000000,1.000000,1.000000}%
\pgfsetfillcolor{currentfill}%
\pgfsetlinewidth{0.010037pt}%
\definecolor{currentstroke}{rgb}{1.000000,1.000000,1.000000}%
\pgfsetstrokecolor{currentstroke}%
\pgfsetdash{}{0pt}%
\pgfusepath{stroke,fill}%
\end{pgfscope}%
\begin{pgfscope}%
\pgfpathrectangle{\pgfqpoint{4.171222in}{0.395076in}}{\pgfqpoint{0.062248in}{1.244963in}}%
\pgfusepath{clip}%
\pgfsetbuttcap%
\pgfsetroundjoin%
\definecolor{currentfill}{rgb}{0.223529,0.231373,0.474510}%
\pgfsetfillcolor{currentfill}%
\pgfsetlinewidth{0.000000pt}%
\definecolor{currentstroke}{rgb}{0.000000,0.000000,0.000000}%
\pgfsetstrokecolor{currentstroke}%
\pgfsetdash{}{0pt}%
\pgfpathmoveto{\pgfqpoint{4.171222in}{0.395076in}}%
\pgfpathlineto{\pgfqpoint{4.233470in}{0.395076in}}%
\pgfpathlineto{\pgfqpoint{4.233470in}{0.457324in}}%
\pgfpathlineto{\pgfqpoint{4.171222in}{0.457324in}}%
\pgfpathlineto{\pgfqpoint{4.171222in}{0.395076in}}%
\pgfusepath{fill}%
\end{pgfscope}%
\begin{pgfscope}%
\pgfpathrectangle{\pgfqpoint{4.171222in}{0.395076in}}{\pgfqpoint{0.062248in}{1.244963in}}%
\pgfusepath{clip}%
\pgfsetbuttcap%
\pgfsetroundjoin%
\definecolor{currentfill}{rgb}{0.321569,0.329412,0.639216}%
\pgfsetfillcolor{currentfill}%
\pgfsetlinewidth{0.000000pt}%
\definecolor{currentstroke}{rgb}{0.000000,0.000000,0.000000}%
\pgfsetstrokecolor{currentstroke}%
\pgfsetdash{}{0pt}%
\pgfpathmoveto{\pgfqpoint{4.171222in}{0.457324in}}%
\pgfpathlineto{\pgfqpoint{4.233470in}{0.457324in}}%
\pgfpathlineto{\pgfqpoint{4.233470in}{0.519573in}}%
\pgfpathlineto{\pgfqpoint{4.171222in}{0.519573in}}%
\pgfpathlineto{\pgfqpoint{4.171222in}{0.457324in}}%
\pgfusepath{fill}%
\end{pgfscope}%
\begin{pgfscope}%
\pgfpathrectangle{\pgfqpoint{4.171222in}{0.395076in}}{\pgfqpoint{0.062248in}{1.244963in}}%
\pgfusepath{clip}%
\pgfsetbuttcap%
\pgfsetroundjoin%
\definecolor{currentfill}{rgb}{0.419608,0.431373,0.811765}%
\pgfsetfillcolor{currentfill}%
\pgfsetlinewidth{0.000000pt}%
\definecolor{currentstroke}{rgb}{0.000000,0.000000,0.000000}%
\pgfsetstrokecolor{currentstroke}%
\pgfsetdash{}{0pt}%
\pgfpathmoveto{\pgfqpoint{4.171222in}{0.519573in}}%
\pgfpathlineto{\pgfqpoint{4.233470in}{0.519573in}}%
\pgfpathlineto{\pgfqpoint{4.233470in}{0.581821in}}%
\pgfpathlineto{\pgfqpoint{4.171222in}{0.581821in}}%
\pgfpathlineto{\pgfqpoint{4.171222in}{0.519573in}}%
\pgfusepath{fill}%
\end{pgfscope}%
\begin{pgfscope}%
\pgfpathrectangle{\pgfqpoint{4.171222in}{0.395076in}}{\pgfqpoint{0.062248in}{1.244963in}}%
\pgfusepath{clip}%
\pgfsetbuttcap%
\pgfsetroundjoin%
\definecolor{currentfill}{rgb}{0.611765,0.619608,0.870588}%
\pgfsetfillcolor{currentfill}%
\pgfsetlinewidth{0.000000pt}%
\definecolor{currentstroke}{rgb}{0.000000,0.000000,0.000000}%
\pgfsetstrokecolor{currentstroke}%
\pgfsetdash{}{0pt}%
\pgfpathmoveto{\pgfqpoint{4.171222in}{0.581821in}}%
\pgfpathlineto{\pgfqpoint{4.233470in}{0.581821in}}%
\pgfpathlineto{\pgfqpoint{4.233470in}{0.644069in}}%
\pgfpathlineto{\pgfqpoint{4.171222in}{0.644069in}}%
\pgfpathlineto{\pgfqpoint{4.171222in}{0.581821in}}%
\pgfusepath{fill}%
\end{pgfscope}%
\begin{pgfscope}%
\pgfpathrectangle{\pgfqpoint{4.171222in}{0.395076in}}{\pgfqpoint{0.062248in}{1.244963in}}%
\pgfusepath{clip}%
\pgfsetbuttcap%
\pgfsetroundjoin%
\definecolor{currentfill}{rgb}{0.388235,0.474510,0.223529}%
\pgfsetfillcolor{currentfill}%
\pgfsetlinewidth{0.000000pt}%
\definecolor{currentstroke}{rgb}{0.000000,0.000000,0.000000}%
\pgfsetstrokecolor{currentstroke}%
\pgfsetdash{}{0pt}%
\pgfpathmoveto{\pgfqpoint{4.171222in}{0.644069in}}%
\pgfpathlineto{\pgfqpoint{4.233470in}{0.644069in}}%
\pgfpathlineto{\pgfqpoint{4.233470in}{0.706317in}}%
\pgfpathlineto{\pgfqpoint{4.171222in}{0.706317in}}%
\pgfpathlineto{\pgfqpoint{4.171222in}{0.644069in}}%
\pgfusepath{fill}%
\end{pgfscope}%
\begin{pgfscope}%
\pgfpathrectangle{\pgfqpoint{4.171222in}{0.395076in}}{\pgfqpoint{0.062248in}{1.244963in}}%
\pgfusepath{clip}%
\pgfsetbuttcap%
\pgfsetroundjoin%
\definecolor{currentfill}{rgb}{0.549020,0.635294,0.321569}%
\pgfsetfillcolor{currentfill}%
\pgfsetlinewidth{0.000000pt}%
\definecolor{currentstroke}{rgb}{0.000000,0.000000,0.000000}%
\pgfsetstrokecolor{currentstroke}%
\pgfsetdash{}{0pt}%
\pgfpathmoveto{\pgfqpoint{4.171222in}{0.706317in}}%
\pgfpathlineto{\pgfqpoint{4.233470in}{0.706317in}}%
\pgfpathlineto{\pgfqpoint{4.233470in}{0.768565in}}%
\pgfpathlineto{\pgfqpoint{4.171222in}{0.768565in}}%
\pgfpathlineto{\pgfqpoint{4.171222in}{0.706317in}}%
\pgfusepath{fill}%
\end{pgfscope}%
\begin{pgfscope}%
\pgfpathrectangle{\pgfqpoint{4.171222in}{0.395076in}}{\pgfqpoint{0.062248in}{1.244963in}}%
\pgfusepath{clip}%
\pgfsetbuttcap%
\pgfsetroundjoin%
\definecolor{currentfill}{rgb}{0.709804,0.811765,0.419608}%
\pgfsetfillcolor{currentfill}%
\pgfsetlinewidth{0.000000pt}%
\definecolor{currentstroke}{rgb}{0.000000,0.000000,0.000000}%
\pgfsetstrokecolor{currentstroke}%
\pgfsetdash{}{0pt}%
\pgfpathmoveto{\pgfqpoint{4.171222in}{0.768565in}}%
\pgfpathlineto{\pgfqpoint{4.233470in}{0.768565in}}%
\pgfpathlineto{\pgfqpoint{4.233470in}{0.830813in}}%
\pgfpathlineto{\pgfqpoint{4.171222in}{0.830813in}}%
\pgfpathlineto{\pgfqpoint{4.171222in}{0.768565in}}%
\pgfusepath{fill}%
\end{pgfscope}%
\begin{pgfscope}%
\pgfpathrectangle{\pgfqpoint{4.171222in}{0.395076in}}{\pgfqpoint{0.062248in}{1.244963in}}%
\pgfusepath{clip}%
\pgfsetbuttcap%
\pgfsetroundjoin%
\definecolor{currentfill}{rgb}{0.807843,0.858824,0.611765}%
\pgfsetfillcolor{currentfill}%
\pgfsetlinewidth{0.000000pt}%
\definecolor{currentstroke}{rgb}{0.000000,0.000000,0.000000}%
\pgfsetstrokecolor{currentstroke}%
\pgfsetdash{}{0pt}%
\pgfpathmoveto{\pgfqpoint{4.171222in}{0.830813in}}%
\pgfpathlineto{\pgfqpoint{4.233470in}{0.830813in}}%
\pgfpathlineto{\pgfqpoint{4.233470in}{0.893061in}}%
\pgfpathlineto{\pgfqpoint{4.171222in}{0.893061in}}%
\pgfpathlineto{\pgfqpoint{4.171222in}{0.830813in}}%
\pgfusepath{fill}%
\end{pgfscope}%
\begin{pgfscope}%
\pgfpathrectangle{\pgfqpoint{4.171222in}{0.395076in}}{\pgfqpoint{0.062248in}{1.244963in}}%
\pgfusepath{clip}%
\pgfsetbuttcap%
\pgfsetroundjoin%
\definecolor{currentfill}{rgb}{0.549020,0.427451,0.192157}%
\pgfsetfillcolor{currentfill}%
\pgfsetlinewidth{0.000000pt}%
\definecolor{currentstroke}{rgb}{0.000000,0.000000,0.000000}%
\pgfsetstrokecolor{currentstroke}%
\pgfsetdash{}{0pt}%
\pgfpathmoveto{\pgfqpoint{4.171222in}{0.893061in}}%
\pgfpathlineto{\pgfqpoint{4.233470in}{0.893061in}}%
\pgfpathlineto{\pgfqpoint{4.233470in}{0.955310in}}%
\pgfpathlineto{\pgfqpoint{4.171222in}{0.955310in}}%
\pgfpathlineto{\pgfqpoint{4.171222in}{0.893061in}}%
\pgfusepath{fill}%
\end{pgfscope}%
\begin{pgfscope}%
\pgfpathrectangle{\pgfqpoint{4.171222in}{0.395076in}}{\pgfqpoint{0.062248in}{1.244963in}}%
\pgfusepath{clip}%
\pgfsetbuttcap%
\pgfsetroundjoin%
\definecolor{currentfill}{rgb}{0.741176,0.619608,0.223529}%
\pgfsetfillcolor{currentfill}%
\pgfsetlinewidth{0.000000pt}%
\definecolor{currentstroke}{rgb}{0.000000,0.000000,0.000000}%
\pgfsetstrokecolor{currentstroke}%
\pgfsetdash{}{0pt}%
\pgfpathmoveto{\pgfqpoint{4.171222in}{0.955310in}}%
\pgfpathlineto{\pgfqpoint{4.233470in}{0.955310in}}%
\pgfpathlineto{\pgfqpoint{4.233470in}{1.017558in}}%
\pgfpathlineto{\pgfqpoint{4.171222in}{1.017558in}}%
\pgfpathlineto{\pgfqpoint{4.171222in}{0.955310in}}%
\pgfusepath{fill}%
\end{pgfscope}%
\begin{pgfscope}%
\pgfpathrectangle{\pgfqpoint{4.171222in}{0.395076in}}{\pgfqpoint{0.062248in}{1.244963in}}%
\pgfusepath{clip}%
\pgfsetbuttcap%
\pgfsetroundjoin%
\definecolor{currentfill}{rgb}{0.905882,0.729412,0.321569}%
\pgfsetfillcolor{currentfill}%
\pgfsetlinewidth{0.000000pt}%
\definecolor{currentstroke}{rgb}{0.000000,0.000000,0.000000}%
\pgfsetstrokecolor{currentstroke}%
\pgfsetdash{}{0pt}%
\pgfpathmoveto{\pgfqpoint{4.171222in}{1.017558in}}%
\pgfpathlineto{\pgfqpoint{4.233470in}{1.017558in}}%
\pgfpathlineto{\pgfqpoint{4.233470in}{1.079806in}}%
\pgfpathlineto{\pgfqpoint{4.171222in}{1.079806in}}%
\pgfpathlineto{\pgfqpoint{4.171222in}{1.017558in}}%
\pgfusepath{fill}%
\end{pgfscope}%
\begin{pgfscope}%
\pgfpathrectangle{\pgfqpoint{4.171222in}{0.395076in}}{\pgfqpoint{0.062248in}{1.244963in}}%
\pgfusepath{clip}%
\pgfsetbuttcap%
\pgfsetroundjoin%
\definecolor{currentfill}{rgb}{0.905882,0.796078,0.580392}%
\pgfsetfillcolor{currentfill}%
\pgfsetlinewidth{0.000000pt}%
\definecolor{currentstroke}{rgb}{0.000000,0.000000,0.000000}%
\pgfsetstrokecolor{currentstroke}%
\pgfsetdash{}{0pt}%
\pgfpathmoveto{\pgfqpoint{4.171222in}{1.079806in}}%
\pgfpathlineto{\pgfqpoint{4.233470in}{1.079806in}}%
\pgfpathlineto{\pgfqpoint{4.233470in}{1.142054in}}%
\pgfpathlineto{\pgfqpoint{4.171222in}{1.142054in}}%
\pgfpathlineto{\pgfqpoint{4.171222in}{1.079806in}}%
\pgfusepath{fill}%
\end{pgfscope}%
\begin{pgfscope}%
\pgfpathrectangle{\pgfqpoint{4.171222in}{0.395076in}}{\pgfqpoint{0.062248in}{1.244963in}}%
\pgfusepath{clip}%
\pgfsetbuttcap%
\pgfsetroundjoin%
\definecolor{currentfill}{rgb}{0.517647,0.235294,0.223529}%
\pgfsetfillcolor{currentfill}%
\pgfsetlinewidth{0.000000pt}%
\definecolor{currentstroke}{rgb}{0.000000,0.000000,0.000000}%
\pgfsetstrokecolor{currentstroke}%
\pgfsetdash{}{0pt}%
\pgfpathmoveto{\pgfqpoint{4.171222in}{1.142054in}}%
\pgfpathlineto{\pgfqpoint{4.233470in}{1.142054in}}%
\pgfpathlineto{\pgfqpoint{4.233470in}{1.204302in}}%
\pgfpathlineto{\pgfqpoint{4.171222in}{1.204302in}}%
\pgfpathlineto{\pgfqpoint{4.171222in}{1.142054in}}%
\pgfusepath{fill}%
\end{pgfscope}%
\begin{pgfscope}%
\pgfpathrectangle{\pgfqpoint{4.171222in}{0.395076in}}{\pgfqpoint{0.062248in}{1.244963in}}%
\pgfusepath{clip}%
\pgfsetbuttcap%
\pgfsetroundjoin%
\definecolor{currentfill}{rgb}{0.678431,0.286275,0.290196}%
\pgfsetfillcolor{currentfill}%
\pgfsetlinewidth{0.000000pt}%
\definecolor{currentstroke}{rgb}{0.000000,0.000000,0.000000}%
\pgfsetstrokecolor{currentstroke}%
\pgfsetdash{}{0pt}%
\pgfpathmoveto{\pgfqpoint{4.171222in}{1.204302in}}%
\pgfpathlineto{\pgfqpoint{4.233470in}{1.204302in}}%
\pgfpathlineto{\pgfqpoint{4.233470in}{1.266550in}}%
\pgfpathlineto{\pgfqpoint{4.171222in}{1.266550in}}%
\pgfpathlineto{\pgfqpoint{4.171222in}{1.204302in}}%
\pgfusepath{fill}%
\end{pgfscope}%
\begin{pgfscope}%
\pgfpathrectangle{\pgfqpoint{4.171222in}{0.395076in}}{\pgfqpoint{0.062248in}{1.244963in}}%
\pgfusepath{clip}%
\pgfsetbuttcap%
\pgfsetroundjoin%
\definecolor{currentfill}{rgb}{0.839216,0.380392,0.419608}%
\pgfsetfillcolor{currentfill}%
\pgfsetlinewidth{0.000000pt}%
\definecolor{currentstroke}{rgb}{0.000000,0.000000,0.000000}%
\pgfsetstrokecolor{currentstroke}%
\pgfsetdash{}{0pt}%
\pgfpathmoveto{\pgfqpoint{4.171222in}{1.266550in}}%
\pgfpathlineto{\pgfqpoint{4.233470in}{1.266550in}}%
\pgfpathlineto{\pgfqpoint{4.233470in}{1.328798in}}%
\pgfpathlineto{\pgfqpoint{4.171222in}{1.328798in}}%
\pgfpathlineto{\pgfqpoint{4.171222in}{1.266550in}}%
\pgfusepath{fill}%
\end{pgfscope}%
\begin{pgfscope}%
\pgfpathrectangle{\pgfqpoint{4.171222in}{0.395076in}}{\pgfqpoint{0.062248in}{1.244963in}}%
\pgfusepath{clip}%
\pgfsetbuttcap%
\pgfsetroundjoin%
\definecolor{currentfill}{rgb}{0.905882,0.588235,0.611765}%
\pgfsetfillcolor{currentfill}%
\pgfsetlinewidth{0.000000pt}%
\definecolor{currentstroke}{rgb}{0.000000,0.000000,0.000000}%
\pgfsetstrokecolor{currentstroke}%
\pgfsetdash{}{0pt}%
\pgfpathmoveto{\pgfqpoint{4.171222in}{1.328798in}}%
\pgfpathlineto{\pgfqpoint{4.233470in}{1.328798in}}%
\pgfpathlineto{\pgfqpoint{4.233470in}{1.391046in}}%
\pgfpathlineto{\pgfqpoint{4.171222in}{1.391046in}}%
\pgfpathlineto{\pgfqpoint{4.171222in}{1.328798in}}%
\pgfusepath{fill}%
\end{pgfscope}%
\begin{pgfscope}%
\pgfpathrectangle{\pgfqpoint{4.171222in}{0.395076in}}{\pgfqpoint{0.062248in}{1.244963in}}%
\pgfusepath{clip}%
\pgfsetbuttcap%
\pgfsetroundjoin%
\definecolor{currentfill}{rgb}{0.482353,0.254902,0.450980}%
\pgfsetfillcolor{currentfill}%
\pgfsetlinewidth{0.000000pt}%
\definecolor{currentstroke}{rgb}{0.000000,0.000000,0.000000}%
\pgfsetstrokecolor{currentstroke}%
\pgfsetdash{}{0pt}%
\pgfpathmoveto{\pgfqpoint{4.171222in}{1.391046in}}%
\pgfpathlineto{\pgfqpoint{4.233470in}{1.391046in}}%
\pgfpathlineto{\pgfqpoint{4.233470in}{1.453295in}}%
\pgfpathlineto{\pgfqpoint{4.171222in}{1.453295in}}%
\pgfpathlineto{\pgfqpoint{4.171222in}{1.391046in}}%
\pgfusepath{fill}%
\end{pgfscope}%
\begin{pgfscope}%
\pgfpathrectangle{\pgfqpoint{4.171222in}{0.395076in}}{\pgfqpoint{0.062248in}{1.244963in}}%
\pgfusepath{clip}%
\pgfsetbuttcap%
\pgfsetroundjoin%
\definecolor{currentfill}{rgb}{0.647059,0.317647,0.580392}%
\pgfsetfillcolor{currentfill}%
\pgfsetlinewidth{0.000000pt}%
\definecolor{currentstroke}{rgb}{0.000000,0.000000,0.000000}%
\pgfsetstrokecolor{currentstroke}%
\pgfsetdash{}{0pt}%
\pgfpathmoveto{\pgfqpoint{4.171222in}{1.453295in}}%
\pgfpathlineto{\pgfqpoint{4.233470in}{1.453295in}}%
\pgfpathlineto{\pgfqpoint{4.233470in}{1.515543in}}%
\pgfpathlineto{\pgfqpoint{4.171222in}{1.515543in}}%
\pgfpathlineto{\pgfqpoint{4.171222in}{1.453295in}}%
\pgfusepath{fill}%
\end{pgfscope}%
\begin{pgfscope}%
\pgfpathrectangle{\pgfqpoint{4.171222in}{0.395076in}}{\pgfqpoint{0.062248in}{1.244963in}}%
\pgfusepath{clip}%
\pgfsetbuttcap%
\pgfsetroundjoin%
\definecolor{currentfill}{rgb}{0.807843,0.427451,0.741176}%
\pgfsetfillcolor{currentfill}%
\pgfsetlinewidth{0.000000pt}%
\definecolor{currentstroke}{rgb}{0.000000,0.000000,0.000000}%
\pgfsetstrokecolor{currentstroke}%
\pgfsetdash{}{0pt}%
\pgfpathmoveto{\pgfqpoint{4.171222in}{1.515543in}}%
\pgfpathlineto{\pgfqpoint{4.233470in}{1.515543in}}%
\pgfpathlineto{\pgfqpoint{4.233470in}{1.577791in}}%
\pgfpathlineto{\pgfqpoint{4.171222in}{1.577791in}}%
\pgfpathlineto{\pgfqpoint{4.171222in}{1.515543in}}%
\pgfusepath{fill}%
\end{pgfscope}%
\begin{pgfscope}%
\pgfpathrectangle{\pgfqpoint{4.171222in}{0.395076in}}{\pgfqpoint{0.062248in}{1.244963in}}%
\pgfusepath{clip}%
\pgfsetbuttcap%
\pgfsetroundjoin%
\definecolor{currentfill}{rgb}{0.870588,0.619608,0.839216}%
\pgfsetfillcolor{currentfill}%
\pgfsetlinewidth{0.000000pt}%
\definecolor{currentstroke}{rgb}{0.000000,0.000000,0.000000}%
\pgfsetstrokecolor{currentstroke}%
\pgfsetdash{}{0pt}%
\pgfpathmoveto{\pgfqpoint{4.171222in}{1.577791in}}%
\pgfpathlineto{\pgfqpoint{4.233470in}{1.577791in}}%
\pgfpathlineto{\pgfqpoint{4.233470in}{1.640039in}}%
\pgfpathlineto{\pgfqpoint{4.171222in}{1.640039in}}%
\pgfpathlineto{\pgfqpoint{4.171222in}{1.577791in}}%
\pgfusepath{fill}%
\end{pgfscope}%
\begin{pgfscope}%
\pgfsetbuttcap%
\pgfsetroundjoin%
\definecolor{currentfill}{rgb}{0.000000,0.000000,0.000000}%
\pgfsetfillcolor{currentfill}%
\pgfsetlinewidth{0.803000pt}%
\definecolor{currentstroke}{rgb}{0.000000,0.000000,0.000000}%
\pgfsetstrokecolor{currentstroke}%
\pgfsetdash{}{0pt}%
\pgfsys@defobject{currentmarker}{\pgfqpoint{0.000000in}{0.000000in}}{\pgfqpoint{0.048611in}{0.000000in}}{%
\pgfpathmoveto{\pgfqpoint{0.000000in}{0.000000in}}%
\pgfpathlineto{\pgfqpoint{0.048611in}{0.000000in}}%
\pgfusepath{stroke,fill}%
}%
\begin{pgfscope}%
\pgfsys@transformshift{4.233470in}{0.670122in}%
\pgfsys@useobject{currentmarker}{}%
\end{pgfscope}%
\end{pgfscope}%
\begin{pgfscope}%
\definecolor{textcolor}{rgb}{0.000000,0.000000,0.000000}%
\pgfsetstrokecolor{textcolor}%
\pgfsetfillcolor{textcolor}%
\pgftext[x=4.330693in, y=0.612084in, left, base]{\color{textcolor}\sffamily\fontsize{11.000000}{13.200000}\selectfont \ensuremath{-}5}%
\end{pgfscope}%
\begin{pgfscope}%
\pgfsetbuttcap%
\pgfsetroundjoin%
\definecolor{currentfill}{rgb}{0.000000,0.000000,0.000000}%
\pgfsetfillcolor{currentfill}%
\pgfsetlinewidth{0.803000pt}%
\definecolor{currentstroke}{rgb}{0.000000,0.000000,0.000000}%
\pgfsetstrokecolor{currentstroke}%
\pgfsetdash{}{0pt}%
\pgfsys@defobject{currentmarker}{\pgfqpoint{0.000000in}{0.000000in}}{\pgfqpoint{0.048611in}{0.000000in}}{%
\pgfpathmoveto{\pgfqpoint{0.000000in}{0.000000in}}%
\pgfpathlineto{\pgfqpoint{0.048611in}{0.000000in}}%
\pgfusepath{stroke,fill}%
}%
\begin{pgfscope}%
\pgfsys@transformshift{4.233470in}{1.230355in}%
\pgfsys@useobject{currentmarker}{}%
\end{pgfscope}%
\end{pgfscope}%
\begin{pgfscope}%
\definecolor{textcolor}{rgb}{0.000000,0.000000,0.000000}%
\pgfsetstrokecolor{textcolor}%
\pgfsetfillcolor{textcolor}%
\pgftext[x=4.330693in, y=1.172318in, left, base]{\color{textcolor}\sffamily\fontsize{11.000000}{13.200000}\selectfont \ensuremath{-}4}%
\end{pgfscope}%
\begin{pgfscope}%
\definecolor{textcolor}{rgb}{0.000000,0.000000,0.000000}%
\pgfsetstrokecolor{textcolor}%
\pgfsetfillcolor{textcolor}%
\pgftext[x=4.601737in,y=1.017558in,,top,rotate=90.000000]{\color{textcolor}\sffamily\fontsize{12.000000}{14.400000}\selectfont \(\displaystyle log_{10}(\alpha \lambda)\)}%
\end{pgfscope}%
\begin{pgfscope}%
\pgfsetrectcap%
\pgfsetmiterjoin%
\pgfsetlinewidth{0.803000pt}%
\definecolor{currentstroke}{rgb}{0.000000,0.000000,0.000000}%
\pgfsetstrokecolor{currentstroke}%
\pgfsetdash{}{0pt}%
\pgfpathmoveto{\pgfqpoint{4.171222in}{0.395076in}}%
\pgfpathlineto{\pgfqpoint{4.202346in}{0.395076in}}%
\pgfpathlineto{\pgfqpoint{4.233470in}{0.395076in}}%
\pgfpathlineto{\pgfqpoint{4.233470in}{1.640039in}}%
\pgfpathlineto{\pgfqpoint{4.202346in}{1.640039in}}%
\pgfpathlineto{\pgfqpoint{4.171222in}{1.640039in}}%
\pgfpathlineto{\pgfqpoint{4.171222in}{0.395076in}}%
\pgfpathclose%
\pgfusepath{stroke}%
\end{pgfscope}%
\end{pgfpicture}%
\makeatother%
\endgroup%

%% file: figures/tlength_acc.pgf
\begingroup%
\makeatletter%
\begin{pgfpicture}%
\pgfpathrectangle{\pgfpointorigin}{\pgfqpoint{5.000000in}{3.000000in}}%
\pgfusepath{use as bounding box, clip}%
\begin{pgfscope}%
\pgfsetbuttcap%
\pgfsetmiterjoin%
\definecolor{currentfill}{rgb}{1.000000,1.000000,1.000000}%
\pgfsetfillcolor{currentfill}%
\pgfsetlinewidth{0.000000pt}%
\definecolor{currentstroke}{rgb}{1.000000,1.000000,1.000000}%
\pgfsetstrokecolor{currentstroke}%
\pgfsetdash{}{0pt}%
\pgfpathmoveto{\pgfqpoint{0.000000in}{0.000000in}}%
\pgfpathlineto{\pgfqpoint{5.000000in}{0.000000in}}%
\pgfpathlineto{\pgfqpoint{5.000000in}{3.000000in}}%
\pgfpathlineto{\pgfqpoint{0.000000in}{3.000000in}}%
\pgfpathlineto{\pgfqpoint{0.000000in}{0.000000in}}%
\pgfpathclose%
\pgfusepath{fill}%
\end{pgfscope}%
\begin{pgfscope}%
\pgfsetbuttcap%
\pgfsetmiterjoin%
\definecolor{currentfill}{rgb}{1.000000,1.000000,1.000000}%
\pgfsetfillcolor{currentfill}%
\pgfsetlinewidth{0.000000pt}%
\definecolor{currentstroke}{rgb}{0.000000,0.000000,0.000000}%
\pgfsetstrokecolor{currentstroke}%
\pgfsetstrokeopacity{0.000000}%
\pgfsetdash{}{0pt}%
\pgfpathmoveto{\pgfqpoint{0.658477in}{0.611927in}}%
\pgfpathlineto{\pgfqpoint{4.850000in}{0.611927in}}%
\pgfpathlineto{\pgfqpoint{4.850000in}{2.814431in}}%
\pgfpathlineto{\pgfqpoint{0.658477in}{2.814431in}}%
\pgfpathlineto{\pgfqpoint{0.658477in}{0.611927in}}%
\pgfpathclose%
\pgfusepath{fill}%
\end{pgfscope}%
\begin{pgfscope}%
\pgfsetbuttcap%
\pgfsetroundjoin%
\definecolor{currentfill}{rgb}{0.000000,0.000000,0.000000}%
\pgfsetfillcolor{currentfill}%
\pgfsetlinewidth{0.803000pt}%
\definecolor{currentstroke}{rgb}{0.000000,0.000000,0.000000}%
\pgfsetstrokecolor{currentstroke}%
\pgfsetdash{}{0pt}%
\pgfsys@defobject{currentmarker}{\pgfqpoint{0.000000in}{-0.048611in}}{\pgfqpoint{0.000000in}{0.000000in}}{%
\pgfpathmoveto{\pgfqpoint{0.000000in}{0.000000in}}%
\pgfpathlineto{\pgfqpoint{0.000000in}{-0.048611in}}%
\pgfusepath{stroke,fill}%
}%
\begin{pgfscope}%
\pgfsys@transformshift{0.849001in}{0.611927in}%
\pgfsys@useobject{currentmarker}{}%
\end{pgfscope}%
\end{pgfscope}%
\begin{pgfscope}%
\definecolor{textcolor}{rgb}{0.000000,0.000000,0.000000}%
\pgfsetstrokecolor{textcolor}%
\pgfsetfillcolor{textcolor}%
\pgftext[x=0.849001in,y=0.514705in,,top]{\color{textcolor}\sffamily\fontsize{11.000000}{13.200000}\selectfont 1}%
\end{pgfscope}%
\begin{pgfscope}%
\pgfsetbuttcap%
\pgfsetroundjoin%
\definecolor{currentfill}{rgb}{0.000000,0.000000,0.000000}%
\pgfsetfillcolor{currentfill}%
\pgfsetlinewidth{0.803000pt}%
\definecolor{currentstroke}{rgb}{0.000000,0.000000,0.000000}%
\pgfsetstrokecolor{currentstroke}%
\pgfsetdash{}{0pt}%
\pgfsys@defobject{currentmarker}{\pgfqpoint{0.000000in}{-0.048611in}}{\pgfqpoint{0.000000in}{0.000000in}}{%
\pgfpathmoveto{\pgfqpoint{0.000000in}{0.000000in}}%
\pgfpathlineto{\pgfqpoint{0.000000in}{-0.048611in}}%
\pgfusepath{stroke,fill}%
}%
\begin{pgfscope}%
\pgfsys@transformshift{2.119159in}{0.611927in}%
\pgfsys@useobject{currentmarker}{}%
\end{pgfscope}%
\end{pgfscope}%
\begin{pgfscope}%
\definecolor{textcolor}{rgb}{0.000000,0.000000,0.000000}%
\pgfsetstrokecolor{textcolor}%
\pgfsetfillcolor{textcolor}%
\pgftext[x=2.119159in,y=0.514705in,,top]{\color{textcolor}\sffamily\fontsize{11.000000}{13.200000}\selectfont 3}%
\end{pgfscope}%
\begin{pgfscope}%
\pgfsetbuttcap%
\pgfsetroundjoin%
\definecolor{currentfill}{rgb}{0.000000,0.000000,0.000000}%
\pgfsetfillcolor{currentfill}%
\pgfsetlinewidth{0.803000pt}%
\definecolor{currentstroke}{rgb}{0.000000,0.000000,0.000000}%
\pgfsetstrokecolor{currentstroke}%
\pgfsetdash{}{0pt}%
\pgfsys@defobject{currentmarker}{\pgfqpoint{0.000000in}{-0.048611in}}{\pgfqpoint{0.000000in}{0.000000in}}{%
\pgfpathmoveto{\pgfqpoint{0.000000in}{0.000000in}}%
\pgfpathlineto{\pgfqpoint{0.000000in}{-0.048611in}}%
\pgfusepath{stroke,fill}%
}%
\begin{pgfscope}%
\pgfsys@transformshift{3.389318in}{0.611927in}%
\pgfsys@useobject{currentmarker}{}%
\end{pgfscope}%
\end{pgfscope}%
\begin{pgfscope}%
\definecolor{textcolor}{rgb}{0.000000,0.000000,0.000000}%
\pgfsetstrokecolor{textcolor}%
\pgfsetfillcolor{textcolor}%
\pgftext[x=3.389318in,y=0.514705in,,top]{\color{textcolor}\sffamily\fontsize{11.000000}{13.200000}\selectfont 10}%
\end{pgfscope}%
\begin{pgfscope}%
\pgfsetbuttcap%
\pgfsetroundjoin%
\definecolor{currentfill}{rgb}{0.000000,0.000000,0.000000}%
\pgfsetfillcolor{currentfill}%
\pgfsetlinewidth{0.803000pt}%
\definecolor{currentstroke}{rgb}{0.000000,0.000000,0.000000}%
\pgfsetstrokecolor{currentstroke}%
\pgfsetdash{}{0pt}%
\pgfsys@defobject{currentmarker}{\pgfqpoint{0.000000in}{-0.048611in}}{\pgfqpoint{0.000000in}{0.000000in}}{%
\pgfpathmoveto{\pgfqpoint{0.000000in}{0.000000in}}%
\pgfpathlineto{\pgfqpoint{0.000000in}{-0.048611in}}%
\pgfusepath{stroke,fill}%
}%
\begin{pgfscope}%
\pgfsys@transformshift{4.659476in}{0.611927in}%
\pgfsys@useobject{currentmarker}{}%
\end{pgfscope}%
\end{pgfscope}%
\begin{pgfscope}%
\definecolor{textcolor}{rgb}{0.000000,0.000000,0.000000}%
\pgfsetstrokecolor{textcolor}%
\pgfsetfillcolor{textcolor}%
\pgftext[x=4.659476in,y=0.514705in,,top]{\color{textcolor}\sffamily\fontsize{11.000000}{13.200000}\selectfont 22}%
\end{pgfscope}%
\begin{pgfscope}%
\definecolor{textcolor}{rgb}{0.000000,0.000000,0.000000}%
\pgfsetstrokecolor{textcolor}%
\pgfsetfillcolor{textcolor}%
\pgftext[x=2.754238in,y=0.311295in,,top]{\color{textcolor}\sffamily\fontsize{12.000000}{14.400000}\selectfont WRN-16-(x)}%
\end{pgfscope}%
\begin{pgfscope}%
\pgfsetbuttcap%
\pgfsetroundjoin%
\definecolor{currentfill}{rgb}{0.000000,0.000000,0.000000}%
\pgfsetfillcolor{currentfill}%
\pgfsetlinewidth{0.803000pt}%
\definecolor{currentstroke}{rgb}{0.000000,0.000000,0.000000}%
\pgfsetstrokecolor{currentstroke}%
\pgfsetdash{}{0pt}%
\pgfsys@defobject{currentmarker}{\pgfqpoint{-0.048611in}{0.000000in}}{\pgfqpoint{-0.000000in}{0.000000in}}{%
\pgfpathmoveto{\pgfqpoint{-0.000000in}{0.000000in}}%
\pgfpathlineto{\pgfqpoint{-0.048611in}{0.000000in}}%
\pgfusepath{stroke,fill}%
}%
\begin{pgfscope}%
\pgfsys@transformshift{0.658477in}{0.618558in}%
\pgfsys@useobject{currentmarker}{}%
\end{pgfscope}%
\end{pgfscope}%
\begin{pgfscope}%
\definecolor{textcolor}{rgb}{0.000000,0.000000,0.000000}%
\pgfsetstrokecolor{textcolor}%
\pgfsetfillcolor{textcolor}%
\pgftext[x=0.366851in, y=0.560520in, left, base]{\color{textcolor}\sffamily\fontsize{11.000000}{13.200000}\selectfont 58}%
\end{pgfscope}%
\begin{pgfscope}%
\pgfsetbuttcap%
\pgfsetroundjoin%
\definecolor{currentfill}{rgb}{0.000000,0.000000,0.000000}%
\pgfsetfillcolor{currentfill}%
\pgfsetlinewidth{0.803000pt}%
\definecolor{currentstroke}{rgb}{0.000000,0.000000,0.000000}%
\pgfsetstrokecolor{currentstroke}%
\pgfsetdash{}{0pt}%
\pgfsys@defobject{currentmarker}{\pgfqpoint{-0.048611in}{0.000000in}}{\pgfqpoint{-0.000000in}{0.000000in}}{%
\pgfpathmoveto{\pgfqpoint{-0.000000in}{0.000000in}}%
\pgfpathlineto{\pgfqpoint{-0.048611in}{0.000000in}}%
\pgfusepath{stroke,fill}%
}%
\begin{pgfscope}%
\pgfsys@transformshift{0.658477in}{1.053363in}%
\pgfsys@useobject{currentmarker}{}%
\end{pgfscope}%
\end{pgfscope}%
\begin{pgfscope}%
\definecolor{textcolor}{rgb}{0.000000,0.000000,0.000000}%
\pgfsetstrokecolor{textcolor}%
\pgfsetfillcolor{textcolor}%
\pgftext[x=0.366851in, y=0.995325in, left, base]{\color{textcolor}\sffamily\fontsize{11.000000}{13.200000}\selectfont 60}%
\end{pgfscope}%
\begin{pgfscope}%
\pgfsetbuttcap%
\pgfsetroundjoin%
\definecolor{currentfill}{rgb}{0.000000,0.000000,0.000000}%
\pgfsetfillcolor{currentfill}%
\pgfsetlinewidth{0.803000pt}%
\definecolor{currentstroke}{rgb}{0.000000,0.000000,0.000000}%
\pgfsetstrokecolor{currentstroke}%
\pgfsetdash{}{0pt}%
\pgfsys@defobject{currentmarker}{\pgfqpoint{-0.048611in}{0.000000in}}{\pgfqpoint{-0.000000in}{0.000000in}}{%
\pgfpathmoveto{\pgfqpoint{-0.000000in}{0.000000in}}%
\pgfpathlineto{\pgfqpoint{-0.048611in}{0.000000in}}%
\pgfusepath{stroke,fill}%
}%
\begin{pgfscope}%
\pgfsys@transformshift{0.658477in}{1.488168in}%
\pgfsys@useobject{currentmarker}{}%
\end{pgfscope}%
\end{pgfscope}%
\begin{pgfscope}%
\definecolor{textcolor}{rgb}{0.000000,0.000000,0.000000}%
\pgfsetstrokecolor{textcolor}%
\pgfsetfillcolor{textcolor}%
\pgftext[x=0.366851in, y=1.430130in, left, base]{\color{textcolor}\sffamily\fontsize{11.000000}{13.200000}\selectfont 62}%
\end{pgfscope}%
\begin{pgfscope}%
\pgfsetbuttcap%
\pgfsetroundjoin%
\definecolor{currentfill}{rgb}{0.000000,0.000000,0.000000}%
\pgfsetfillcolor{currentfill}%
\pgfsetlinewidth{0.803000pt}%
\definecolor{currentstroke}{rgb}{0.000000,0.000000,0.000000}%
\pgfsetstrokecolor{currentstroke}%
\pgfsetdash{}{0pt}%
\pgfsys@defobject{currentmarker}{\pgfqpoint{-0.048611in}{0.000000in}}{\pgfqpoint{-0.000000in}{0.000000in}}{%
\pgfpathmoveto{\pgfqpoint{-0.000000in}{0.000000in}}%
\pgfpathlineto{\pgfqpoint{-0.048611in}{0.000000in}}%
\pgfusepath{stroke,fill}%
}%
\begin{pgfscope}%
\pgfsys@transformshift{0.658477in}{1.922973in}%
\pgfsys@useobject{currentmarker}{}%
\end{pgfscope}%
\end{pgfscope}%
\begin{pgfscope}%
\definecolor{textcolor}{rgb}{0.000000,0.000000,0.000000}%
\pgfsetstrokecolor{textcolor}%
\pgfsetfillcolor{textcolor}%
\pgftext[x=0.366851in, y=1.864935in, left, base]{\color{textcolor}\sffamily\fontsize{11.000000}{13.200000}\selectfont 64}%
\end{pgfscope}%
\begin{pgfscope}%
\pgfsetbuttcap%
\pgfsetroundjoin%
\definecolor{currentfill}{rgb}{0.000000,0.000000,0.000000}%
\pgfsetfillcolor{currentfill}%
\pgfsetlinewidth{0.803000pt}%
\definecolor{currentstroke}{rgb}{0.000000,0.000000,0.000000}%
\pgfsetstrokecolor{currentstroke}%
\pgfsetdash{}{0pt}%
\pgfsys@defobject{currentmarker}{\pgfqpoint{-0.048611in}{0.000000in}}{\pgfqpoint{-0.000000in}{0.000000in}}{%
\pgfpathmoveto{\pgfqpoint{-0.000000in}{0.000000in}}%
\pgfpathlineto{\pgfqpoint{-0.048611in}{0.000000in}}%
\pgfusepath{stroke,fill}%
}%
\begin{pgfscope}%
\pgfsys@transformshift{0.658477in}{2.357777in}%
\pgfsys@useobject{currentmarker}{}%
\end{pgfscope}%
\end{pgfscope}%
\begin{pgfscope}%
\definecolor{textcolor}{rgb}{0.000000,0.000000,0.000000}%
\pgfsetstrokecolor{textcolor}%
\pgfsetfillcolor{textcolor}%
\pgftext[x=0.366851in, y=2.299740in, left, base]{\color{textcolor}\sffamily\fontsize{11.000000}{13.200000}\selectfont 66}%
\end{pgfscope}%
\begin{pgfscope}%
\pgfsetbuttcap%
\pgfsetroundjoin%
\definecolor{currentfill}{rgb}{0.000000,0.000000,0.000000}%
\pgfsetfillcolor{currentfill}%
\pgfsetlinewidth{0.803000pt}%
\definecolor{currentstroke}{rgb}{0.000000,0.000000,0.000000}%
\pgfsetstrokecolor{currentstroke}%
\pgfsetdash{}{0pt}%
\pgfsys@defobject{currentmarker}{\pgfqpoint{-0.048611in}{0.000000in}}{\pgfqpoint{-0.000000in}{0.000000in}}{%
\pgfpathmoveto{\pgfqpoint{-0.000000in}{0.000000in}}%
\pgfpathlineto{\pgfqpoint{-0.048611in}{0.000000in}}%
\pgfusepath{stroke,fill}%
}%
\begin{pgfscope}%
\pgfsys@transformshift{0.658477in}{2.792582in}%
\pgfsys@useobject{currentmarker}{}%
\end{pgfscope}%
\end{pgfscope}%
\begin{pgfscope}%
\definecolor{textcolor}{rgb}{0.000000,0.000000,0.000000}%
\pgfsetstrokecolor{textcolor}%
\pgfsetfillcolor{textcolor}%
\pgftext[x=0.366851in, y=2.734545in, left, base]{\color{textcolor}\sffamily\fontsize{11.000000}{13.200000}\selectfont 68}%
\end{pgfscope}%
\begin{pgfscope}%
\definecolor{textcolor}{rgb}{0.000000,0.000000,0.000000}%
\pgfsetstrokecolor{textcolor}%
\pgfsetfillcolor{textcolor}%
\pgftext[x=0.311295in,y=1.713179in,,bottom,rotate=90.000000]{\color{textcolor}\sffamily\fontsize{12.000000}{14.400000}\selectfont Max Test Accuracy (\%)}%
\end{pgfscope}%
\begin{pgfscope}%
\pgfpathrectangle{\pgfqpoint{0.658477in}{0.611927in}}{\pgfqpoint{4.191523in}{2.202504in}}%
\pgfusepath{clip}%
\pgfsetrectcap%
\pgfsetroundjoin%
\pgfsetlinewidth{2.509375pt}%
\definecolor{currentstroke}{rgb}{0.121569,0.466667,0.705882}%
\pgfsetstrokecolor{currentstroke}%
\pgfsetdash{}{0pt}%
\pgfpathmoveto{\pgfqpoint{0.849001in}{0.712041in}}%
\pgfpathlineto{\pgfqpoint{2.119159in}{1.664264in}}%
\pgfpathlineto{\pgfqpoint{3.389318in}{2.177333in}}%
\pgfpathlineto{\pgfqpoint{4.659476in}{2.249076in}}%
\pgfusepath{stroke}%
\end{pgfscope}%
\begin{pgfscope}%
\pgfpathrectangle{\pgfqpoint{0.658477in}{0.611927in}}{\pgfqpoint{4.191523in}{2.202504in}}%
\pgfusepath{clip}%
\pgfsetbuttcap%
\pgfsetroundjoin%
\definecolor{currentfill}{rgb}{0.121569,0.466667,0.705882}%
\pgfsetfillcolor{currentfill}%
\pgfsetlinewidth{1.003750pt}%
\definecolor{currentstroke}{rgb}{0.121569,0.466667,0.705882}%
\pgfsetstrokecolor{currentstroke}%
\pgfsetdash{}{0pt}%
\pgfsys@defobject{currentmarker}{\pgfqpoint{-0.055556in}{-0.055556in}}{\pgfqpoint{0.055556in}{0.055556in}}{%
\pgfpathmoveto{\pgfqpoint{0.000000in}{-0.055556in}}%
\pgfpathcurveto{\pgfqpoint{0.014734in}{-0.055556in}}{\pgfqpoint{0.028866in}{-0.049702in}}{\pgfqpoint{0.039284in}{-0.039284in}}%
\pgfpathcurveto{\pgfqpoint{0.049702in}{-0.028866in}}{\pgfqpoint{0.055556in}{-0.014734in}}{\pgfqpoint{0.055556in}{0.000000in}}%
\pgfpathcurveto{\pgfqpoint{0.055556in}{0.014734in}}{\pgfqpoint{0.049702in}{0.028866in}}{\pgfqpoint{0.039284in}{0.039284in}}%
\pgfpathcurveto{\pgfqpoint{0.028866in}{0.049702in}}{\pgfqpoint{0.014734in}{0.055556in}}{\pgfqpoint{0.000000in}{0.055556in}}%
\pgfpathcurveto{\pgfqpoint{-0.014734in}{0.055556in}}{\pgfqpoint{-0.028866in}{0.049702in}}{\pgfqpoint{-0.039284in}{0.039284in}}%
\pgfpathcurveto{\pgfqpoint{-0.049702in}{0.028866in}}{\pgfqpoint{-0.055556in}{0.014734in}}{\pgfqpoint{-0.055556in}{0.000000in}}%
\pgfpathcurveto{\pgfqpoint{-0.055556in}{-0.014734in}}{\pgfqpoint{-0.049702in}{-0.028866in}}{\pgfqpoint{-0.039284in}{-0.039284in}}%
\pgfpathcurveto{\pgfqpoint{-0.028866in}{-0.049702in}}{\pgfqpoint{-0.014734in}{-0.055556in}}{\pgfqpoint{0.000000in}{-0.055556in}}%
\pgfpathlineto{\pgfqpoint{0.000000in}{-0.055556in}}%
\pgfpathclose%
\pgfusepath{stroke,fill}%
}%
\begin{pgfscope}%
\pgfsys@transformshift{0.849001in}{0.712041in}%
\pgfsys@useobject{currentmarker}{}%
\end{pgfscope}%
\begin{pgfscope}%
\pgfsys@transformshift{2.119159in}{1.664264in}%
\pgfsys@useobject{currentmarker}{}%
\end{pgfscope}%
\begin{pgfscope}%
\pgfsys@transformshift{3.389318in}{2.177333in}%
\pgfsys@useobject{currentmarker}{}%
\end{pgfscope}%
\begin{pgfscope}%
\pgfsys@transformshift{4.659476in}{2.249076in}%
\pgfsys@useobject{currentmarker}{}%
\end{pgfscope}%
\end{pgfscope}%
\begin{pgfscope}%
\pgfpathrectangle{\pgfqpoint{0.658477in}{0.611927in}}{\pgfqpoint{4.191523in}{2.202504in}}%
\pgfusepath{clip}%
\pgfsetrectcap%
\pgfsetroundjoin%
\pgfsetlinewidth{2.509375pt}%
\definecolor{currentstroke}{rgb}{1.000000,0.498039,0.054902}%
\pgfsetstrokecolor{currentstroke}%
\pgfsetdash{}{0pt}%
\pgfpathmoveto{\pgfqpoint{0.849001in}{1.579477in}}%
\pgfpathlineto{\pgfqpoint{2.119159in}{1.979497in}}%
\pgfpathlineto{\pgfqpoint{3.389318in}{2.523003in}}%
\pgfpathlineto{\pgfqpoint{4.659476in}{2.714317in}}%
\pgfusepath{stroke}%
\end{pgfscope}%
\begin{pgfscope}%
\pgfpathrectangle{\pgfqpoint{0.658477in}{0.611927in}}{\pgfqpoint{4.191523in}{2.202504in}}%
\pgfusepath{clip}%
\pgfsetbuttcap%
\pgfsetroundjoin%
\definecolor{currentfill}{rgb}{1.000000,0.498039,0.054902}%
\pgfsetfillcolor{currentfill}%
\pgfsetlinewidth{1.003750pt}%
\definecolor{currentstroke}{rgb}{1.000000,0.498039,0.054902}%
\pgfsetstrokecolor{currentstroke}%
\pgfsetdash{}{0pt}%
\pgfsys@defobject{currentmarker}{\pgfqpoint{-0.055556in}{-0.055556in}}{\pgfqpoint{0.055556in}{0.055556in}}{%
\pgfpathmoveto{\pgfqpoint{0.000000in}{-0.055556in}}%
\pgfpathcurveto{\pgfqpoint{0.014734in}{-0.055556in}}{\pgfqpoint{0.028866in}{-0.049702in}}{\pgfqpoint{0.039284in}{-0.039284in}}%
\pgfpathcurveto{\pgfqpoint{0.049702in}{-0.028866in}}{\pgfqpoint{0.055556in}{-0.014734in}}{\pgfqpoint{0.055556in}{0.000000in}}%
\pgfpathcurveto{\pgfqpoint{0.055556in}{0.014734in}}{\pgfqpoint{0.049702in}{0.028866in}}{\pgfqpoint{0.039284in}{0.039284in}}%
\pgfpathcurveto{\pgfqpoint{0.028866in}{0.049702in}}{\pgfqpoint{0.014734in}{0.055556in}}{\pgfqpoint{0.000000in}{0.055556in}}%
\pgfpathcurveto{\pgfqpoint{-0.014734in}{0.055556in}}{\pgfqpoint{-0.028866in}{0.049702in}}{\pgfqpoint{-0.039284in}{0.039284in}}%
\pgfpathcurveto{\pgfqpoint{-0.049702in}{0.028866in}}{\pgfqpoint{-0.055556in}{0.014734in}}{\pgfqpoint{-0.055556in}{0.000000in}}%
\pgfpathcurveto{\pgfqpoint{-0.055556in}{-0.014734in}}{\pgfqpoint{-0.049702in}{-0.028866in}}{\pgfqpoint{-0.039284in}{-0.039284in}}%
\pgfpathcurveto{\pgfqpoint{-0.028866in}{-0.049702in}}{\pgfqpoint{-0.014734in}{-0.055556in}}{\pgfqpoint{0.000000in}{-0.055556in}}%
\pgfpathlineto{\pgfqpoint{0.000000in}{-0.055556in}}%
\pgfpathclose%
\pgfusepath{stroke,fill}%
}%
\begin{pgfscope}%
\pgfsys@transformshift{0.849001in}{1.579477in}%
\pgfsys@useobject{currentmarker}{}%
\end{pgfscope}%
\begin{pgfscope}%
\pgfsys@transformshift{2.119159in}{1.979497in}%
\pgfsys@useobject{currentmarker}{}%
\end{pgfscope}%
\begin{pgfscope}%
\pgfsys@transformshift{3.389318in}{2.523003in}%
\pgfsys@useobject{currentmarker}{}%
\end{pgfscope}%
\begin{pgfscope}%
\pgfsys@transformshift{4.659476in}{2.714317in}%
\pgfsys@useobject{currentmarker}{}%
\end{pgfscope}%
\end{pgfscope}%
\begin{pgfscope}%
\pgfsetrectcap%
\pgfsetmiterjoin%
\pgfsetlinewidth{1.505625pt}%
\definecolor{currentstroke}{rgb}{0.000000,0.000000,0.000000}%
\pgfsetstrokecolor{currentstroke}%
\pgfsetdash{}{0pt}%
\pgfpathmoveto{\pgfqpoint{0.658477in}{0.611927in}}%
\pgfpathlineto{\pgfqpoint{0.658477in}{2.814431in}}%
\pgfusepath{stroke}%
\end{pgfscope}%
\begin{pgfscope}%
\pgfsetrectcap%
\pgfsetmiterjoin%
\pgfsetlinewidth{1.505625pt}%
\definecolor{currentstroke}{rgb}{0.000000,0.000000,0.000000}%
\pgfsetstrokecolor{currentstroke}%
\pgfsetdash{}{0pt}%
\pgfpathmoveto{\pgfqpoint{0.658477in}{0.611927in}}%
\pgfpathlineto{\pgfqpoint{4.850000in}{0.611927in}}%
\pgfusepath{stroke}%
\end{pgfscope}%
\begin{pgfscope}%
\pgfsetbuttcap%
\pgfsetmiterjoin%
\definecolor{currentfill}{rgb}{1.000000,1.000000,1.000000}%
\pgfsetfillcolor{currentfill}%
\pgfsetfillopacity{0.800000}%
\pgfsetlinewidth{1.003750pt}%
\definecolor{currentstroke}{rgb}{0.800000,0.800000,0.800000}%
\pgfsetstrokecolor{currentstroke}%
\pgfsetstrokeopacity{0.800000}%
\pgfsetdash{}{0pt}%
\pgfpathmoveto{\pgfqpoint{3.468880in}{0.695261in}}%
\pgfpathlineto{\pgfqpoint{4.733333in}{0.695261in}}%
\pgfpathquadraticcurveto{\pgfqpoint{4.766667in}{0.695261in}}{\pgfqpoint{4.766667in}{0.728594in}}%
\pgfpathlineto{\pgfqpoint{4.766667in}{1.201185in}}%
\pgfpathquadraticcurveto{\pgfqpoint{4.766667in}{1.234518in}}{\pgfqpoint{4.733333in}{1.234518in}}%
\pgfpathlineto{\pgfqpoint{3.468880in}{1.234518in}}%
\pgfpathquadraticcurveto{\pgfqpoint{3.435547in}{1.234518in}}{\pgfqpoint{3.435547in}{1.201185in}}%
\pgfpathlineto{\pgfqpoint{3.435547in}{0.728594in}}%
\pgfpathquadraticcurveto{\pgfqpoint{3.435547in}{0.695261in}}{\pgfqpoint{3.468880in}{0.695261in}}%
\pgfpathlineto{\pgfqpoint{3.468880in}{0.695261in}}%
\pgfpathclose%
\pgfusepath{stroke,fill}%
\end{pgfscope}%
\begin{pgfscope}%
\pgfsetrectcap%
\pgfsetroundjoin%
\pgfsetlinewidth{2.509375pt}%
\definecolor{currentstroke}{rgb}{0.121569,0.466667,0.705882}%
\pgfsetstrokecolor{currentstroke}%
\pgfsetdash{}{0pt}%
\pgfpathmoveto{\pgfqpoint{3.502213in}{1.099557in}}%
\pgfpathlineto{\pgfqpoint{3.668880in}{1.099557in}}%
\pgfpathlineto{\pgfqpoint{3.835547in}{1.099557in}}%
\pgfusepath{stroke}%
\end{pgfscope}%
\begin{pgfscope}%
\pgfsetbuttcap%
\pgfsetroundjoin%
\definecolor{currentfill}{rgb}{0.121569,0.466667,0.705882}%
\pgfsetfillcolor{currentfill}%
\pgfsetlinewidth{1.003750pt}%
\definecolor{currentstroke}{rgb}{0.121569,0.466667,0.705882}%
\pgfsetstrokecolor{currentstroke}%
\pgfsetdash{}{0pt}%
\pgfsys@defobject{currentmarker}{\pgfqpoint{-0.055556in}{-0.055556in}}{\pgfqpoint{0.055556in}{0.055556in}}{%
\pgfpathmoveto{\pgfqpoint{0.000000in}{-0.055556in}}%
\pgfpathcurveto{\pgfqpoint{0.014734in}{-0.055556in}}{\pgfqpoint{0.028866in}{-0.049702in}}{\pgfqpoint{0.039284in}{-0.039284in}}%
\pgfpathcurveto{\pgfqpoint{0.049702in}{-0.028866in}}{\pgfqpoint{0.055556in}{-0.014734in}}{\pgfqpoint{0.055556in}{0.000000in}}%
\pgfpathcurveto{\pgfqpoint{0.055556in}{0.014734in}}{\pgfqpoint{0.049702in}{0.028866in}}{\pgfqpoint{0.039284in}{0.039284in}}%
\pgfpathcurveto{\pgfqpoint{0.028866in}{0.049702in}}{\pgfqpoint{0.014734in}{0.055556in}}{\pgfqpoint{0.000000in}{0.055556in}}%
\pgfpathcurveto{\pgfqpoint{-0.014734in}{0.055556in}}{\pgfqpoint{-0.028866in}{0.049702in}}{\pgfqpoint{-0.039284in}{0.039284in}}%
\pgfpathcurveto{\pgfqpoint{-0.049702in}{0.028866in}}{\pgfqpoint{-0.055556in}{0.014734in}}{\pgfqpoint{-0.055556in}{0.000000in}}%
\pgfpathcurveto{\pgfqpoint{-0.055556in}{-0.014734in}}{\pgfqpoint{-0.049702in}{-0.028866in}}{\pgfqpoint{-0.039284in}{-0.039284in}}%
\pgfpathcurveto{\pgfqpoint{-0.028866in}{-0.049702in}}{\pgfqpoint{-0.014734in}{-0.055556in}}{\pgfqpoint{0.000000in}{-0.055556in}}%
\pgfpathlineto{\pgfqpoint{0.000000in}{-0.055556in}}%
\pgfpathclose%
\pgfusepath{stroke,fill}%
}%
\begin{pgfscope}%
\pgfsys@transformshift{3.668880in}{1.099557in}%
\pgfsys@useobject{currentmarker}{}%
\end{pgfscope}%
\end{pgfscope}%
\begin{pgfscope}%
\definecolor{textcolor}{rgb}{0.000000,0.000000,0.000000}%
\pgfsetstrokecolor{textcolor}%
\pgfsetfillcolor{textcolor}%
\pgftext[x=3.968880in,y=1.041224in,left,base]{\color{textcolor}\sffamily\fontsize{12.000000}{14.400000}\selectfont 25k iters}%
\end{pgfscope}%
\begin{pgfscope}%
\pgfsetrectcap%
\pgfsetroundjoin%
\pgfsetlinewidth{2.509375pt}%
\definecolor{currentstroke}{rgb}{1.000000,0.498039,0.054902}%
\pgfsetstrokecolor{currentstroke}%
\pgfsetdash{}{0pt}%
\pgfpathmoveto{\pgfqpoint{3.502213in}{0.854928in}}%
\pgfpathlineto{\pgfqpoint{3.668880in}{0.854928in}}%
\pgfpathlineto{\pgfqpoint{3.835547in}{0.854928in}}%
\pgfusepath{stroke}%
\end{pgfscope}%
\begin{pgfscope}%
\pgfsetbuttcap%
\pgfsetroundjoin%
\definecolor{currentfill}{rgb}{1.000000,0.498039,0.054902}%
\pgfsetfillcolor{currentfill}%
\pgfsetlinewidth{1.003750pt}%
\definecolor{currentstroke}{rgb}{1.000000,0.498039,0.054902}%
\pgfsetstrokecolor{currentstroke}%
\pgfsetdash{}{0pt}%
\pgfsys@defobject{currentmarker}{\pgfqpoint{-0.055556in}{-0.055556in}}{\pgfqpoint{0.055556in}{0.055556in}}{%
\pgfpathmoveto{\pgfqpoint{0.000000in}{-0.055556in}}%
\pgfpathcurveto{\pgfqpoint{0.014734in}{-0.055556in}}{\pgfqpoint{0.028866in}{-0.049702in}}{\pgfqpoint{0.039284in}{-0.039284in}}%
\pgfpathcurveto{\pgfqpoint{0.049702in}{-0.028866in}}{\pgfqpoint{0.055556in}{-0.014734in}}{\pgfqpoint{0.055556in}{0.000000in}}%
\pgfpathcurveto{\pgfqpoint{0.055556in}{0.014734in}}{\pgfqpoint{0.049702in}{0.028866in}}{\pgfqpoint{0.039284in}{0.039284in}}%
\pgfpathcurveto{\pgfqpoint{0.028866in}{0.049702in}}{\pgfqpoint{0.014734in}{0.055556in}}{\pgfqpoint{0.000000in}{0.055556in}}%
\pgfpathcurveto{\pgfqpoint{-0.014734in}{0.055556in}}{\pgfqpoint{-0.028866in}{0.049702in}}{\pgfqpoint{-0.039284in}{0.039284in}}%
\pgfpathcurveto{\pgfqpoint{-0.049702in}{0.028866in}}{\pgfqpoint{-0.055556in}{0.014734in}}{\pgfqpoint{-0.055556in}{0.000000in}}%
\pgfpathcurveto{\pgfqpoint{-0.055556in}{-0.014734in}}{\pgfqpoint{-0.049702in}{-0.028866in}}{\pgfqpoint{-0.039284in}{-0.039284in}}%
\pgfpathcurveto{\pgfqpoint{-0.028866in}{-0.049702in}}{\pgfqpoint{-0.014734in}{-0.055556in}}{\pgfqpoint{0.000000in}{-0.055556in}}%
\pgfpathlineto{\pgfqpoint{0.000000in}{-0.055556in}}%
\pgfpathclose%
\pgfusepath{stroke,fill}%
}%
\begin{pgfscope}%
\pgfsys@transformshift{3.668880in}{0.854928in}%
\pgfsys@useobject{currentmarker}{}%
\end{pgfscope}%
\end{pgfscope}%
\begin{pgfscope}%
\definecolor{textcolor}{rgb}{0.000000,0.000000,0.000000}%
\pgfsetstrokecolor{textcolor}%
\pgfsetfillcolor{textcolor}%
\pgftext[x=3.968880in,y=0.796595in,left,base]{\color{textcolor}\sffamily\fontsize{12.000000}{14.400000}\selectfont 75k iters}%
\end{pgfscope}%
\end{pgfpicture}%
\makeatother%
\endgroup%